\documentclass[sigconf]{acmart}

\AtBeginDocument{%
  }

\setcopyright{acmlicensed}
\copyrightyear{2018}
\acmYear{2018}
\acmDOI{XXXXXXX.XXXXXXX}
\acmConference[Conference acronym 'XX]{Make sure to enter the correct
  conference title from your rights confirmation email}{June 03--05,
  2018}{Woodstock, NY}
\acmISBN{978-1-4503-XXXX-X/2018/06}

\usepackage{xspace}
\newcommand{\method}{LiveHouse-TS\xspace}
\usepackage{color}
\usepackage{booktabs}
\usepackage{pifont} 
\usepackage{enumitem} 
\usepackage{listings}
\usepackage{graphicx}
\usepackage{subcaption}
\usepackage[most]{tcolorbox}
\usepackage{multirow}
\usepackage{tabularx}

\usepackage{xcolor}

\newcommand{\appref}[1]{\textcolor{olive}{Appx. ~\ref{#1}}}

\renewcommand\footnotetextcopyrightpermission[1]{}

\usepackage{listings}
\usepackage{xcolor}

\definecolor{keyword}{rgb}{0.26, 0.44, 0.76}
\definecolor{comment}{rgb}{0.5, 0.5, 0.5}
\definecolor{string}{rgb}{0.56, 0.93, 0.56}
\definecolor{backcolour}{rgb}{0.95, 0.95, 0.92}

\lstdefinestyle{bashbox}{
    language=bash,
    basicstyle=\ttfamily\footnotesize,
    keywordstyle=\color{keyword}\bfseries,
    commentstyle=\color{comment}\itshape,
    stringstyle=\color{string},
    backgroundcolor=\color{backcolour},
    showstringspaces=false,
    breaklines=true,
    frame=single,
    captionpos=b
}
\lstdefinestyle{codebox}{
    basicstyle=\ttfamily\footnotesize,
    backgroundcolor=\color{backcolour},
    showstringspaces=false,
    breaklines=true,
    frame=single,
    captionpos=b
}
\begin{document}

\title{LiveHouse-TS: An Open-world Living Benchmark \\ for  Time Series Foundation Models}





\author{Haomin Wen\textsuperscript{\rm 2,\dag}, Ziyu Zhou\textsuperscript{\rm 1,\dag},  Qingxiang Liu\textsuperscript{\rm 1,\dag},  Siru Zhong\textsuperscript{\rm 1,\dag}, Yuxuan Liang\textsuperscript{\rm 1,*}} 
\affiliation{%
  \institution{
  \textsuperscript{\rm 1}The Hong Kong University of Science and Technology (Guangzhou) \hspace{0.1em} \\
  \textsuperscript{\rm 2} Shanghai Innovation Institute; \\
  \textsuperscript{$\dag$}  Equal contribution; \textsuperscript{*} Corresponding author\\ 
  }
  \city{} 
  \state{}
  \country{}
}

\email{{wenhaomin.whm,zziyuzhou,qingxiangliu737}@gmail.com,}
\email{yuxliang@outlook.com, szhong691@connect.hkust-gz.edu.cn}

\renewcommand{\shortauthors}{Trovato et al.}


\begin{abstract}
    Time Series Foundation Models (TSFMs) have recently emerged as a highly promising paradigm for cross-domain zero-shot forecasting. However, existing evaluation protocols predominantly rely on static benchmarks with fixed historical test windows. While these benchmarks provide a valuable baseline snapshot, they evaluate an average performance on a fixed history, failing to capture how models behave in continuously evolving real-world environments characterized by seasonal variations, distribution shifts, and unexpected events. To bridge this gap, we introduce LiveHouse-TS, the first open-world living benchmark infrastructure for TSFMs.  By evaluating models prequentially on real future data in open-world environments, LiveHouse-TS shifts time series benchmarking from snapshot accuracy to continuous temporal validity. Rather than acting as a one-off leaderboard, our infrastructure serves as a continuous time series infrastructure designed to explore vital, long-term scientific questions: Can model rankings be maintained over the long term? Which models remain genuinely robust under distribution shifts? Extensive streaming evaluations across 11 domains with 17 datasets demonstrate that static rankings undergo a dramatic reshuffling under a live protocol. Code, dataset, and leaderboard are available at: \url{https://huggingface.co/spaces/CityMindDev/LiveHouse-TS}.

    
\end{abstract}





\maketitle

\section{Introduction}  \label{sec:intro}
Time series forecasting is a foundational task across a wide spectrum of industrial and scientific domains, ranging from energy management and financial planning to climate modeling. Time series foundation models (TSFMs) have shown to be highly promising paradigm for zero-shot forecasting across domains, driven by large-scale pretraining and the ability to perform  zero-shot inference~\citep{garza2023timegpt,das2024timesfm,ansari2024chronos,woo2024moirai,goswami2024moment}. This paradigm shift has sparked massive research interest, yielding hundreds of papers in the last two years.

\begin{figure}[!t]
    \centering
    \includegraphics[width=1\linewidth]{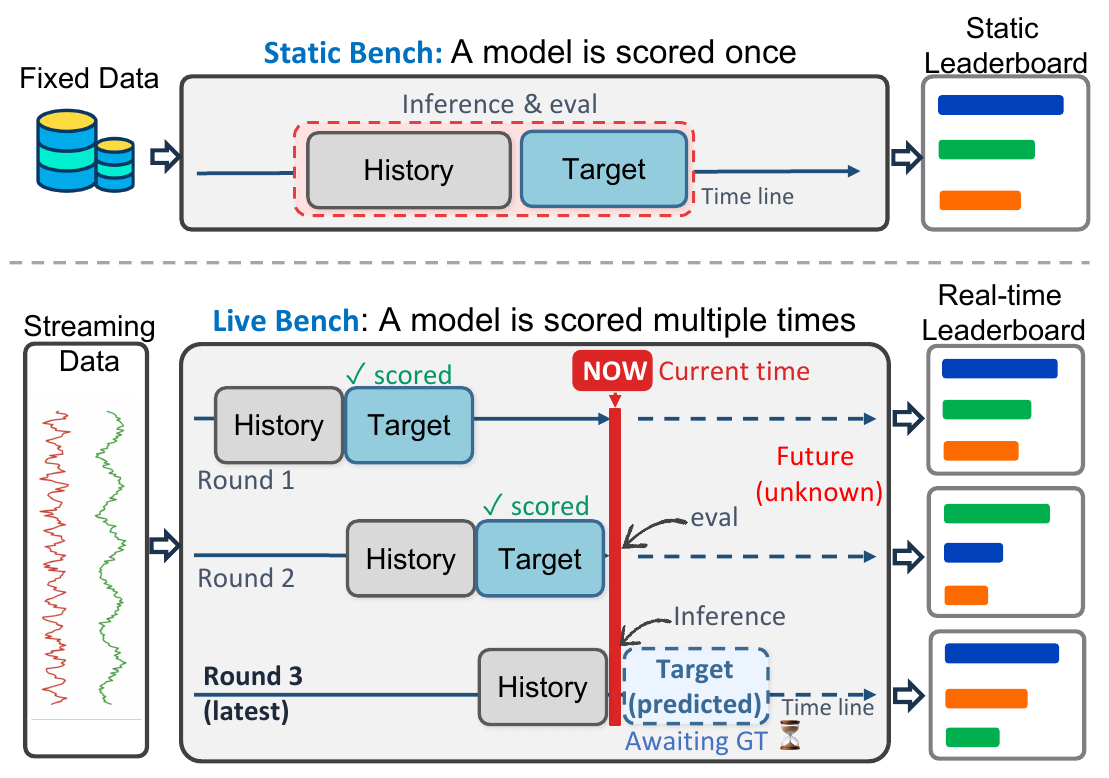}
    \caption{Comparison between proposed \method and static benchmark. \method introduce a streaming evaluation protocol to capture model robustness under real-world operational conditions. It eliminates data leakage by requiring models to forecast at the current world time before ground truth is observed, with evaluation and the leaderboard updated continuously as new data arrives.}
    \label{fig:intro}
\end{figure}

Concurrently, the rapid evolution of these models drives the demand for reliable evaluation protocols. As shown in Figure~\ref{fig:intro}, current practice relies almost entirely on static benchmarks (e.g., GIFT-eval~\citep{aksu2024gifteval}, fev-bench~\citep{shchur2025fevbench}, TSFM-Bench~\cite{li2025tsfm}), where public datasets are split into predetermined, frozen train and test windows. While static leaderboards offer a controlled environment for initial verification, they introduce a fundamental limitation: snapshot evaluation ignores the operational realities of real-world deployment, where non-stationarity, concept drift, and sudden exogenous shifts continuously alter the underlying data-generating processes.  In a static benchmark, a model that captures the top spot remains there indefinitely because standard datasets yield a permanent, immutable rank once computed. In reality, two forecasting models might achieve the exact same performance metric (e.g., a Mean Absolute Error of 0.42) on a static split, rendering them indistinguishable offline. Yet, when subjected to a rolling, real-world timeline, one model might swiftly degrade under a seasonal shift while the other maintains consistent reliability. \textbf{\textit{A model's operational superiority is not a permanent attribute; its performance and ranking must be continuously tested and earned as the world changes.}}


\begin{table*}[t]
\centering
\footnotesize
\caption{Comparison between representative TSFM benchmarks and \method. (Abbr; LR: Leakage-resistant by nature; RTL: Real-time Leaderboard; OND: Open-to-new-data; FT: Forecasting Task, Prob.: Probability Forecasting). \method provides the first systematic solution for benchmarking the live performance of time series foundation models. }
\label{tab:benchmark-comparison}
\resizebox{\linewidth}{!}{%
\begin{tabular}{lccccccccccc}
\toprule
\textbf{Benchmark} & \textbf{Year} & \multicolumn{3}{c}{\textbf{Dataset}} & \multicolumn{4}{c}{\textbf{Evaluation}} & \multicolumn{3}{c}{\textbf{Leaderboard}}\\
 \cmidrule(lr){3-5} \cmidrule(lr){6-9} \cmidrule(lr){10-12}
 &  & \textbf{Live/Static} & \textbf{\#Domain} & \textbf{\#Data} & \textbf{LR} & \textbf{Test zero-shot?} & \textbf{FT} & \textbf{Multivariate} & \textbf{RTL} & \textbf{OND} & \textbf{Metrics}\\
\midrule
Monash~\citep{godahewa2021monash} & 2021 & Static & 7 & 20 & \ding{56} & \ding{56} & Point & \ding{56} & \ding{56} & \ding{56} & MASE, sMAPE, msMAPE, MAE, RMSE\\
BasicTS~\citep{shao2023basicts} & 2023 & Static & 5 & 20 & \ding{56} & \ding{56} & point & \ding{52} & \ding{56} & \ding{56} & MAE, RMSE, MAPE, WAPE\\
TFB~\citep{qiu2024tfb} & 2024 & Static & 10 & 41 & \ding{56} & \ding{56} & Point & \ding{52} & \ding{56} & \ding{52} & MAE, MSE, MASE, MSMAPE\\

ProbTS~\citep{zhang2024probts} & 2024 & Static & 6 & 12 & \ding{56} & \ding{52} & Point/Prob. & \ding{52} & \ding{56} & \ding{56} & NMAE, CRPS\\
CiK~\citep{williams2024cik} & 2025 & Static & 7 & 9 & \ding{56} & \ding{52} & Prob. & \ding{56} & \ding{56} & \ding{56} & CRPS\\

GIFT-Eval~\citep{aksu2024gifteval} & 2024 & Static & 7 & 23 & \ding{56} & \ding{52} & Point/Prob. & \ding{52} & \ding{56} & \ding{56} & MAPE, CRPS\\
fev-bench~\citep{shchur2025fevbench} & 2025 & Static & 7 & 96 & \ding{56} & \ding{52} & Point/Prob. & \ding{52} & \ding{56} & \ding{52} & MASE, SQL\\
BOOM~\citep{cohen2025boom} & 2025 & Static & 5 & - & \ding{56} & \ding{52} & Point/Prob. & \ding{52} & \ding{56} & \ding{56} & MASE, CRPS\\
TSFM-Bench \citep{li2025tsfm} & 2025 & Static & 10 & 21 & \ding{56} & \ding{52} & Point & \ding{52} & \ding{56} & \ding{56} & MAE, MSE\\
\midrule
Impermanent~\citep{garza2026impermanent} & 2026 & \textbf{Live} & 1 & 1 & \ding{52} & \ding{52} & Point/Prob. & \ding{52} & \ding{56} & \ding{56} & MASE, CRPS\\

TS-Arena ~\citep{Meyer_2026} & 2026 & Live & 1 & 3 & \ding{52} & \ding{52} & Point & \ding{52} & \ding{52}  & \ding{52}  & MASE \\

\method & 2026 & \textbf{Live} & 11 & 17 & \ding{52} & \ding{52} & Point/Prob. & \ding{52} & \ding{52} & \ding{52} & RMSE,MAPE, CRPS, \textbf{Stability, Improvement}\\
\bottomrule
\end{tabular}%
}
\end{table*}


To address these limitations, we introduce \method, the first open-world living benchmark infrastructure for TSFMs. We define our open-world setting as a temporally open system featuring continuous streaming observations and an extensible registry for dynamically expanding data sources. As illustrated in Figure~\ref{fig:intro}, \method enforces a strict prequential evaluation protocol: predictions must be made at the current world time before the corresponding ground-truth values exist, with metrics updated continuously as new observations arrive. Ultimately, \method shifts time series benchmarking from snapshot accuracy to continuous temporal validity. We list the detailed comparsion of \method and current benchmarks in Table~\ref{tab:benchmark-comparison}. Crucially, rather than serving as a one-off leaderboard, \method is conceptualized as a living time series infrastructure designed to spark and systematically answer new scientific questions vital to the community. For example,  can model rankings be maintained long-term? Which models are genuinely robust under real-world deployment?  In summary, our core contributions are:

\begin{itemize}[leftmargin=*]
  \item \textbf{New Paradigm}: We identify a critical evaluation gap in the snapshot paradigm and introduce a streaming evaluation protocol centered on continuous temporal validity to capture model robustness under real-world operational conditions.
  
  \item \textbf{New Evaluation Infrastructure}: We propose \method, a leakage-resistant open-world live benchmark infrastructure featuring a real-time leaderboard, an extensible registry for streams and models, and new metrics specifically tailored for temporal stability and monotone performance improvement.
  
  \item \textbf{New Insights}: Current TSFMs generalize well for zero-shot forecasting on real future data. However, the rankings on \method differ from those on prior static benchmarks, suggesting strong performance on static benchmarks may not necessarily translate to practical deployment.
  

  
\end{itemize}

\section{Related Work}


\noindent \textbf{Time Series Foundation Models.} TSFMs are pretrained on a large cross-domain time series corpus and then applied zero-shot or with light fine-tuning to unseen datasets~\citep{wangconflux,garza2023timegpt,das2024timesfm,ansari2024chronos,woo2024moirai,goswami2024moment,unravel,cao2025,timedit}. They vary in tokenization, architecture, and pre-training objectives, including DeepAR, N-BEATS, N-HiTS, PatchTST, DLinear, and TimesNet~\citep{salinas2017deepar,oreshkin2020nbeats,challu2022nhits,nie2023patchtst,zeng2023dlinear,wu2023timesnet}; later models include Informer, Autoformer, FEDformer, Pyraformer, Crossformer, SCINet, and TiDE~\citep{zhou2021informer,wu2021autoformer,zhou2022fedformer,liu2022pyraformer,zhang2023crossformer,liu2022scinet,das2023tide}. We refer the reader to \citet{liang2024foundation} for a broader survey. To name a few examples, TimesFM~\citep{das2024timesfm} and Timer~\citep{liu2024timer} follow a decoder-only design that models time series as patches, whereas Chronos~\citep{ansari2024chronos,ansari2025chronos2} discretizes (via scaling and quantization) continuous values into a token vocabulary to reuse language-model backbones. In contrast, encoder-style masked pretraining is adopted by MOIRAI~\citep{woo2024moirai} (masked any-variate modeling) and MOMENT~\citep{goswami2024moment} (masked multi-task pretraining), while Lag-Llama~\citep{rasul2023lagllama} produces probabilistic forecasts from lag-based features. Beyond modeling choices, Toto~\citep{cohen2024toto} and TTM~\citep{ekambaram2024ttm} emphasize observability and lightweight deployment, and Time-MoE~\citep{shi2025timemoe} and Moirai-MoE~\citep{liu2024moiraimoe} scale up via sparse mixture-of-experts routing. In parallel, another line of work reprograms or fine-tunes frozen language models for forecasting~\citep{jin2024timellm,zhou2023onefitsall,Speakllm}.




\noindent \textbf{Time Series Forecasting Benchmark.} Early competitions and archives fixed the unit of comparison, as M4 and M5~\citep{makridakis2020m4,makridakis2022m5} standardized point and probabilistic scoring over large series collections and the Monash archive~\citep{godahewa2021monash} consolidated datasets into one format that later became a common pretraining source. 
A second wave then targeted fair and reproducible comparison, where TFB~\citep{qiu2024tfb} and BasicTS~\citep{shao2023basicts} control preprocessing and dataset heterogeneity while ProbTS~\citep{zhang2024probts} and CiK~\citep{williams2024cik} broaden the evaluation axis to distributional and context-aware forecasting.
Most recently, GIFT-Eval~\citep{aksu2024gifteval,li2024foundts}, TSFM-Bench~\citep{li2025tsfm}, fev-bench~\citep{shchur2025fevbench}, and BOOM~\citep{cohen2025boom} target foundation models directly to test their zero-shot ability. Across all three waves, a curated set of public series is frozen with predetermined train/test splits, and models are scored once over the held-out windows. Such a design introduces the potential data leakage and may not reflect model performance in real deployment (as discussed in Sec~\ref{sec:intro}), which motivates our live and open-world benchmark. Overall, we refer to Table~\ref{tab:benchmark-comparison} for a detailed comparison with \method and related ones.

\noindent \textbf{Live Benchmark.} A growing number of works in language and code evaluation address data contamination by making the benchmarks themselves \emph{live}, continuously refreshing test data or gating them by release date so that every scored example is released \emph{after} the model training.
LiveBench~\citep{white2024livebench}, LiveCodeBench~\citep{jain2025livecodebench}, and the multimodal LiveXiv~\citep{shabtay2025livexiv} follow this principle and are supported by literature that analyzes why static benchmarks fail once their data leaks into pretraining~\citep{sainz2023contamination,golchin2024timetravel}.
The methodological basis comes from the stream-learning literature, where prequential evaluation enforces that predictions are always made before observing the corresponding labels, thereby preventing look-ahead bias~\citep{gama2013evaluating} and enabling adaptive performance tracking under concept drift~\citep{bifet2007adwin}.
This live paradigm is well-suited for benchmarking general time series forecasting, since real-world series are generated continuously and provide a natural supply of strictly post-cutoff evaluation data. 


\begin{figure*}[t]
    \centering
    \includegraphics[width=1\textwidth]{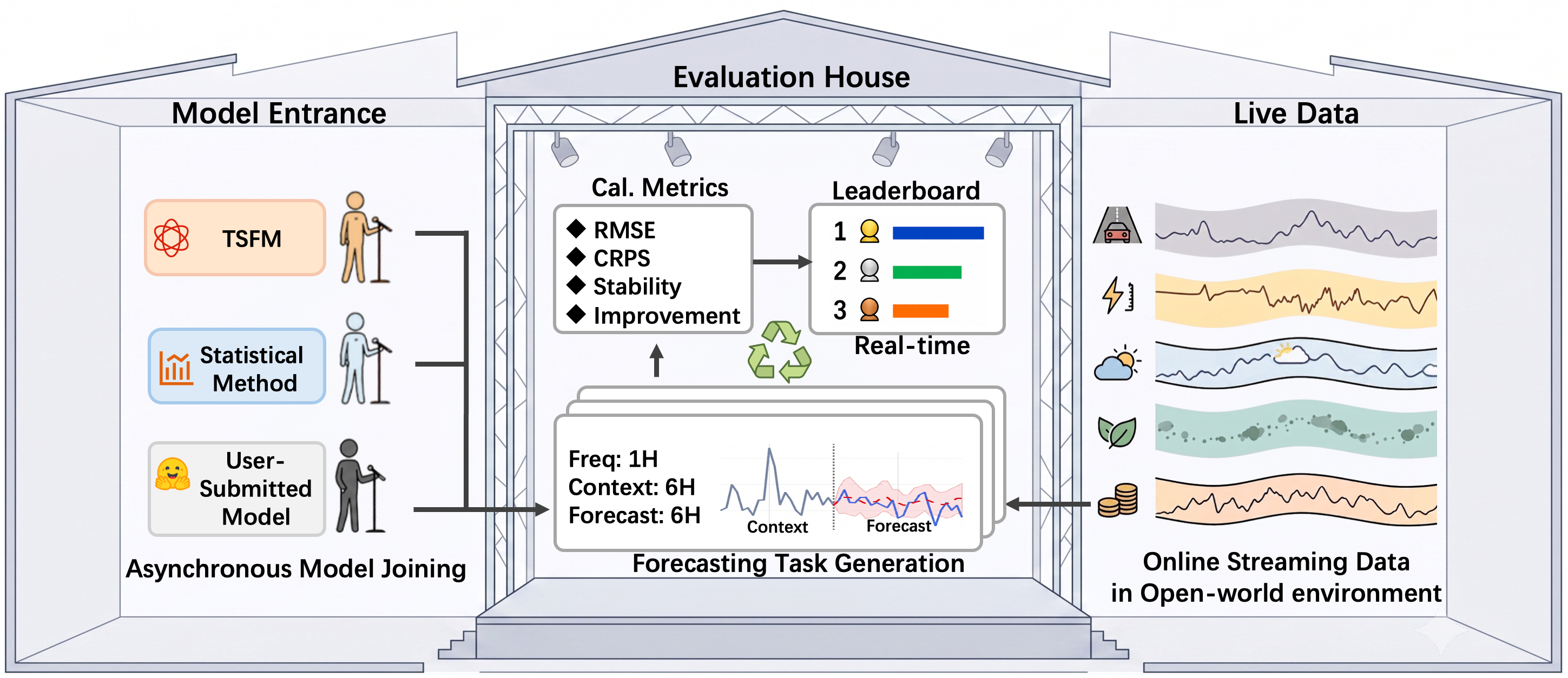}
    \vspace{-5mm}
    \caption{Overall architecture of \method, as an analogy in which models are performers and the benchmark is a \emph{live house}.  It contains three decoupled yet coordinated components: $ i)$ a \emph{Model Entrance} that ``checks tickets'' and standardizes heterogeneous forecasters before they take the stage; $ ii)$ \emph{Live Data} that turns public streams into a rolling set of forecasting tasks according to the dataset description (e.g., frequency, context length, and forecast horizon); and $ iii)$ an \emph{Evaluation House} (the live house) that enforces the future-only rule, and updates the leaderboard.
   }
    \label{fig:overall_architecture}
\end{figure*}

\section{Benchmark Details}

\subsection{Task Formulation}
We consider a time series $\{\mathbf{y}_t\}_{t\ge 1}$ observed at regular time steps, where $\mathbf{y}_t\in\mathbb{R}^D$ collects $D$ target variables. At time $t$, the goal is to forecast the next $H$ vectors given the most recent $L$ observations (with optional covariates/metadata). In this setting, a point forecasting model outputs
\begin{equation}
\hat{\mathbf{y}}_{t+1:t+H} = f\big(\mathbf{y}_{t-L+1:t},\, \mathbf{x}_{t-L+1:t},\, \mathbf{m}\big),
\end{equation}
where $\mathbf{x}$ denotes covariates (e.g., calendar features) and $\mathbf{m}$ denotes metadata such as frequency or horizon. Probabilistic forecasting instead targets a full predictive distribution over future trajectories,
\begin{equation}
P\big(\mathbf{y}_{t+1:t+H} \mid \mathbf{y}_{t-L+1:t},\, \mathbf{x}_{t-L+1:t},\, \mathbf{m}\big).
\end{equation}
\emph{Zero-shot} time series forecasting refers to applying a pretrained model to a novel dataset or unseen series without fine-tuning, using only the provided context window at inference time.

\subsection{Overall Architecture}
 
\method is guided by the following three design
principles: 
\begin{tcolorbox}[
  notitle,
  rounded corners,
  colframe=darkgray,
  colback=blue!5,
  boxrule=0.75pt,
  boxsep=0pt,
  left=0.15cm,
  right=0.17cm,
  enhanced,
  shadow={1pt}{-1pt}{0pt}{opacity=0.5,gray},
  toprule=0.75pt,
  before skip=0.65em,
  after skip=0.75em
]
\textbf{Remark: Design principles.}  \\
($i$) \textbf{Leakge-resistant:} Tasks are constructed from continually arriving public streams, and designed to test model's ability on the real future to prevent the potential data leakage.\\
($ii$) \textbf{Fairness:} Ensuring fair comparisons across methods over time, since methods that join at different times may be evaluated over different time spans. \\
($iii$) \textbf{Easy-to-scale:} It should be easy for researchers and practitioners to join the leaderboard or contribute a new data source. Since we hope \method serves as an infrastructure to evaluate the model's generalizability in the open-world environment.
\end{tcolorbox}

As in Figure~\ref{fig:overall_architecture}, we realize these principles with three decoupled yet coordinated components---as an analogy in which models are performers and the benchmark is a \emph{live house}: $ i)$ a \emph{Model Entrance} that ``checks tickets'' and standardizes heterogeneous forecasters before they take the stage; $ ii)$ \emph{Live Data} that turns public streams into a rolling set of forecasting tasks according to a dataset description (e.g., frequency, context length, and horizon); and $ iii)$ an \emph{Evaluation House} (the live house) that enforces the future-only rule, and updates the leaderboard. Details are provided in \appref{appendix:implementation_details}.

\textbf{Model Entrance,} which exposes a unified forecasting interface for forecasters, including hosted TSFMs and lightweight statistical baselines. Given a context window and dataset metadata (e.g., sampling frequency and horizon), each predictor is required to return forecasts aligned with the requested prediction horizon. The entrance adapter then validates the output shape and converts heterogeneous model outputs into a common scoring representation: a mean forecast for point-error metrics such as MSE/RMSE, a median forecast for quantile-based point metrics such as MAPE. And a fixed set of quantile forecasts at predefined levels for probabilistic metrics such as CRPS when available. If a method only provides point forecasts, we treat the point prediction as a degenerate predictive distribution for the evaluator. This canonical representation ensures that all methods, regardless of whether they are local TSFMs, user-submitted models, or statistical baselines, are scored by the same metric implementation under the same horizon and target alignment. Details in \appref{appx:model-entrance}.

\textbf{Live Data,} where collectors periodically retrieve fresh observations from multiple domains (see
Sec~\ref{sec:streaming-data} for more details). Each stream is cleaned and mapped into a shared
schema before being windowed into tasks. Rather than imposing a single global
setting, task construction follows the per-dataset specification (context length, prediction horizon, and frequency), ensuring that all models evaluated on a
given dataset receive identical inputs and targets while respecting the natural
time scale of each stream.


\textbf{Evaluation House.}
For each newly created task, the evaluation engine retrieves the historical
context available at issue time and packages it into a standardized forecasting
instance. A future-only gate then compares the task timestamp with each model's
admission time, filtering out any tasks issued before the model entered the
leaderboard. The remaining eligible tasks are dispatched through the unified
forecasting interface, and their forecasts are evaluated once the corresponding
future targets become observable. Details of the evaluation house are provided in
\appref{appx:evaluation-house}.


\subsection{Streaming Data} 
\label{sec:streaming-data}

\subsubsection{Data Curation}
\begin{figure}[htbp]
    \centering
    \includegraphics[width=\linewidth]{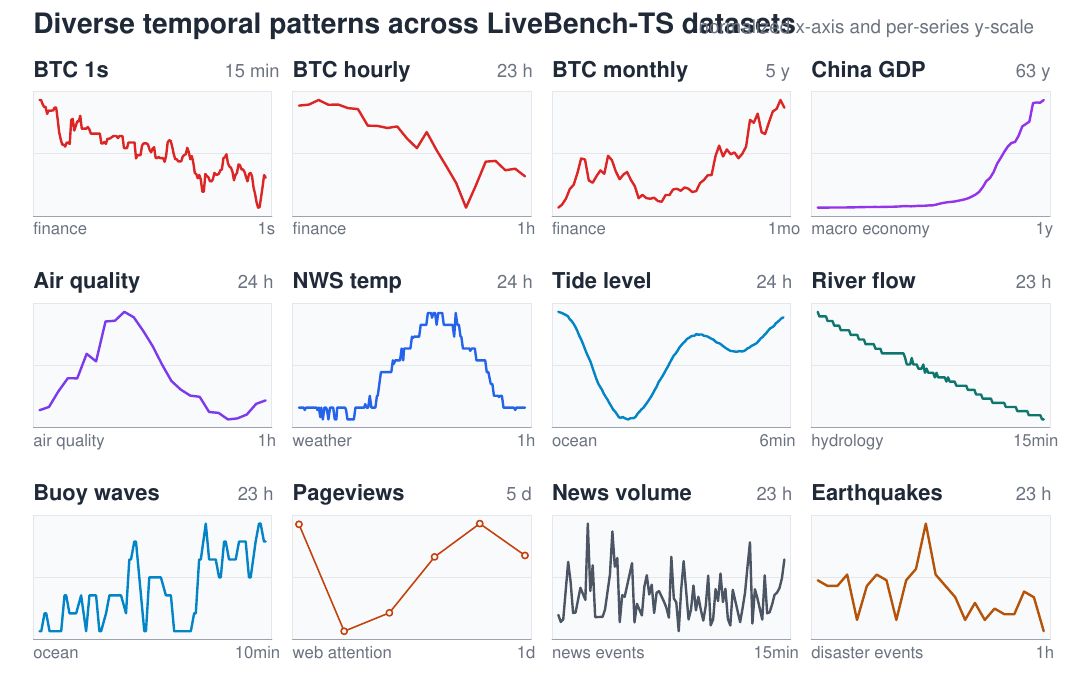}
    \caption{Dataset examples with representative temporal patterns. Each panel shows one selected dataset window at its native cadence; axes are normalized independently to highlight temporal shape rather than absolute magnitude. The upper-right label gives the displayed window span, and the lower-right label is the native sampling frequency.}
    \label{fig:overview-patterns}
\end{figure}

\method builds its evaluation stream from public, continuously updated time series rather than from a frozen archive. Figure~\ref{fig:overview-patterns} showcases representative examples from various domains. The current registry contains \textbf{17 benchmark datasets} across \textbf{15 public sources}, \textbf{11 domains}, and \textbf{8 native frequencies}, as detailed in Table~\ref{tab:registry-overview}.
The registry contains both directly reported time series (e.g., sensor readings, market prices and macro indicators) and event-derived time series, where timestamped events are aggregated into regular buckets, such as GDELT document volume and USGS earthquake counts. Each dataset is described by a single registry entry containing its source identifier, entity granularity, native data frequency, recommended evaluation frequency, history length, forecast horizon, target variable, and optional covariates. Fourteen of the seventeen datasets include covariates for multivariate evaluation.
History and forecast windows are measured in native-frequency steps. High-rate and daily operational streams provide short live contexts from minutes to weeks; while monthly and annual series preserve the longer seasonal and structural context (\appref{app:dataset-inventory}~Table~\ref{tab:dataset-inventory}). We treat the native data frequency as a property of the series. This design choice ensures that changing how often we fetch data does not alter the forecasting problem itself—only the native frequency does.

\begin{table*}[t]
\centering
\small
\caption{Overview of streaming data in \method ($n{=}17$ datasets, 15 public sources). \emph{Native freq.} is the sampling rate; \emph{Eval freq.} is the recommended evaluation frequency. Both exclude crawler polling frequency.}
\label{tab:registry-overview}
\scalebox{0.8}{
\begin{tabular}{@{}l c l l l l l@{}}
\toprule
\textbf{Group} & \textbf{\#} & \textbf{Domains} & \textbf{Native freq.} & \textbf{Eval freq.} & \textbf{Sources} & \textbf{Targets}\\
\midrule
Environment & 5 & weather, air quality, weather-energy & 1h (4), 1d & 1h (4), 1d & Open-Meteo, NASA POWER, NWS, NOAA NCEI & temperature, PM\textsubscript{2.5}\\
Water & 3 & hydrology, ocean & 6min, 10min, 15min & 1h & USGS Water, NOAA CO-OPS, NOAA NDBC & discharge, water level, wave height\\
Mobility & 1 & traffic & 15min & 15min & GBFS Citi Bike & available bikes\\
Finance & 4 & finance & 1s, 1h (2), 1mo & 1s, 1h (2), 1mo & Binance, CoinGecko & close, market price\\
Society/economy & 2 & web attention, macro-economy & 1d, 1y & 1d, 1y & Wikimedia, World Bank & pageviews, GDP\\
Events & 2 & news events, disaster events & 15min, 1h & 1d & GDELT, USGS Earthquake & event volume, count\\
\bottomrule
\end{tabular}}
\end{table*}

\subsubsection{Data Characteristics}
\par  The curated datasets have two properties: diversity (covering qualitatively different forecasting regimes) and liveness (streaming coming data). 


\begin{figure}[t]
\centering
\includegraphics[width=\linewidth]{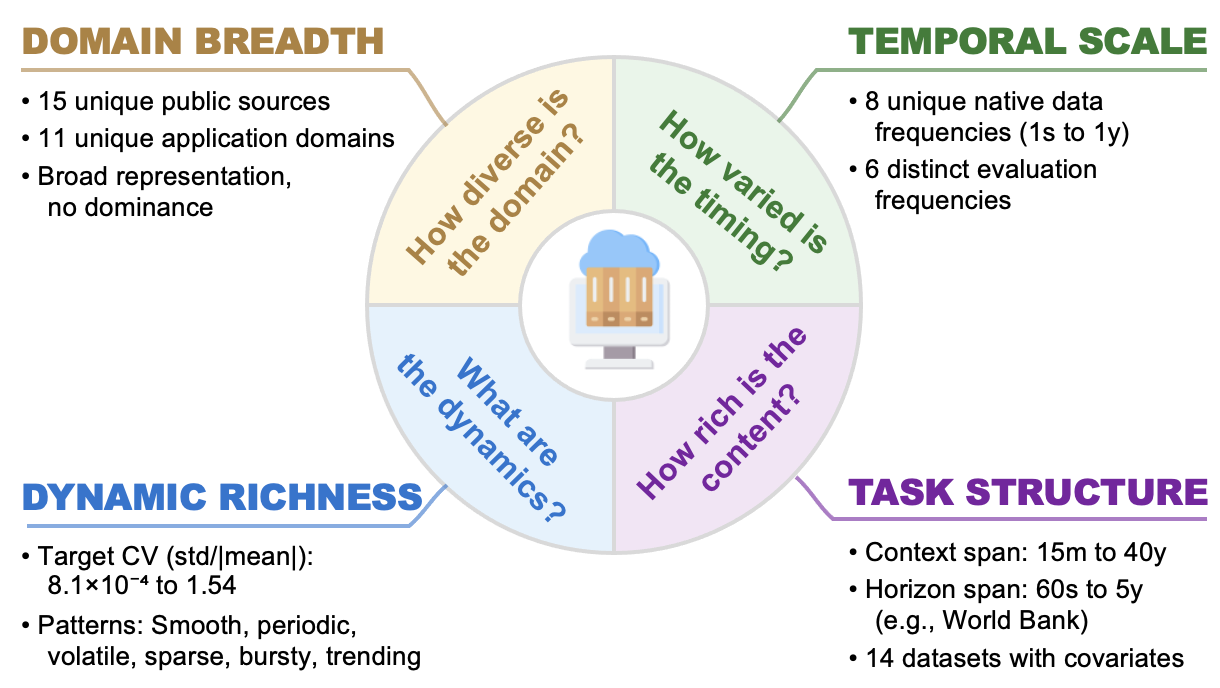}
\caption{\textbf{Four measurements for dataset diversity.} \method is diverse enough both in terms of domain breadth, temporal scale, statistical variability, and task structure.}
\vspace{-2em}
\label{fig:dataset-diversity}
\end{figure}
\begin{figure}[htbp]
    \centering
    \vspace{-1em}
    \includegraphics[width=0.9 \linewidth]{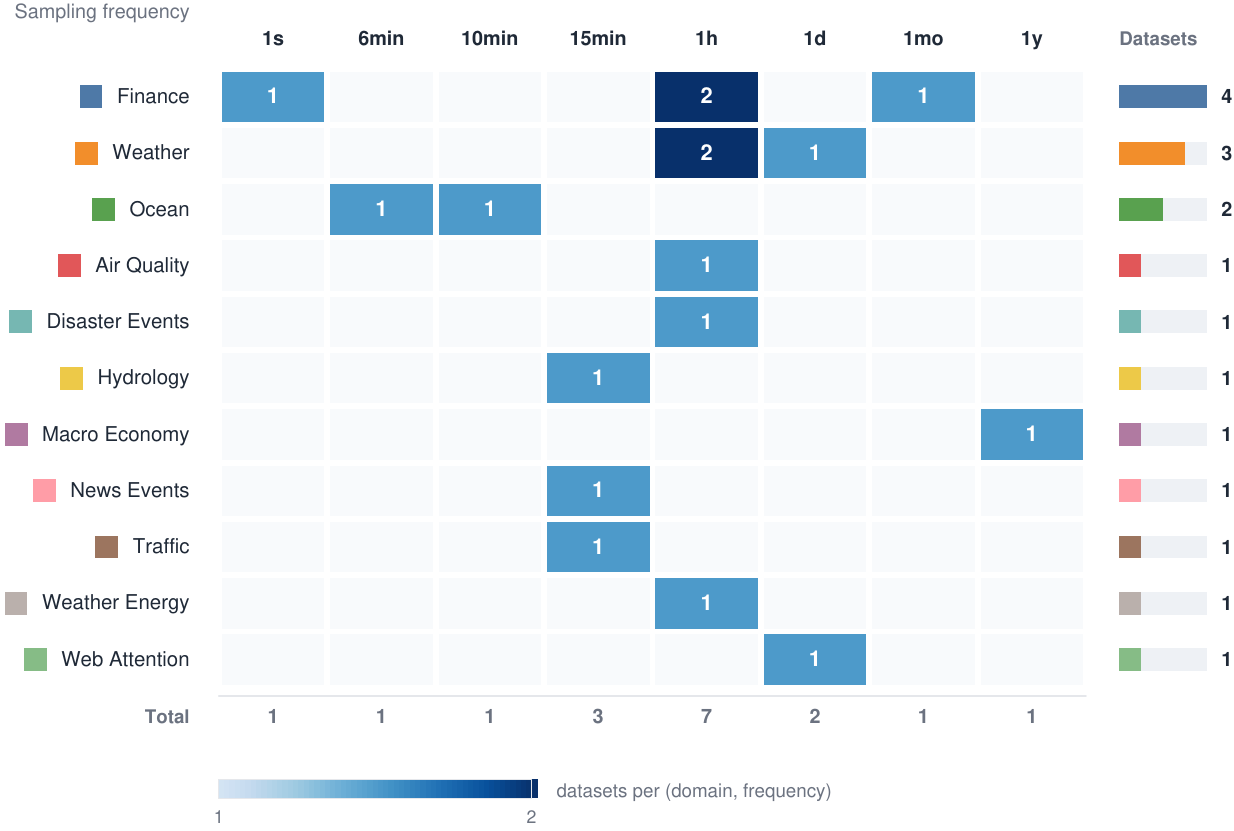}
    \caption{Coverage by domain and native sampling frequency. \method evaluation covers from short-horizon high-rate forecast to long-horizon low-rate forecast.}
    \label{fig:domain-overview}
\end{figure}

\textbf{Diversity.}
To avoid a high score being driven by matching a single domain, sampling rate, or smoothness pattern, we curate datasets along four complementary axes: \emph{source/domain breadth}, \emph{temporal-scale breadth}, \emph{task-structure breadth}, and \emph{dynamic richness} (Figure~\ref{fig:dataset-diversity}).
Concretely, the current registry spans 15 public sources, 11 domains, and 8 native frequencies from one second to one year.
Figure~\ref{fig:domain-overview} summarizes this coverage by domain, native sampling frequency, and their joint distribution across the 17 datasets.
Hourly series are the most common (7 datasets), followed by 15-minute (3) and daily (2); the remaining datasets cover 1s, 6min, 10min, 1mo, and 1y regimes.
This range allows \method to evaluate short-horizon high-rate forecasting, ordinary sensor forecasting, event-volume forecasting, and slow low-frequency forecasting under one protocol.  Moreover, \method is paried with dynamic richness diverse temporal behaviors such as seasonality, bursts, and regime shifts, and task-structure breadth that spans distinct forecasting setups (e.g., horizons, targets, and available covariates).

\textbf{Liveness.}
Observations are collected from public sources as they are released, which enables evaluation on values that post-date a model's participation in the leaderboard.
The leaderboard therefore evolves as new collection rounds complete and previously issued forecasts become scoreable.
Every canonical observation row links back to a raw response record and parser version, and the verification run parsed 2{,}672 observations across all 17 datasets.
To keep the main paper focused, per-dataset registry fields and verification-slice statistics are reported in \appref{app:dataset-inventory}.


\subsubsection{Data pipeline}
The crawled raw data are archived with request metadata, and converted into the forecasting task used for model evaluation. A future-only filter drops any context row whose observation was not yet available at forecast issue time.
\appref{app:data-process} details collection, canonical parsing, and task filtering.
\subsection{Evaluation Mechanism}
\label{sec_exp_metrics}
\subsubsection{Metrics.} Like most static benchmark, for point accuracy we use RMSE ($\downarrow$) computed on $z$-normalized series to make magnitudes comparable across datasets, and we additionally report MAPE ($\downarrow$) when targets are bounded away from zero. For probabilistic forecasts we use CRPS ($\downarrow$), estimated from the quantiles emitted by each model. To summarize performance across datasets, we report Average Rank~\citep{zhang2026mitra} ($\downarrow$) across datasets, Win Rate~\citep{zhang2026mitra} ($\uparrow$) from pairwise wins, and an Elo rating~\citep{elo1978rating} ($\uparrow$) that weights wins over stronger opponents more heavily and yields a robust leaderboard score.  We refer to \appref{appx:exp_metrics} for more details.

Moreover, since \method is a live benchmark, we additionally introduce live-specific metrics that capture what static leaderboards cannot. Let $s_t$ denote a base metric computed within a time window (i.e., a day) indexed by $t\in\{1,\dots,T\}$ on a forecasting task. We define \emph{Temporal Stability} ($\downarrow$) as the standard deviation over evaluations; it tests how stable a model's performance is over time:
\begin{equation}
\mathrm{Stability}=\sqrt{\frac{1}{T-1}\sum_{t=1}^{T}(s_t-\bar{s})^2},\qquad \bar{s}=\frac{1}{T}\sum_{t=1}^{T}s_t,
\end{equation}
We next introduce  \emph{Improvement} ($\downarrow$) to test how reliably a model improves as new targets are revealed. Fix a target time $u$ (which will be realized later). As the benchmark evolves, the model may issue multiple forecasts for this same $u$ at different issue times $t_1<\cdots<t_{T_u}$; let $s_{u,k}$ be the resulting error (lower-is-better) once $y_u$ is revealed. We quantify the monotone trend of $\{s_{u,k}\}_{k=1}^{T_u}$ with Kendall's $\tau$,
\begin{equation}
\begin{gathered}
\tau_u = \frac{2}{T_u(T_u-1)}\sum_{1\le i<j\le T_u}\mathrm{sign}(s_{u,j}-s_{u,i}),\\
\mathrm{Improvement} = \frac{1}{|\mathcal{U}|}\sum_{u\in\mathcal{U}}\tau_u,
\end{gathered}
\end{equation}
where more negative values indicate a consistently decreasing (i.e., improving) error sequence as the issue time approaches the target. We choose Kendall's $\tau$ because it is non-linear and robust to spikes, capturing whether performance predominantly improves as new information arrives (details in \appref{appx:exp_metrics}).



\begin{figure}[!ht]
    \centering
    \includegraphics[width=\linewidth,trim=18 22 20 15,clip]{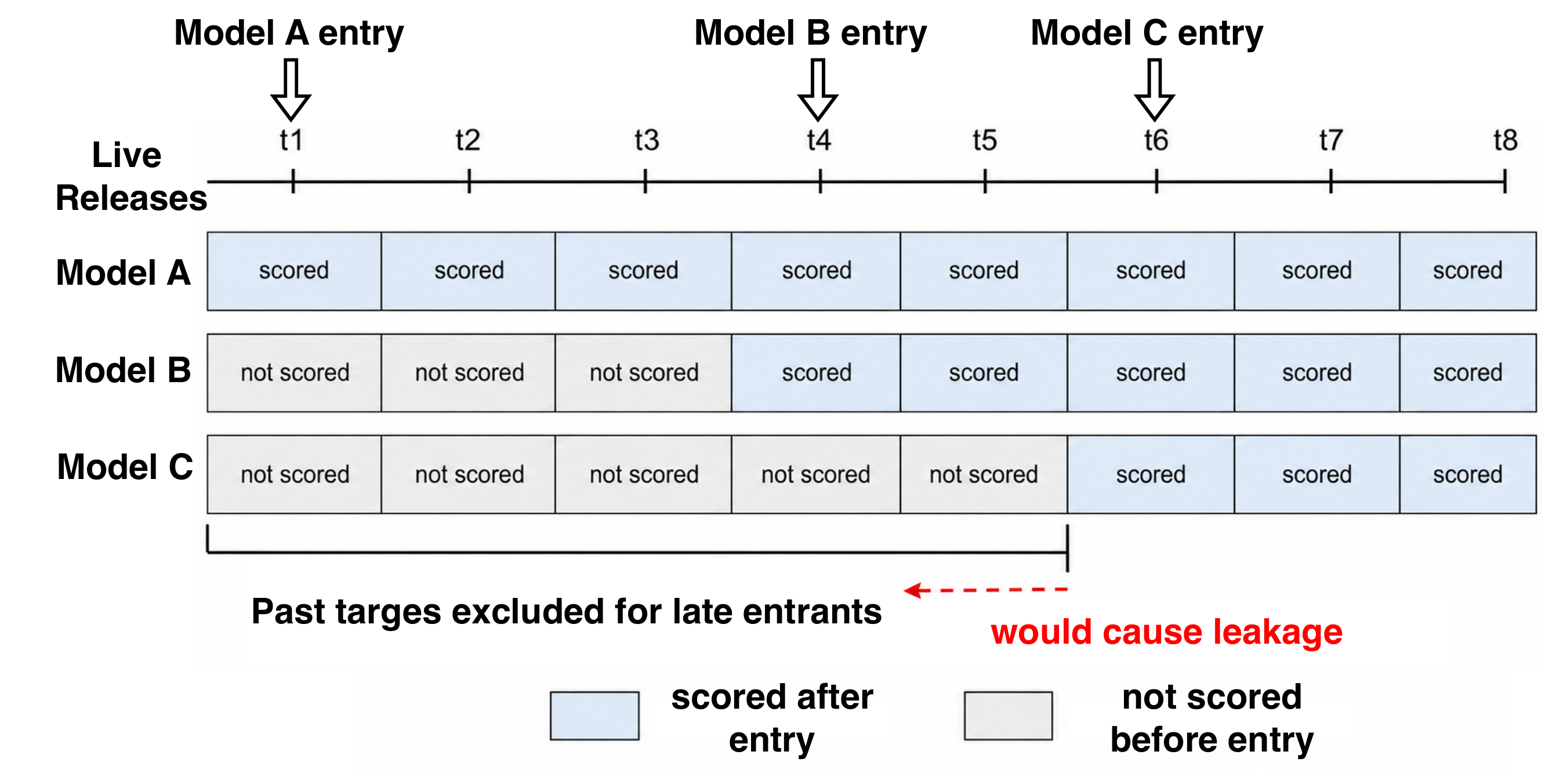}
    \vspace{-7mm}
    \caption{Future-only evaluation under asynchronous model entry: models are scored only on targets released after they join the live leaderboard, to avoid leakage.}
    \label{fig:liverelease}
\end{figure}

\begin{table*}[t]
\centering
\caption{Overall performance on \method evaluated by RMSE, MAPE, and CRPS using Average Rank ($\downarrow$), Win Rate ($\uparrow$), and Elo Rating ($\uparrow$).
All baselines are ordered by their average rank across the three metrics.
The best and second-best results in each row are highlighted in \textcolor{red}{\textbf{red bold}} and \textcolor{blue}{\underline{blue underline}}, respectively.
TSFMs consistently outperform classical statistical methods. Moirai-2.0 dominates probabilistic forecasting, while TimesFM-2.5 and Chronos--2 excel primarily in point accuracy.}
\label{tab_rq1_overall_part}
\resizebox{\linewidth}{!}{%
\begin{tabular}{l l rrrrrrrr rrrr}
\toprule
\multirow{2}{*}{Measure} & \multirow{2}{*}{Metric} & \multicolumn{8}{c}{Foundation models} & \multicolumn{4}{c}{Classical baselines} \\
\cmidrule(lr){3-10} \cmidrule(lr){11-14}
& & Moi2 & TFM & Chr2 & TiRex & TabPFN & Sundial & ChrB & Toto1 & ARIMA & MovAvg & ETS & SNaive \\
\midrule
\multirow{3}{*}{\textbf{Average Rank} ($\downarrow$)}
  & RMSE & 4.70 & \textcolor{red}{\textbf{3.30}} & \textcolor{blue}{\underline{3.80}} & 4.00 & 5.80 & 6.40 & 7.10 & 10.70 & 7.45 & 7.75 & 8.00 & 9.00 \\
  & MAPE & \textcolor{red}{\textbf{3.70}} & \textcolor{blue}{\underline{3.85}} & 4.90 & 4.30 & 5.10 & 6.45 & 6.15 & 5.05 & 8.70 & 9.65 & 9.80 & 10.35 \\
  & CRPS & \textcolor{red}{\textbf{2.30}} & 9.20 & 7.25 & 7.05 & 4.70 & 3.15 & 3.90 & \textcolor{blue}{\underline{2.95}} & 8.70 & 9.35 & 9.45 & 10.00 \\
\multirow{3}{*}{\textbf{Win Rate} ($\uparrow$)}
  & RMSE & 0.66 & \textcolor{red}{\textbf{0.79}} & \textcolor{blue}{\underline{0.75}} & 0.73 & 0.56 & 0.51 & 0.45 & 0.12 & 0.41 & 0.39 & 0.36 & 0.27 \\
  & MAPE & \textcolor{red}{\textbf{0.76}} & \textcolor{blue}{\underline{0.74}} & 0.65 & 0.70 & 0.63 & 0.51 & 0.53 & 0.63 & 0.30 & 0.21 & 0.20 & 0.15 \\
  & CRPS & \textcolor{red}{\textbf{0.88}} & 0.26 & 0.43 & 0.45 & 0.66 & 0.81 & 0.74 & \textcolor{blue}{\underline{0.82}} & 0.30 & 0.24 & 0.23 & 0.18 \\
\multirow{3}{*}{\textbf{Elo Rating} ($\uparrow$)}
  & RMSE & 1218 & \textcolor{red}{\textbf{1254}} & \textcolor{blue}{\underline{1225}} & 1194 & 1161 & 1070 & 947 & 637 & 834 & 867 & 789 & 804 \\
  & MAPE & \textcolor{red}{\textbf{1253}} & 1139 & 1163 & \textcolor{blue}{\underline{1251}} & 1177 & 998 & 1064 & 1055 & 846 & 739 & 650 & 666 \\
  & CRPS & \textcolor{red}{\textbf{1470}} & 780 & 930 & 1003 & 1243 & \textcolor{blue}{\underline{1358}} & 1265 & 1325 & 726 & 666 & 623 & 611 \\
\bottomrule
\end{tabular}}
\end{table*}

\subsubsection{Fairness} Note that in the real case, the model can join the leaderboard at different times, which introduces a significant challenge to fair comparison for the live benchmark:  As shown in Figure~\ref{fig:liverelease}, models that join later may have already observed the ground-truth outcomes from earlier forecasting rounds, so scoring them on earlier periods would give an information advantage and introduce potential data leakage. To address this challenge, we propose evaluating each model only on targets released after it joins the leaderboard. This keeps comparisons fair across entry times. 

However,  models can therefore have different evaluation horizons under the future-only evaluation paradigm; we propose two solutions $(i)$  fixed-horizon leaderboards (daily, weekly, and monthly) so that models are compared over the same evaluation window. $(ii)$ pair-wise historical ranking ranks models using only the forecasting tasks on which they were evaluated together (see more details in \appref{app:pairwise-historical-ranking}). For each eligible model pair, it computes a dataset-balanced win rate from per-release comparisons (using both RMSE and CRPS), then aggregates these pairwise win rates into a model-level score by macro-averaging over eligible opponents.

\newpage

\section{Experiments}

\label{sec_experiments}
We conduct extensive experiments to answer the following four research questions: \textbf{RQ1 (Zero-shot ability)}. Do TSFMs genuinely have strong zero-shot forecasting ability when evaluated on the real future? \textbf{RQ2 (Static vs. live rankings).} Do rankings of TSFMs on \method change significantly compared to static ones? \textbf{RQ3 (Drift robustness).} How does TSFM performance evolve as the live distribution drifts? \textbf{RQ4 (Ranking Stability.)} Can model rankings be maintained long-term?




\noindent \textbf{Models.} We evaluate six TSFMs in frozen zero-shot mode, namely TiRex~\citep{auer2025tirexzeroshotforecastinglong}, Chronos-2~\citep{ansari2025chronos2}, and TimesFM-2.5~\citep{das2024timesfm}.  
Toto-1.0~\citep{cohen2026this}, 
Moirai-2.0~\citep{liu2026moirai20timeseries}
Chronos-Bolt~\citep{ansari2024chronos},
TabPFN-TS~\citep{hoo2026tablestimeextendingtabpfnv2},
Sundial~\citep{liu2025sundial}.
We select these models as the union of the TSFMs compared in widely used benchmarks, including GIFT-Eval~\citep{aksu2024gifteval}, fev-bench~\citep{shchur2025fevbench}, and TIME~\citep{qiao2026s}. 
Collectively, these models represent major paradigms of modern TSFMs, including tokenization-based forecasting (Chronos-2), direct continuous-value prediction (Chronos-Bolt), decoder-only large-scale forecasting (TimesFM-2.5), retrieval-enhanced forecasting (TiRex), probabilistic modeling (Moirai-2.0), diffusion-based forecasting (Sundial), large-scale autoregressive pre-training (Toto-1.0), and table foundation model adaptation for time series forecasting (TabPFN-TS).
We also include four classical baselines, namely Seasonal Naive~\citep{hyndman2018fpp2}, Moving Average~\citep{box2015time}, ARIMA~\citep{box2015time}, and ETS~\citep{hyndman2002state}, which provide the evaluation baselines. 
The foundation models are run with frozen weights and a fixed context window. 
To keep the result tables compact we abbreviate the models as Chr2 for Chronos-2, TFM for TimesFM-2.5, Toto1 for Toto-1.0, Moi2 for Moirai-2.0, MovAvg for Moving-Average, TabPFN for TabPFN-TS, Chr2 for Chronos-2, and SNaive for Seasonal-Naive, while TiRex, ARIMA, and ETS keep their origianl names. 

\noindent \textbf{Settings.} 
All models are scored with a rolling mode, so that a forecast at time $t$ uses only observations up to $t$ and is graded once the ground truth arrives.
The detailed forecasting settings of each dataset are presented in \appref{app:verification}~Table~\ref{tab:dataset-inventory}.
Performance is assessed with the point metrics RMSE, and MAPE and the probabilistic metric CRPS, together with the aggregate measures Average Rank, Win Rate, and Elo detailed in  \appref{appx:exp_metrics}. 
Following GIFT-Eval~\citep{aksu2024gifteval}, each Rank reported in the tables assigns every model a per-dataset rank by the metric (best${}={}1$) and averages these ranks within the reported group. The evaluation horizon reported in the results is from 01/06/2026 to 01/07/2026. Among the 17 datasets, 10 are consistently available; the remaining 7 occasionally have missing observations during data collection. We therefore report results on the 10 stable datasets and reserve the other 7 for future benchmarks.

\subsection{RQ1. Zero-shot ability}
\label{sec_exp_rq1}

We evaluate whether TSFMs deliver strong zero-shot forecasts by comparing their predictions across the 10 eligible datasets in the live snapshot. We summarize the overall results in Table~\ref{tab_rq1_overall_part} and report per-dataset scores aggregated by domain, sampling frequency, and forecasting horizon in \appref{app_perdataset}~Table~\ref{tab_rq1_domain}, Table~\ref{tab_rq1_frequency}, and Table~\ref{tab_rq1_predlen}.

\noindent\textbf{Overall.}
Table~\ref{tab_rq1_overall_part} presents the overall performance of all baselines across three ranking metrics computed from MAE, MAPE, and CRPS, with detailed per-dataset results reported in \appref{app_perdataset}~Table~\ref{tab_rq1_overall}.
Overall, TSFMs consistently outperform statistical baselines.
Specifically, Moirai-2.0 and TimesFM-2.5 achieve the best Average Rank, whereas Toto-1.0 performs consistently worse.
The Elo and Win Rate results further reveal substantial differences among TSFMs. Moirai-2.0 dominates probabilistic forecasting, while TimesFM-2.5 excels primarily in point accuracy (also see \appref{app:pairwise-historical-ranking}~Figure~\ref{fig:overall-ranking-mse} and Figure~\ref{fig:overall-ranking-crps} for more results).

\begin{figure}[htbp]
\centering
\includegraphics[width=\linewidth]{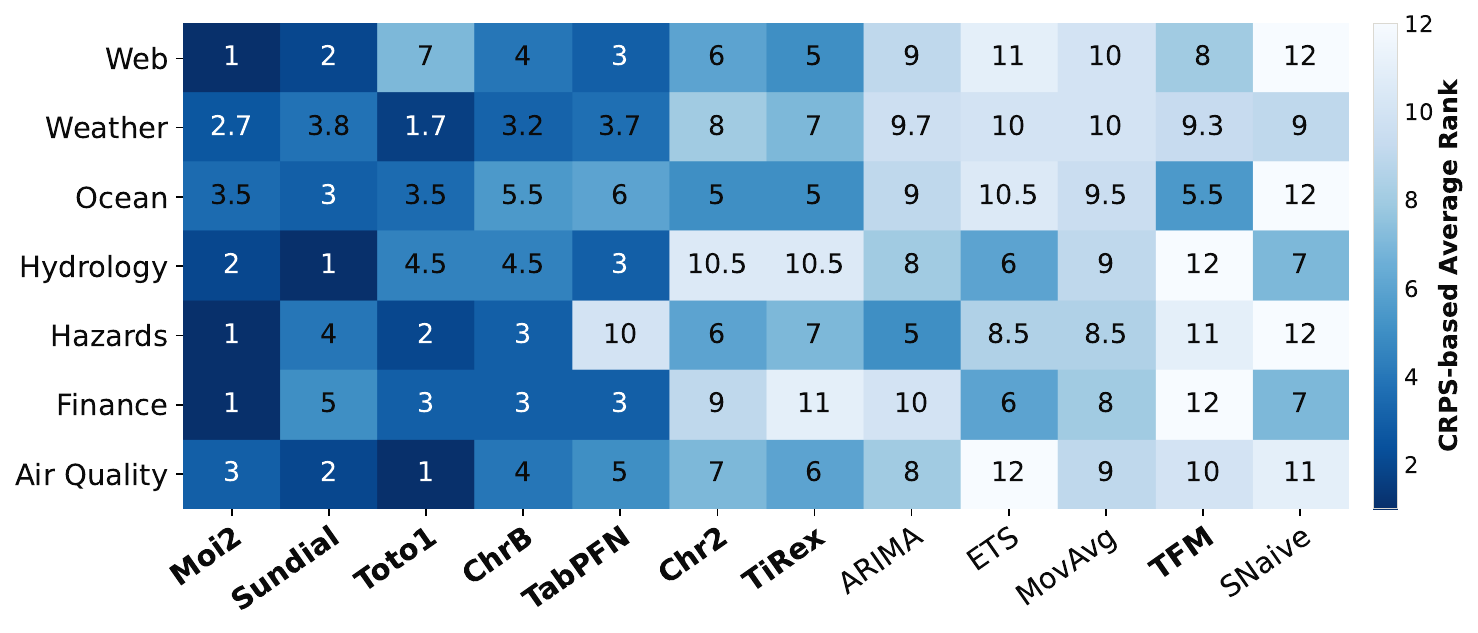}
\caption{CRPS-based average rank across seven domains. The results show that  different TSFMs excel in distinct domains. Notably, the top-ranked Moirai-2.0 performs poorly on weather-related domains (Weather, Air Quality, and Ocean).}
\label{fig_rq1_crps_heatmap}
\end{figure}

\noindent\textbf{By different domains.}
Figure~\ref{fig_rq1_crps_heatmap} summarizes CRPS-based ranks across the domains; detailed results for other metrics, sampling frequencies, and forecasting horizons are reported in \appref{app_perdataset}~Table~\ref{tab_rq1_domain}, Table~\ref{tab_rq1_frequency}, and~Table~\ref{tab_rq1_predlen}.
Moirai-2.0 stays in the Top-2 CRPS rank on most domains, and different TSFMs lead under different domains, frequencies, and forecast horizons. Notably, the top-ranked Moirai-2.0 performs poorly on weather-related domains (Weather, Air Quality, and Ocean), which highlights a clear opportunity to improve performance on such domains.


\noindent\textbf{Findings.}
TSFMs consistently outperform classical statistical baselines in the zero-shot setting, yet no single model dominates both point forecasting and probabilistic forecasting. Among tested TSFMs, Moirai-2.0 achives the best in probabilistic forecasting while  TimesFM-2.5 and TiRex are good at point forecasting. 


\subsection{RQ2. Static versus live rankings}
\label{sec_exp_rq2}
\begin{figure}[htbp]
\centering
\includegraphics[width=0.9 \linewidth]{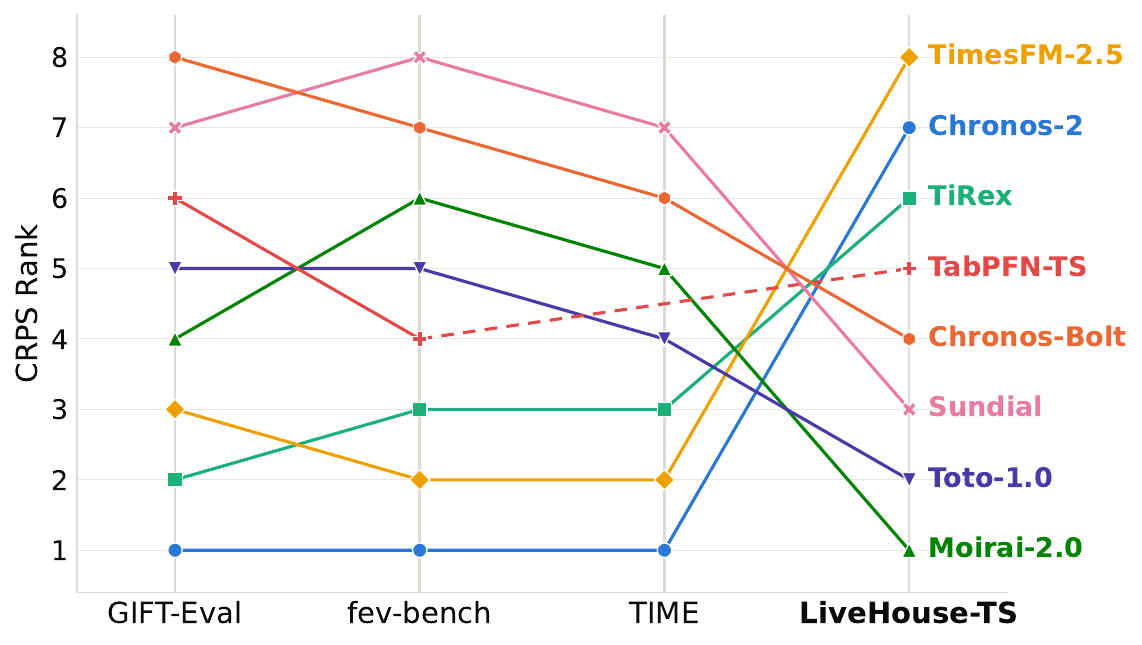}
\caption{CRPS ranking of the shared TSFMs across three static benchmarks and the live benchmark \method. The dotted segment for TabPFN-TS indicates that it is not evaluated in TIME. The three static benchmarks reach a similar consensus, whereas the live benchmark produces a pronounced ranking inversion, showing that offline (static) rankings can diverge from online performance and that a live benchmark is necessary to assess models under real deployment conditions.}
\label{fig_rq2_rank}
\end{figure}

To avoid biases in a single leaderboard, we compare \method with three popular static benchmarks, i.e., GIFT-Eval~\citep{aksu2024gifteval}, fev-bench~\citep{shchur2025fevbench}, and TIME~\citep{qiao2026s}. The comparison is restricted to the eight shared TSFMs with identical model versions (TabPFN-TS is unavailable in TIME). All benchmarks are ranked by CRPS. Figure~\ref{fig_rq2_rank} presents the ranking comparison.

\noindent\textbf{Findings.}
The three static benchmarks exhibit remarkable agreement. Chronos-2 consistently ranks first, followed by TiRex and TimesFM-2.5, while Chronos-Bolt and Sundial remain near the bottom. In contrast, \method produces a quite different ranking. Moirai-2.0 and Toto-1.0 rise to the top, while Chronos-2 and TimesFM-2.5 fall to the bottom. Together with the RQ1 observation that Chronos-2 and TimesFM-2.5 achieve strong point accuracy but poor probabilistic calibration, these results suggest that static benchmarks fail to capture aspects of robustness that become apparent only under continuous live evaluation.


\subsection{RQ3. Drift robustness}
\label{sec_exp_rq3}

\begin{figure}[htbp]
\centering
\includegraphics[width=\linewidth]{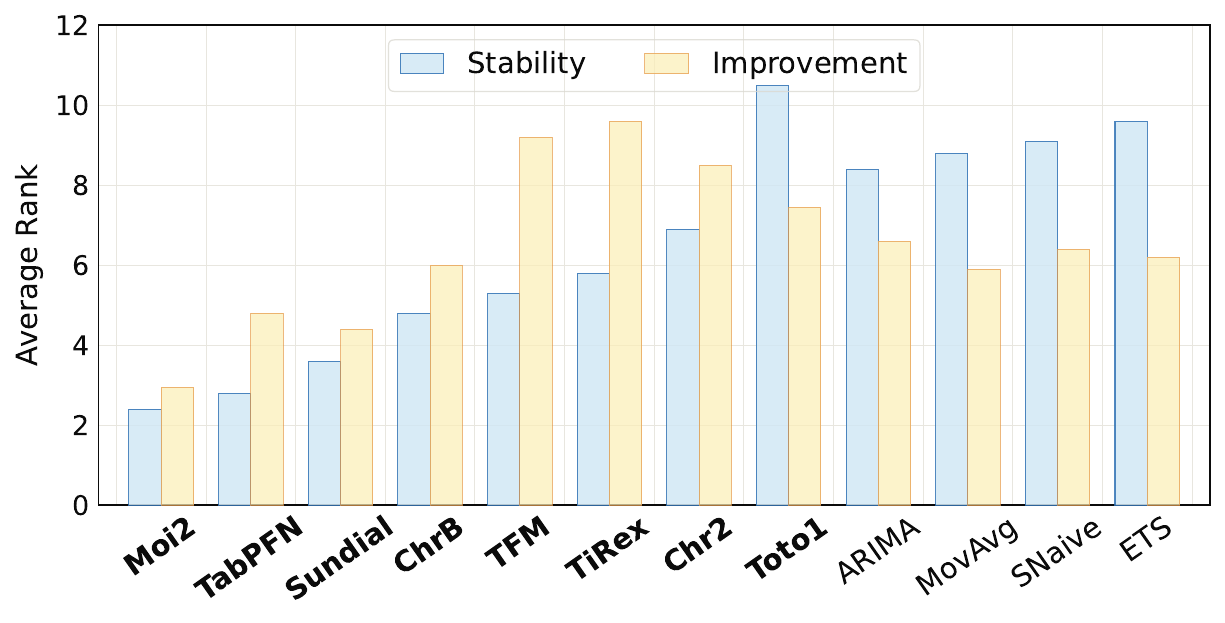}
\caption{Average Rank of Temporal Stability (blue) and Improvement (yellow). Moirai-2.0 ranks highest on both drift-robustness metrics, while several top static performers exhibit substantially lower temporal stability.}
\label{fig_rq3_stability_improvement}
\end{figure}

Unlike static benchmarks, which evaluate a fixed test set, the live benchmark continuously assesses baselines on newly arriving observations and therefore reveals their robustness to temporal distribution shift.
Figure~\ref{fig_rq3_stability_improvement} reports the Average Rank of the proposed Temporal Stability ($\downarrow$) and Improvement ($\downarrow$) metrics.
The detailed numderical results are provided in \appref{app_perdataset}~Table~\ref{tab_perdataset_stability} and~Table~\ref{tab_perdataset_improvement}.
Moirai-2.0 achieves the best performance on both metrics, consistent with its top ranking in the live benchmark.
In contrast, Chronos-2 and TimesFM-2.5, which consistently lead the static benchmarks, rank near the bottom among TSFMs, indicating substantial degradation under temporal drift.
Toto-1.0 further illustrates the difference between average accuracy and robustness. Despite ranking second in live CRPS, its Stability rank is only 10.50, revealing large performance fluctuations over time.

\noindent\textbf{Findings.} 
Drift robustness largely explains the ranking inversion observed in RQ2.
Models that remain stable under evolving data also achieve stronger live benchmarking performance, while strong static accuracy alone does not guarantee robust deployment.
These complementary metrics provide aspects of forecasting quality that static evaluations cannot capture.

\subsection{RQ4. Ranking Stability}
\label{sec_rank_stab}
\begin{figure}[htbp]
\centering
\includegraphics[width=\linewidth]{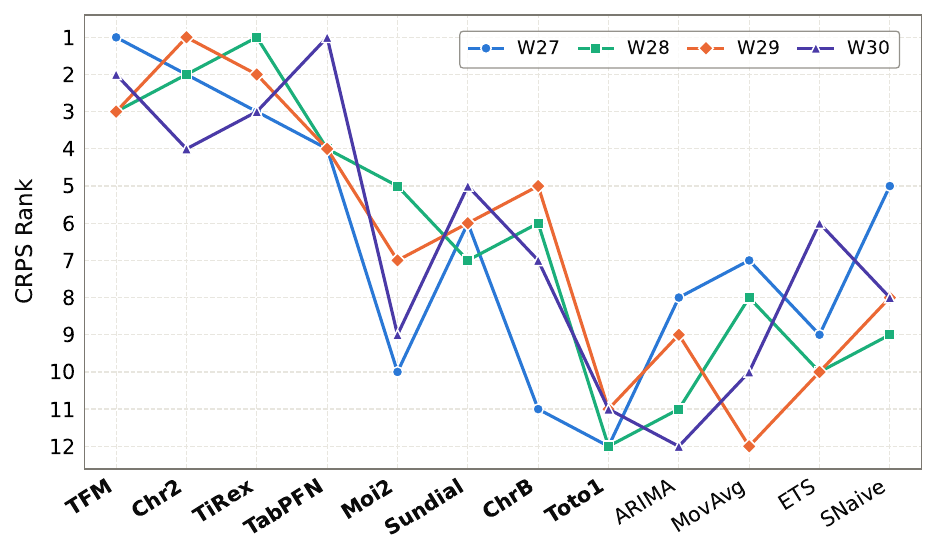}
\caption{CRPS rankings over four consecutive weekly snapshots. The dynamic rankings demonstrate that model performance evolves with the incoming data stream, highlighting the need for continuous live evaluation rather than a single leaderboard snapshot.}
\label{fig_rq4_rank_stability}
\end{figure}

Can model
rankings be maintained long-term? In this subsection, we investigate whether model rankings remain stable across consecutive weekly snapshots of the live benchmark. Unlike RQ2, which compares static leaderboards with \method, this analysis focuses on temporal ranking dynamics under the same evaluation protocol. Figure~\ref{fig_rq4_rank_stability} shows CRPS rankings from W27 to W30 for all twelve baselines.

\begin{figure*}[!t]
\centering
\begin{subfigure}[t]{0.49\linewidth}
\centering
\includegraphics[width=\linewidth]{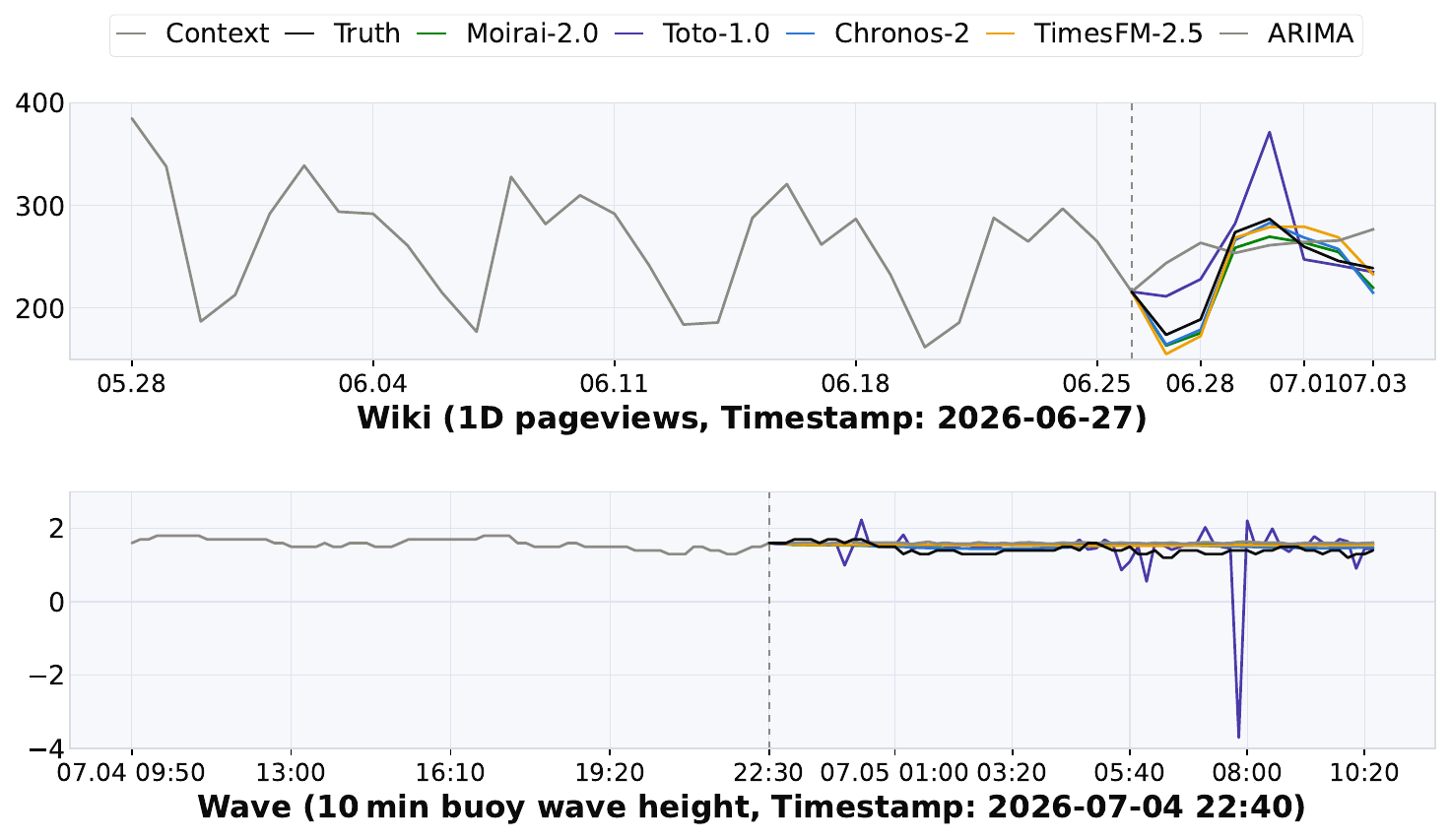}
\caption{Wiki and Wave.}
\end{subfigure}\hfill
\begin{subfigure}[t]{0.49\linewidth}
\centering
\includegraphics[width=\linewidth]{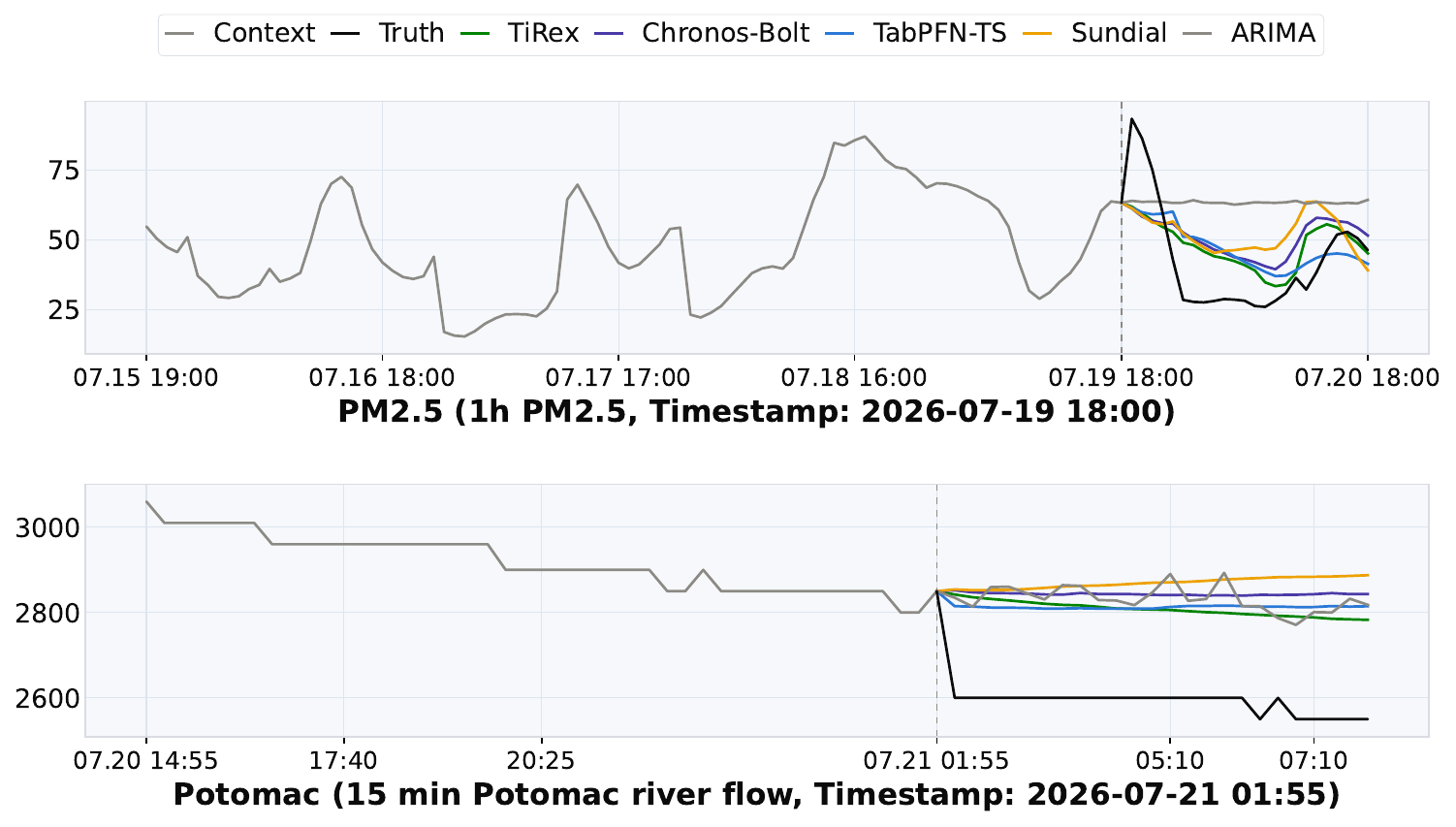}
\caption{PM2.5 and Potomac.}
\end{subfigure}
\caption{Forecasting visualization on Wiki, Wave, PM~2.5, and Potomac datasets. 
The left of the dashed line denotes the historical context and the right the forecasting horizon. Per-model visualizations are provided in Figure~\ref{fig_appendix_all_models} and Figure~\ref{fig_appendix_all_models2}.
Although different TSFMs perform well on different datasets, they share common failure modes under evolving data distributions, including oversmoothing, delayed adaptation, and underestimated distribution shifts.}
\label{fig_case_study}
\end{figure*}

\noindent\textbf{Finding.} Rankings continue to evolve even over consecutive weekly snapshots. Rather than converging to a fixed ordering, the leading position alternates among Chronos-2, TiRex, TimesFM-2.5, and TabPFN-TS, while several mid-ranked models exchange positions across weeks. These observations suggest that no single model can dominated the live benchmark all the way, even though it is the best method in the static benchmark. This further highlights the necessity of the proposed \method and its supporting infrastructure for establishing a realistic testbed to constantly assess model performance under real-world deployment.


\subsection{Case study}
\label{sec_exp_case_study}
We visualize the predictions on four representative datasets from different domains and with diverse temporal characteristics: Wiki (daily Wikimedia page views), PM2.5 (hourly air quality), Wave (10-minute buoy wave height), and Potomac river flow (15-minute hydrology). 
Figure~\ref{fig_case_study} shows the visualization comparison of these TSFMs together with ARIMA, and we provide the per-model showcases in \appref{app_perdataset}~ Figure~\ref{fig_appendix_all_models} and ~\ref{fig_appendix_all_models2}.

As shown in Figure~\ref{fig_case_study}(a), on Wiki dataset, Moirai-2.0 and Chronos-2 can yield similar predictions with the ground truth throughout the forecasting horizon, whereas Toto-1.0 exhibits unstable oscillations with wider prediction intervals. 
Wave further reveals the complexity of long-term temporal dynamics. 
Chronos-2, TimesFM-2.5, and TiRex initially produce less fluctuating predictions, but their predictions gradually weaken and tend towards constant trajectories rather than maintaining the underlying periodicity. 
As shown in Figure \ref{fig_case_study}(b), although TiRex, TabPFN-TS, Chronos-Bolt and Sundial capture the overall upeard trend of PM2.5 in later stages, they consistently underestimate the sharp increase in the earlier stages, while ARIMA remaines anchored near historical levels and failed to predict the shift.
On the Potomac River dataset, the river flow drops rapidly after the predicted boundary, but all the highlighted methods react too slowly, consistently exceeding the true trajectory despite varying rates of decline.

\noindent\textbf{Findings.} 
Across these datasets, TSFM failures exhibit consistent patterns rather than isolated errors. Models smooth out abrupt PM~2.5 spikes, bias toward historical levels in Potomac, or collapse long-term forecasts into smoothed trajectories in Wave when periodic dynamics are under-observed. Crucially, these errors often occur simultaneously, indicating that prediction consistency does not imply reliability. Static benchmarks struggle to identify this behavior, merely averaging results over a fixed horizon. In contrast, live benchmarks iteratively evaluate TSFMs across evolving sources, revealing real-world performance degradation.

\section{Conclusion}
We presented a benchmark and live leaderboard for evaluating time series foundation models across diverse datasets and forecasting horizons. By standardizing data processing, evaluation protocols, this work aims to make comparisons more transparent and reproducible for the community. Our results highlight both the strengths of modern TSFMs and the remaining gaps in robustness and generalization when conditions shift across domains. Future work will expand the dataset coverage and tasks, and incorporate richer modalities to better reflect real-world deployment needs.

\clearpage


\par \noindent \textbf{Limitations and Ethical Considerations}. \method evaluates forecasting models using publicly accessible time series streams and is not designed to collect private or personally identifiable information; therefore, individual consent is generally not applicable to the current datasets. Data sources are reviewed for accessibility, licensing, and provenance, and future contributors are expected to exclude sensitive personal data. Nevertheless, geographic, domain, availability, and measurement biases in the selected streams may affect model scores and rankings, which should not be interpreted as evidence of universal superiority or downstream fairness. Finally, the reported forecasts and rankings are research artifacts rather than operational advice and should not be used directly for high-stakes financial, environmental, or public-safety decisions without domain-specific validation and human oversight.

\bibliographystyle{ACM-Reference-Format}
\bibliography{sample-base}

\clearpage
\appendix

\section{Getting Started}
We introduce how external participants can connect their TSFMs and contribute datasets to the live leaderboard. Because the benchmark updates in real time, each model must support sustainable repeated inference. Therefore, the leaderboard does not download model weights or execute user code. Instead, participants host a forecasting endpoint—such as a Hugging Face Space or Inference Endpoint—which the leaderboard calls via a standardized API. Participants manage inference resources, while the leaderboard handles data collection, evaluation, aggregation, and display.

\subsection{How to join the leaderboard}
\label{sec:join-leaderboard}

To join the leaderboard, participants provide a public Hugging Face model
repository, a public URL for the endpoint implementation, and a stable HTTPS
forecasting endpoint. Inference runs on participant-controlled infrastructure:
the model owner supplies the inference compute, while the leaderboard handles
live task generation, evaluation, aggregation, and display. A paid Hugging Face
Space, a participant-owned domain, and a public server IP are not required.

The recommended workflow is:

\begin{enumerate}
    \item Initialize the portable endpoint template, replace its
    \texttt{forecast\_one} function with the model inference logic, and start
    the service on the participant's inference server.
    \item Validate the local endpoint using a complete forecasting request.
    \item Run the publishing helper. By default, it exposes the local service
    through a persistent Tailscale Funnel with managed HTTPS and validates the
    resulting public route. A stable institutional HTTPS endpoint may be
    supplied instead.
    \item Submit the generated metadata and validation receipt after both the
    local and public endpoints pass validation.
\end{enumerate}

The default deployment requires Python~3, Docker Engine, and Docker Compose~v2.
When the Tailscale route is used, Tailscale~1.52 or later must also be installed
on the inference server. The complete setup is:

\begin{lstlisting}[style=bashbox]
git clone https://github.com/zhouziyu02/TS-Live.git
cd TS-Live

python3 -m venv .venv
source .venv/bin/activate
python -m pip install -r requirements.txt

export MODEL_ID="your-hf-username/your-model"
export DISPLAY_NAME="YourModelName"
export CODE_URL="https://github.com/your-org/your-endpoint"

# Verify the container runtime and Compose installation.
docker version
docker compose version

# Generate the portable forecasting service.
python scripts/community_model_wizard.py init \
  --output-dir forecast-service

# Replace forecast_one and add model-specific dependencies
# to forecast-service/requirements.txt. Publish this source
# directory at CODE_URL before submitting the model.
docker compose -f forecast-service/compose.yaml \
  up -d --build

# Check the container and wait for the local service.
docker compose -f forecast-service/compose.yaml ps
curl --fail --show-error \
  --retry 30 --retry-delay 2 --retry-connrefused \
  http://127.0.0.1:7860/health

# Send a complete local forecasting request.
python scripts/validate_external_model_endpoint.py \
  --endpoint-url http://127.0.0.1:7860/forecast \
  --model-id "${MODEL_ID}" \
  --allow-http

# One-time Linux setup for the default Tailscale route.
tailscale version
sudo tailscale up
sudo tailscale set --operator="${USER}"
tailscale status

# Validate locally, create the persistent HTTPS Funnel,
# validate the public route, and generate both artifacts.
python scripts/community_model_wizard.py publish \
  --model-id "${MODEL_ID}" \
  --display-name "${DISPLAY_NAME}" \
  --code-url "${CODE_URL}" \
  --output-dir community-submission
\end{lstlisting}

The first Funnel activation may require the participant or a tailnet
administrator to approve Funnel and HTTPS access in a browser. If the publishing
helper displays an approval URL, the participant completes this one-time
authorization and reruns the same \texttt{publish} command. The helper allows up
to ten minutes for the public DNS record and HTTPS route to become available.

Participants who already operate a stable institutional HTTPS endpoint, or a
named Cloudflare Tunnel with a stable hostname, may skip the Tailscale setup and
run:

\begin{lstlisting}[style=bashbox]
python scripts/community_model_wizard.py publish \
  --model-id "${MODEL_ID}" \
  --display-name "${DISPLAY_NAME}" \
  --code-url "${CODE_URL}" \
  --endpoint-url \
    https://forecast.your-domain.example/forecast \
  --output-dir community-submission
\end{lstlisting}

For each evaluation request, the endpoint receives only the causal target
history, an opaque series identifier, the forecast horizon, frequency, and
requested quantile levels. It does not receive future observations, ground
truth, raw dataset names, private metric values, or predictions from other
models. The validator checks \texttt{GET /health} and sends a complete request
to \texttt{POST /forecast}. It verifies the response length, numerical
finiteness, requested quantiles, response-size limit, and HTTPS requirement
before producing a successful receipt.

The default Funnel is initiated from the inference server and therefore
requires no inbound firewall rule, participant-owned domain, or manually
managed TLS certificate. Its \texttt{*.ts.net} address remains stable while the
Tailscale device identity and MagicDNS name are retained. Tailscale Funnel is a
low-friction option but remains a beta service with provider-defined bandwidth
limits. Participants requiring a custom domain or stronger ingress guarantees
should use an institutional HTTPS service or a named Cloudflare Tunnel.
Temporary tunnel URLs are not accepted.

After successful validation, the participant pastes the contents of
\texttt{community-submission/community\_model.yaml} and
\texttt{community-submission/validator\_receipt.json} into the community model
request form at
\url{https://github.com/zhouziyu02/TS-Live/issues/new?template=community-model.yml}.
The submitted endpoint is reviewed before being enabled.

An accepted model is admitted only to future live evaluation rounds and is
never backfilled on releases preceding its admission time. Its results appear
after the next successful evaluation cycle. Detailed release snapshots may be
retained privately for auditing and metric recomputation, while only metric
summaries and aggregate leaderboard tables are published.

\subsection{How to contribute new dataset}

As an open online benchmark, \method accepts new \emph{public} time series streams from the community.
A contribution plugs into the same live loop as the built-in sources in Sec.~\ref{sec:streaming-data}: register the dataset semantics, ingest fresh observations on a recurring schedule, and let the evaluation house form and score forecasting tasks automatically.
Contributors provide (i)~a dataset-level specification that defines the forecasting problem and (ii)~an ingestion adapter that fetches the public source and emits normalized observations.
The full attribute contract and preprocessing requirements are given in Appx.~\ref{appx:live-data}; accepted sources are reviewed for licensing, stability, and schema compliance before entering the public registry.

\section{Implementation Details of \method}  \label{appendix:implementation_details}

\subsection{Model Entrance.}
\label{appx:model-entrance}

The model entrance is the component through which all forecasting methods are registered, adapted, and admitted into \method. Its primary role is to decouple model-specific inference logic from the live data and evaluation pipeline. A model may be a hosted TSFM, an external service, a local predictor, or a statistical baseline; after passing through the entrance, however, all methods expose a uniform forecasting contract to the evaluation house.

\noindent\textbf{Model registry.}
Each model is specified by a registry entry containing a stable model identifier,
a display name, the model type, organization metadata, links to the model and
replication code, and the information needed to instantiate its adapter. For
hosted TSFMs, this includes the remote model identifier and API backend. For
community submissions, the entry may instead specify a validated HTTPS forecast
endpoint. For statistical baselines, the entry records the algorithm family and
its fixed hyperparameters, such as the ARIMA order, moving-average window, or
season length. The registry also records whether a model is enabled for public
evaluation and the time from which its future-only evaluation window begins.
This start time is important: \method never backfills a newly added model on
targets that were already observable before the model entered the benchmark.

\noindent\textbf{Unified forecasting contract.}
For every admitted evaluation task, the entrance presents the model with a
causal context window, the sampling frequency, the prediction horizon, and
minimal task metadata. Formally, the model receives
$\mathbf{x}_{1:T}$, a frequency descriptor $f$, and a horizon $H$, and is asked
to return forecasts for $\mathbf{x}_{T+1:T+H}$. The standardized output contains
a point forecast and, when available, predictive quantiles:
\[
    \left(
    \hat{\boldsymbol{\mu}}_{1:H},
    \{\hat{\mathbf{q}}_{\tau,1:H}\}_{\tau \in \mathcal{Q}}
    \right),
\]
where $\mathcal{Q}$ denotes the requested quantile levels. This contract is
implemented through a common predictor abstraction, so the downstream evaluator
does not need method-specific logic for TSFM APIs, local predictors, endpoints,
or baselines. Forecast arrays are checked for shape, horizon length, and finite
numeric values before they are converted into the common evaluation format.

\noindent\textbf{Adapters for heterogeneous model outputs.}
Different TSFMs expose various native output formats. Some return means and quantile forecasts; others return samples, prediction intervals, or point forecasts. The model entrance normalizes these outputs into a uniform mean-plus-quantiles representation. If a model provides samples, empirical quantiles are computed from the sample paths. If a model exposes only point forecasts, the adapter constructs an approximate predictive distribution via repeated stochastic calls or residual-bootstrap perturbations. This normalization allows point and distributional metrics, such as the quantile-loss approximation to CRPS, to be computed by a single metric engine.

\noindent\textbf{Hosted and external models.}
Built-in hosted TSFMs are queried through authenticated forecast APIs with
fixed request fields: historical target values, frequency, prediction length,
and requested quantile levels. Community models can be integrated through a
self-hosted HTTPS endpoint. In this mode, \method treats the submitted model as
a black box: it does not import user code, download user weights, or execute
third-party dependencies inside the leaderboard process. By default, the
endpoint receives only the causal context, an opaque series identifier, the
frequency, the horizon, and the requested quantiles. Future targets, metric
values, authentication tokens, raw leaderboard internals, and other models'
predictions are never sent to the endpoint. The adapter enforces request
timeouts, retry limits, maximum context lengths, maximum response sizes, and
forecast-validity checks.

\noindent\textbf{Statistical baselines.}
The same entrance also hosts non-pretrained reference methods, including
seasonal naive, moving average, ARIMA, and exponential smoothing. These
baselines are fit or instantiated only from the context window of the current
task and therefore obey the same causal restriction as TSFMs. Since several
classical methods naturally produce point forecasts, the entrance derives
quantile forecasts using residual-bootstrap samples when probabilistic outputs
are required. This makes the baselines compatible with both point-accuracy and
distributional metrics without giving them any special treatment in the
evaluator.

\paragraph{Failure handling and auditability.}
The entrance isolates model failures from the rest of the benchmark. Transient
API failures can be retried, while invalid responses, non-finite forecasts,
short horizons, malformed quantiles, or endpoint errors are rejected before
scoring. For each successful run, \method stores model metadata, task metadata,
metric rows, and evaluation timestamps as persistent artifacts. These records
make it possible to audit which adapter, model identifier, context window,
horizon, and evaluation time produced each leaderboard entry. Thus, the model
entrance provides both a practical integration layer for heterogeneous
forecasting systems and a reproducibility boundary that keeps the live
evaluation protocol model agnostic.



\subsection{Live Data.} \label{appx:live-data}

As a public online benchmark, \method is designed to grow beyond its initial registry.
Researchers and practitioners can contribute additional \emph{public} streams---for example, a weather station feed, an open mobility API, or a government statistics portal---as long as the source is openly accessible and can be mapped into the shared contract below.
Each contribution reuses the same live loop as the built-in sources: raw responses are archived for provenance, observations are normalized into a common schema, forecast tasks are generated from per-dataset window settings, and the evaluation house scores only targets that become observable after a model joins the leaderboard.

\noindent\textbf{Contribution contract.}
A new source requires two complementary pieces.
First, a \emph{dataset specification} that semantically defines the forecasting problem (domain, entity type, native frequency, evaluation frequency, context and horizon lengths, targets, and covariates).
Second, an \emph{ingestion adapter} that periodically retrieves the public endpoint, preserves immutable raw evidence, and emits normalized observation records.
The dataset specification is the single source of semantic truth; polling cadence is an operational scheduling choice and is kept separate from the forecasting definition itself.

\noindent\textbf{Required dataset attributes.}
Each contributed dataset must declare the attributes below.
\begin{itemize}[leftmargin=*, nosep, topsep=2pt]
  \item \textbf{Identity:} dataset identifier, source identifier, human-readable name, domain, entity granularity, and short English description/background for downstream users.
  \item \textbf{Forecasting setup:} native \texttt{data\_frequency}, recommended \texttt{eval\_frequency}, \texttt{history\_length\_steps} and \texttt{forecast\_horizon\_steps} (both measured in native-frequency steps), \texttt{target\_variables}, and optional \texttt{covariate\_variables}.
  \item \textbf{Behavior flags:} whether the series is ready for direct zero-shot leaderboard scoring, and whether raw events must be aggregated to a regular grid before forecasting.
\end{itemize}
History and horizon windows are always expressed in the native series frequency, so a high-rate stream and a monthly macro series can coexist without forcing a single global window length.

\noindent\textbf{Storage model.}
The data layer separates three concerns rather than relying on one monolithic training file.
\begin{itemize}[leftmargin=*, nosep, topsep=2pt]
  \item \textbf{Raw archive:} immutable copies of public responses, together with request metadata, fetch time, content hash, and parser version, so every normalized value remains auditable.
  \item \textbf{Canonical store:} relational long-format tables for datasets, entities, variables, observations, forecast tasks, and quality/run logs. This layer is the source of truth.
  \item \textbf{Task export:} ephemeral model-facing bundles that pair a context window with dataset metadata; future targets are exported separately for the evaluator only.
\end{itemize}

\noindent\textbf{Preprocessing requirements.}
After each fetch, an ingestion adapter must satisfy the following requirements.
\begin{enumerate}[leftmargin=*, nosep, topsep=2pt]
  \item \textbf{Preserve provenance.} Store the untouched response before parsing; never overwrite prior raw evidence.
  \item \textbf{Normalize entities and variables.} Map each forecastable object (station, market, fleet, grid point, country, etc.) to a stable entity identifier, and map measurable quantities to variable identifiers with units and frequency hints.
  \item \textbf{Emit long-format observations.} Represent each value as one record keyed by entity, timestamp, and variable. Every record must carry:
  \begin{itemize}[leftmargin=*, nosep, topsep=1pt]
    \item \texttt{timestamp}: when the measurement refers to;
    \item \texttt{available\_time}: when the value became knowable to a forecaster;
    \item \texttt{ingest\_time}: when \method retrieved it;
    \item numeric \texttt{value}, \texttt{frequency}, \texttt{unit}, and a link back to the raw evidence.
  \end{itemize}
  Event streams (e.g., news documents or earthquake catalogs) must be bucketed to the declared native frequency when aggregation is required.
  \item \textbf{Generate forecast tasks.} Instantiate tasks from the dataset specification: forecast issue time, context window, horizon window, covariate set, and frequency must be identical for all models evaluated on the same dataset.
  \item \textbf{Report data health.} Record missingness, duplicates, freshness delay, outliers, and ingestion failures so unstable or stale sources do not silently enter the leaderboard.
\end{enumerate}

\noindent\textbf{Leakage rule.}
Model inputs must never contain information that was unavailable at prediction time.
For every context observation,
\[
\texttt{available\_time} \leq \texttt{forecast\_issue\_time}.
\]
A measurement timestamp alone is therefore insufficient: the value must also have been publicly knowable before the forecast was issued.
Future targets are kept evaluator-only and are never exposed as model inputs.
This rule is the same leakage-resistant contract used by the built-in streams described in Appx.~\ref{app:data-process}. Once accepted, a contributed source inherits the same task generation, future-only gating, and leaderboard update loop as the core registry entries.

\subsection{Evaluation House.}
\label{appx:evaluation-house}

The evaluation house is the component in \method that turns standardized model
forecasts into auditable leaderboard records. It receives forecasting tasks from
the live data stage and model predictions from the model entrance, then applies
a fixed evaluation protocol to all admitted model--task pairs. Its design goal
is to ensure that every reported score is computed from the same target window,
the same metric implementation, and the same aggregation rule, regardless of
the model's inference backend or output format.

\noindent\textbf{Task admission and future-only scoring.}
For each refreshed data stream, the live data stage materializes a forecasting
task consisting of a historical context window $\mathbf{x}_{1:T}$, a prediction
horizon $H$, dataset metadata, and the future target
$\mathbf{x}_{T+1:T+H}$. The evaluation house first checks whether the task is
eligible for a given model. Let $a_m$ denote the accepted entrance time of model
$m$. A task is scored for $m$ only if its target window is generated from
observations that become available after $a_m$. This future-only gate prevents
newly submitted models from being evaluated retrospectively on data that could
have been inspected during model development or endpoint debugging. The same
gate is applied to hosted TSFMs, external community endpoints, local wrappers,
and statistical baselines.

\noindent\textbf{Forecast execution.}
After admission, the evaluation house dispatches the task through the unified
predictor interface. The model receives only the causal context, frequency, and
horizon; the target values are held out until scoring. The returned forecast is
converted into a common representation containing a mean or median forecast and,
when available, a set of predictive quantiles. The evaluator validates that all
forecast arrays have the required horizon length and contain finite numeric
values. Invalid, malformed, or incomplete outputs are rejected before metrics
are computed, so downstream ranking is never based on partially parsed
forecasts.

\noindent\textbf{Metric computation.}
The metric engine evaluates both point accuracy and probabilistic quality. For
point forecasts, \method reports metrics such as MSE, RMSE, MAE, MASE, MAPE,
sMAPE, NRMSE, and ND when they are well defined. For probabilistic forecasts,
the evaluator computes interval and quantile-based scores, including MSIS and
the mean weighted sum quantile loss, which is used as the CRPS-style
distributional metric in the leaderboard. All metrics are computed on the same
held-out target window for all models admitted to that task. Metrics that are
undefined for a particular target, such as percentage errors near zero, are
marked as unavailable rather than silently imputed.

\noindent\textbf{Reference baselines and relative gain.}
The evaluation house also maintains matched baseline scores for lightweight
reference methods. In particular, Seasonal-Naive is used as a causal reference
because it requires no pretraining and can be instantiated from the same
context window as every other method. For live aggregate reporting, \method
computes a relative real-time gain (RTG) from matched MSE values:
\[
    \mathrm{RTG}(m)
    =
    100 \cdot
    \frac{\mathrm{MSE}_{\mathrm{SNaive}} - \mathrm{MSE}_{m}}
         {\mathrm{MSE}_{\mathrm{SNaive}} + \mathrm{MSE}_{m}} .
\]
This bounded form gives positive values to models that improve over the
seasonal-naive reference and negative values to models that underperform it,
while avoiding instability when absolute errors are small.

\textbf{Aggregation and ranking.}
The evaluation house produces fine-grained and aggregate views. At the lowest level, it stores a metric row for each model--dataset--release combination. Dataset-level scores are obtained by averaging over releases from the same stream. Overall live scores assign equal weight to datasets, preventing frequently refreshed streams from dominating. For grouped analysis, \method reports GIFT-Eval-style aggregates by domain, frequency, and prediction length. These grouped tables normalize MSE and CRPS against matched Seasonal-Naive scores, then aggregate across configurations.

Ranks are computed from matched comparisons rather than from incomparable
partial records. For compact overall presentation, \method reports average
rank, pairwise win rate, and an Elo-style score based on shared releases. The
rank computation uses the lower-is-better ordering of MSE and CRPS and only
compares models on tasks where both models have valid scores. This preserves a
fair comparison when some models join later or when an admitted forecast fails
validation for a particular release.

\noindent\textbf{Persistence and reproducibility.}
Every successful evaluation produces persistent artifacts: the model name and
identifier, dataset configuration, release timestamp, prediction horizon,
metric values, and aggregation metadata. These artifacts are stored separately
from the model execution code and are sufficient to reconstruct the public
leaderboard tables. The system also records metadata for aggregate tables,
including generation time, the normalization baseline, the aggregation rule,
and the number of contributing models and configurations. This separation
allows \method to audit individual leaderboard entries, regenerate aggregate
tables, and distinguish changes caused by new data from changes caused by
model or code updates.

\noindent\textbf{Robustness of the online loop.}
Because \method operates continuously, the evaluation house is designed to
handle partial failures without interrupting the full benchmark. A failed model
call, invalid endpoint response, missing quantile, or undefined metric affects
only the corresponding model--task pair. Other models evaluated on the same
task remain valid, and later releases can still contribute new evidence for the
failed model. As new observations arrive, the same admission, inference,
scoring, persistence, and aggregation steps are repeated. The evaluation house
therefore serves as the reproducible boundary between live forecasting
execution and public leaderboard publication.





\subsection{Evaluation Metrics}
\label{appx:exp_metrics}
We evaluate each models along two complementary perspectives. Basic metrics quantify the absolute quality and cost of a forecast on each dataset, while performance-rank metrics aggregate these per-dataset scores into a single comparable measure of relative standing across the whole benchmark. Throughout we let $\mathbf{y}_t \in \mathbb{R}^{C}$ denote the ground-truth vector at forecast step $t$ and $\hat{\mathbf{y}}_t \in \mathbb{R}^{C}$ the corresponding prediction over a horizon of length $H$, where $C$ is the forecast dimension and $\|\cdot\|$ denotes a vector norm.

\textbf{MSE and RMSE.} The mean squared error (MSE) and root mean squared error (RMSE) are \textit{lower-is-better} point-accuracy metrics. MSE penalizes large deviations quadratically and RMSE reports the same quantity in the original data scale,
\begin{equation}
\mathrm{MSE} = \frac{1}{C H}\sum_{t=1}^{H}\bigl\|\mathbf{y}_t-\hat{\mathbf{y}}_t\bigr\|_2^{2},
\qquad
\mathrm{RMSE} = \sqrt{\mathrm{MSE}} .
\end{equation}
We compute both on $z$-normalized series so that datasets with different magnitudes contribute comparably.

\textbf{MAPE.} The mean absolute percentage error (MAPE) is a \emph{lower-is-better} metric that expresses error relative to the magnitude of each target, which makes it scale-free and comparable across datasets,
\begin{equation}
\mathrm{MAPE} = \frac{1}{C H}\sum_{t=1}^{H}\bigl\|(\mathbf{y}_t-\hat{\mathbf{y}}_t) / \mathbf{y}_t\bigr\|_1 ,
\end{equation}
Because MAPE is undefined when a target is zero and inflates near small targets, we report it on series whose values stay bounded away from zero and rely on MSE and RMSE elsewhere.

\textbf{CRPS.} The continuous ranked probability score (CRPS) is a \emph{lower-is-better} metric that assesses probabilistic forecasts. For a single coordinate with predicted cumulative distribution $F$ and realized value $y$ it is
\begin{equation}
\mathrm{CRPS}(F, y) = \int_{-\infty}^{\infty}\bigl(F(z) - \mathbf{1}\{z \ge y\}\bigr)^{2}\,\mathrm{d}z ,
\end{equation}
and we average it over the $C$ coordinates of $\hat{\mathbf{y}}_t$ and over the $H$ forecast steps. CRPS is a strictly proper scoring rule that jointly rewards calibration and sharpness, and it reduces to absolute error when the forecast is a point mass. We estimate it from the predicted quantiles emitted by each model.



\textbf{Average Rank.} This \emph{lower-is-better} metric ranks models on each dataset by a basic metric and averages these ranks across datasets, providing a simple scale-free indicator of consistent standing.

\textbf{Win Rate.} This \emph{higher-is-better} metric is the fraction of pairwise comparisons a model wins. For each dataset and each opposing model it scores a win when its metric is better, and we report the proportion of wins over all such comparisons.

\textbf{Elo.} Elo rating~\citep{elo1978rating} is a \emph{higher-is-better} metric that treats per-dataset head-to-head outcomes as matches and fits a rating $R_m$ to each model, where expected score of model $m$ against model $n$ is
\begin{equation}
\mathbb{E}[S_{m,n}] = \frac{1}{1 + 10^{(R_n - R_m)/400}} .
\end{equation}
Ratings are updated from observed wins and losses, so Elo rewards beating strong competitors more than weak ones and yields a single rating robust to the inclusion or removal of individual models.





\section{Dataset details}
\label{app:dataset-details}

The main paper summarizes benchmark coverage and task configuration. This appendix provides additional background on series origins, physical or social quantities represented, and public acquisition methods. All sources are openly accessible without proprietary API keys. Table~\ref{tab:app-dataset-background} documents each dataset's monitoring entity, forecasting target, public endpoint, and background notes from our diversity verification samples.

\subsection{Data pipeline}
\label{app:data-process}

\textbf{Collection.}
At each collection round, the pipeline loads the registry, queries the corresponding public endpoints, and stores the response body without modification under a source-specific raw directory.
Collection runs on a source-dependent schedule, from frequent polls for high-rate markets to daily polls for environmental and macroeconomic feeds.
Each round appends new raw response records rather than overwriting prior responses; when sufficient observations arrive, the pipeline exports fresh forecasting tasks whose future windows can be scored against newly observed ground truth.
The raw layer keeps JSON, TXT, or ZIP responses together with request URL, request parameters, HTTP status, fetch time, content hash, parser version, and error messages when applicable.
This raw response archive makes every benchmark observation traceable to the exact public response from which it was parsed.

\textbf{Canonical parsing.}
Successful raw responses are parsed into a small relational schema rather than a single monolithic CSV.
The metadata tables store dataset definitions, entities, variables, raw response records, quality reports, and run logs.
The central table is a long-format observation table:
\begin{quote}
\small
\texttt{dataset\_id, source\_id, entity\_id, variable\_id, timestamp,}\\
\texttt{available\_time, ingest\_time, value, frequency, unit, raw\_id}
\end{quote}
This design preserves provenance and supports heterogeneous sources with different entities, units, and update mechanisms.
For event sources, such as GDELT and USGS Earthquake, raw events are converted into regular aggregate time series, e.g., document volume or earthquake counts per time bucket.

\textbf{Task export.}
Forecasting tasks are generated from the canonical observation table using the history and horizon lengths specified in the registry.
Each task records a target entity, target variable, forecast issue time, context window, horizon window, covariate list, and frequency.
The model-facing export contains \texttt{task.json} and \texttt{context.csv}; labels are isolated in \texttt{future\_target.csv} for the evaluator.
To prevent look-ahead leakage, the exporter enforces
\[
\texttt{available\_time} \leq \texttt{forecast\_issue\_time}
\]
for every context row.
Thus, a timestamp is insufficient for inclusion: values must also be available prior to prediction issuance.

\subsection{Per-dataset background and access}
\label{app:per-dataset}

\begin{table*}[t]
\centering
\scriptsize
\setlength{\tabcolsep}{2pt}
\caption{Background and public access information for each benchmark dataset.
Endpoints are the base URLs used for collection; query parameters depend on
entity, time window, and variables. Covariates may be omitted in univariate
runs. We use the following abbreviations in subsequent tables and figures:
BTC for Binance BTCUSDT, PM2.5 for Open-Meteo air quality, Quake for USGS
earthquake aggregates, Potomac for USGS Potomac discharge, Water for NOAA
CO-OPS water level, Wave for NOAA NDBC buoy observations, T2M for NASA POWER
meteorology, KSFO for NWS KSFO observations, Temp2m for Open-Meteo weather,
and Wiki for Wikimedia pageviews.}
\label{tab:app-dataset-background}
\resizebox{\textwidth}{!}{%
\begin{tabular}{@{}llp{2.1cm}p{1.9cm}p{3.4cm}p{2.5cm}@{}}
\toprule
\textbf{Dataset} & \textbf{Domain} & \textbf{Entity \& region} & \textbf{Primary target} & \textbf{Public endpoint} & \textbf{Background notes}\\
\midrule
Open-Meteo Shanghai weather & weather & Shanghai grid (31.23$^\circ$N, 121.47$^\circ$E) & 2\,m temperature & \url{https://api.open-meteo.com/v1/forecast} & Smooth diurnal weather; humidity, wind, precipitation covariates\\
Open-Meteo Shanghai air quality & air quality & Shanghai grid & PM$_{2.5}$ & \url{https://air-quality-api.open-meteo.com/v1/air-quality} & Pollutant spikes; PM$_{10}$, NO$_2$, CO covariates\\
NASA POWER Shanghai meteorology & weather-energy & Shanghai point & Air temperature (T2M) & \url{https://power.larc.nasa.gov/api/temporal/hourly/point} & Energy-oriented meteorology; evaluated daily despite hourly native series\\
USGS Potomac discharge & hydrology & USGS site 01646500, Potomac River & River discharge & \url{https://waterservices.usgs.gov/nwis/iv/} & High-frequency hydrology; gauge height covariate\\
NOAA CO-OPS San Francisco water level & ocean & Tide station 9414290, San Francisco & Water level & \url{https://api.tidesandcurrents.noaa.gov/api/prod/datagetter} & Six-minute coastal measurements with tide-like oscillation\\
NOAA NDBC buoy 46013 & ocean & NDBC buoy 46013 & Significant wave height & \url{https://www.ndbc.noaa.gov/data/realtime2/46013.txt} & Marine buoy observations; wind, pressure, temperature covariates\\
GBFS Citi Bike station status & traffic & NYC bike-share stations & Available bikes per station & \url{https://gbfs.citibikenyc.com/gbfs/en/station_status.json} & Shared-mobility supply; verification uses a system mean over sampled stations\\
Binance BTCUSDT (hourly) & finance & BTCUSDT spot market & Close price & \url{https://api.binance.com/api/v3/klines} & Hourly OHLCV with regime shifts\\
Binance BTCUSDT (one-second) & finance & BTCUSDT spot market & Close price & \url{https://api.binance.com/api/v3/klines} & Second-level market microstructure stress test\\
Binance BTCUSDT (monthly) & finance & BTCUSDT spot market & Close price & \url{https://api.binance.com/api/v3/klines} & Low-frequency crypto history on a shared asset\\
CoinGecko Bitcoin market chart & finance & Bitcoin (market level) & USD price & \url{https://api.coingecko.com/api/v3/coins/bitcoin/market_chart} & Market-cap and volume covariates\\
Wikimedia Time Series article views & web attention & enwiki article \emph{Time series} & Daily pageviews & \url{https://wikimedia.org/api/rest_v1/metrics/pageviews/per-article/} & Sparse daily attention signal\\
NWS KSFO observations & weather & NWS station KSFO, San Francisco & Temperature & \url{https://api.weather.gov/stations/KSFO/observations} & Official station observations; wind, humidity, pressure covariates\\
NOAA NCEI NYC daily summaries & weather & NCEI station USW00014732, New York City & Daily mean temperature & \url{https://www.ncei.noaa.gov/access/services/data/v1} & Long-running official daily climate summaries\\
World Bank China macro indicators & macro-economy & China (country level) & GDP (current US\$) & \url{https://api.worldbank.org/v2/country/CHN/indicator/} & Annual macro series since 1961; population and inflation covariates\\
GDELT climate-change timeline & news events & Global query: ``climate change'' & Document volume (15\,min) & \url{https://api.gdeltproject.org/api/v2/doc/doc} & Event-attention stream; scored at daily frequency\\
USGS global earthquake aggregates & disaster events & Global earthquake catalog & Hourly earthquake count & \url{https://earthquake.usgs.gov/earthquakes/feed/v1.0/summary/all_week.geojson} & Sparse events bucketed by UTC hour\\
\bottomrule
\end{tabular}%
}
\end{table*}

\subsection{Event-derived series}
\label{app:event-aggregation}

Two datasets arrive as \emph{events} rather than natively regular measurements.
\textbf{GDELT} returns a timeline of relative news-document volume for the query ``climate change'' at 15-minute resolution.
\textbf{USGS earthquakes} publishes a rolling one-week GeoJSON feed of global events.
For both sources, we aggregate timestamped events into regular time buckets before forming forecasting tasks, so that event-derived streams follow the same leaderboard protocol as directly reported time series.

\subsection{Diversity verification samples}
\label{app:verification}

Before live deployment we collected short public samples from every provider to confirm that each source can be downloaded and converted into a regular numeric series.
The verification covered 17 datasets, 11 domains, and 2{,}672 representative observations in total; every source in Table~\ref{tab:app-dataset-background} formed a usable time series in this check.
Table~\ref{tab:dataset-inventory} reports the resulting per-dataset registry windows and descriptive statistics; those numbers are \emph{verification slices}, not fixed train/test sizes for the live leaderboard.

These statistics are generated from the canonical observation rows produced by the data pipeline, not from manual measurements.
For each dataset, the diversity-report script selects the configured target variable (or a documented display variable for visualization), chooses one representative entity or aggregate, deduplicates observations by dataset, entity, variable, and timestamp, and then computes $N$, mean, range, and standard deviation on the selected target values.
The underlying verification slices come from source-dependent sample collection runs rather than from one identical calendar period imposed on every domain-frequency pair: high-rate feeds use short recent windows, backfillable daily or hourly sources use multi-day verification windows, and monthly or annual sources use their available historical records.
This design lets the verification check whether each source is parseable and behaviorally distinct, while leaving the live benchmark free to keep collecting future observations.

\subsection{Dataset inventory and verification statistics}
\label{app:dataset-inventory}

\begin{table*}[t]
\centering
\scriptsize
\setlength{\tabcolsep}{2.5pt}
\caption{Per-dataset registry and verification statistics. \emph{Data freq.} and \emph{Eval freq.} are native series frequency and recommended scoring frequency. Hist.\ and Hor.\ are history and forecast horizon in native-frequency steps. Type distinguishes directly reported time series from event-derived time series. $N$ is \texttt{row\_count} in the diversity verification slice (not a fixed train/test size). Mean, Std, and Range summarize the selected target variable in that slice.}
\label{tab:dataset-inventory}
\resizebox{\textwidth}{!}{%
\begin{tabular}{@{}lllll rr l r rrr@{}}
\toprule
\textbf{Dataset} & \textbf{Domain} & \textbf{Data freq.} & \textbf{Eval freq.} & \textbf{Target} & \textbf{Hist.} & \textbf{Hor.} & \textbf{Type} & \textbf{$N$} & \textbf{Mean} & \textbf{Std} & \textbf{Range}\\
\midrule
Open-Meteo weather & weather & 1h & 1h & temperature\_2m & 336 & 24 & direct & 32 & 26.97 & 2.746 & 24.0--33.7\\
Open-Meteo air quality & air\_quality & 1h & 1h & pm2\_5 & 336 & 24 & direct & 32 & 50.68 & 22.38 & 27.1--99.4\\
NASA POWER meteorology & weather\_energy & 1h & 1d & T2M & 336 & 24 & direct & 112 & 23.62 & 3.679 & 17.4--30.9\\
USGS Potomac water & hydrology & 15min & 1h & usgs\_00060 & 96 & 24 & direct & 275 & 9{,}824 & 652.9 & 8{,}770--11{,}100\\
NOAA CO-OPS water level & ocean & 6min & 1h & water\_level & 240 & 60 & direct & 240 & 1.012 & 0.558 & $-$0.042--1.84\\
NOAA NDBC buoy & ocean & 10min & 1h & wave\_height & 144 & 72 & direct & 76 & 2.034 & 0.187 & 1.8--2.4\\
GBFS Citi Bike status & traffic & 15min & 15min & num\_bikes\_available & 96 & 4 & direct & 6 & 9.467 & 0.462 & 9.26--10.5\\
Binance BTCUSDT hourly & finance & 1h & 1h & close & 96 & 24 & direct & 24 & 65{,}330 & 1{,}458 & 62{,}178--67{,}295\\
Binance BTCUSDT one-second & finance & 1s & 1s & close & 900 & 60 & direct & 999 & 61{,}667 & 50.09 & 61{,}550--61{,}772\\
Binance BTCUSDT monthly & finance & 1mo & 1mo & close & 60 & 12 & direct & 97 & 40{,}957 & 31{,}981 & 3{,}434--115{,}764\\
CoinGecko Bitcoin market & finance & 1h & 1h & price\_usd & 96 & 24 & direct & 289 & 65{,}335 & 1{,}397 & 61{,}557--67{,}327\\
Wikimedia pageviews & web\_attention & 1d & 1d & pageviews & 30 & 7 & direct & 6 & 277.2 & 58.13 & 187--339\\
NWS KSFO observations & weather & 1h & 1h & temperature & 96 & 24 & direct & 313 & 15.67 & 2.999 & 12.0--21.1\\
NOAA NCEI daily summaries & weather & 1d & 1d & daily\_avg\_temperature & 30 & 7 & direct & 4 & 24.20 & 1.602 & 21.7--25.6\\
World Bank China macro & macro\_economy & 1y & 1y & gdp\_current\_usd & 40 & 5 & direct & 64 & 3.65e12 & 5.63e12 & 4.73e10--1.87e13\\
GDELT climate timeline & news\_events & 15min & 1d & document\_volume & 672 & 96 & event-derived & 79 & 1.249 & 0.348 & 0.701--2.347\\
USGS earthquake aggregates & disaster\_events & 1h & 1d & earthquake\_count & 168 & 24 & event-derived & 24 & 10.71 & 4.067 & 4--23\\
\bottomrule
\end{tabular}%
}
\end{table*}

Table~\ref{tab:verification-patterns} summarizes the behavioral patterns seen in the verification charts.
They illustrate why the registry mixes smooth environmental signals, volatile financial series, sparse attention counts, and event-driven streams under one benchmark.

\begin{table}[H]
\centering
\small
\caption{Representative dynamics observed in diversity verification samples (June 2026).}
\label{tab:verification-patterns}
\resizebox{\linewidth}{!}{%
\begin{tabular}{@{}lp{6.2cm}@{}}
\toprule
\textbf{Domain group} & \textbf{Observed pattern in verification samples}\\
\midrule
Weather \& air quality & Diurnal structure (Open-Meteo, NWS) and pollutant spikes (PM$_{2.5}$)\\
Weather-energy & Smooth exogenous meteorology suitable as covariate-rich context\\
Hydrology \& ocean & Slower physical dynamics; sub-hourly coastal tides and buoy waves\\
Urban mobility & System-level bike availability rises/falls with commuting demand\\
Finance & Heavy-tailed movement from 1\,s to monthly BTC series; CoinGecko adds market-level context\\
Web \& macro & Sparse daily pageviews and slow annual GDP growth with long history\\
News \& disaster events & Bursty document volume and hourly earthquake counts from sparse events\\
\bottomrule
\end{tabular}}
\end{table}

\begin{table*}[t]
\centering
\scriptsize
\setlength{\tabcolsep}{2.5pt}
\caption{Per-dataset verification variability used to support the dynamic-richness criterion in Figure~\ref{fig:dataset-diversity}. CV is computed on the selected target variable in the verification slice as $\mathrm{std}/|\mathrm{mean}|$. These short slices illustrate temporal behavior but are not fixed train/test windows.}
\label{tab:app-verification-variability}
\resizebox{\textwidth}{!}{%
\begin{tabular}{@{}llrrp{7.0cm}@{}}
\toprule
\textbf{Dataset} & \textbf{Domain} & \textbf{$N$} & \textbf{CV} & \textbf{Observed dynamic pattern}\\
\midrule
Open-Meteo weather & weather & 32 & 0.102 & Smooth hourly diurnal temperature cycle\\
Open-Meteo air quality & air\_quality & 32 & 0.442 & Pollutant variation with short spikes\\
NASA POWER meteorology & weather\_energy & 112 & 0.156 & Smooth exogenous meteorology over several days\\
USGS Potomac water & hydrology & 275 & 0.066 & High-frequency river discharge with slow physical variation\\
NOAA CO-OPS water level & ocean & 240 & 0.551 & Sub-hourly tidal oscillation with values around zero\\
NOAA NDBC buoy & ocean & 76 & 0.092 & Marine wave-height fluctuations with meteorological covariates\\
GBFS Citi Bike status & traffic & 6 & 0.049 & Short mobility-supply sequence averaged over sampled stations\\
Binance BTCUSDT hourly & finance & 24 & 0.022 & Intraday financial movement\\
Binance BTCUSDT one-second & finance & 999 & $8.1{\times}10^{-4}$ & Second-level market microstructure with small relative variation over the sampled window\\
Binance BTCUSDT monthly & finance & 97 & 0.781 & Long-horizon crypto regime shifts\\
CoinGecko Bitcoin market & finance & 289 & 0.021 & Market-level hourly Bitcoin price movement\\
Wikimedia pageviews & web\_attention & 6 & 0.210 & Sparse daily public-attention counts\\
NWS KSFO observations & weather & 313 & 0.191 & Station weather with diurnal structure and irregular updates\\
NOAA NCEI daily summaries & weather & 4 & 0.066 & Daily official climate summaries in a short verification slice\\
World Bank China macro & macro\_economy & 64 & 1.545 & Long-term annual macroeconomic trend\\
GDELT climate timeline & news\_events & 79 & 0.279 & Bursty event-attention volume aggregated from documents\\
USGS earthquake aggregates & disaster\_events & 24 & 0.380 & Sparse hourly disaster-event counts\\
\bottomrule
\end{tabular}}
\end{table*}

\subsection{Per-dataset verification charts}
\label{app:verification-charts}

\begin{figure*}[t]
    \centering
    \includegraphics[width=\linewidth]{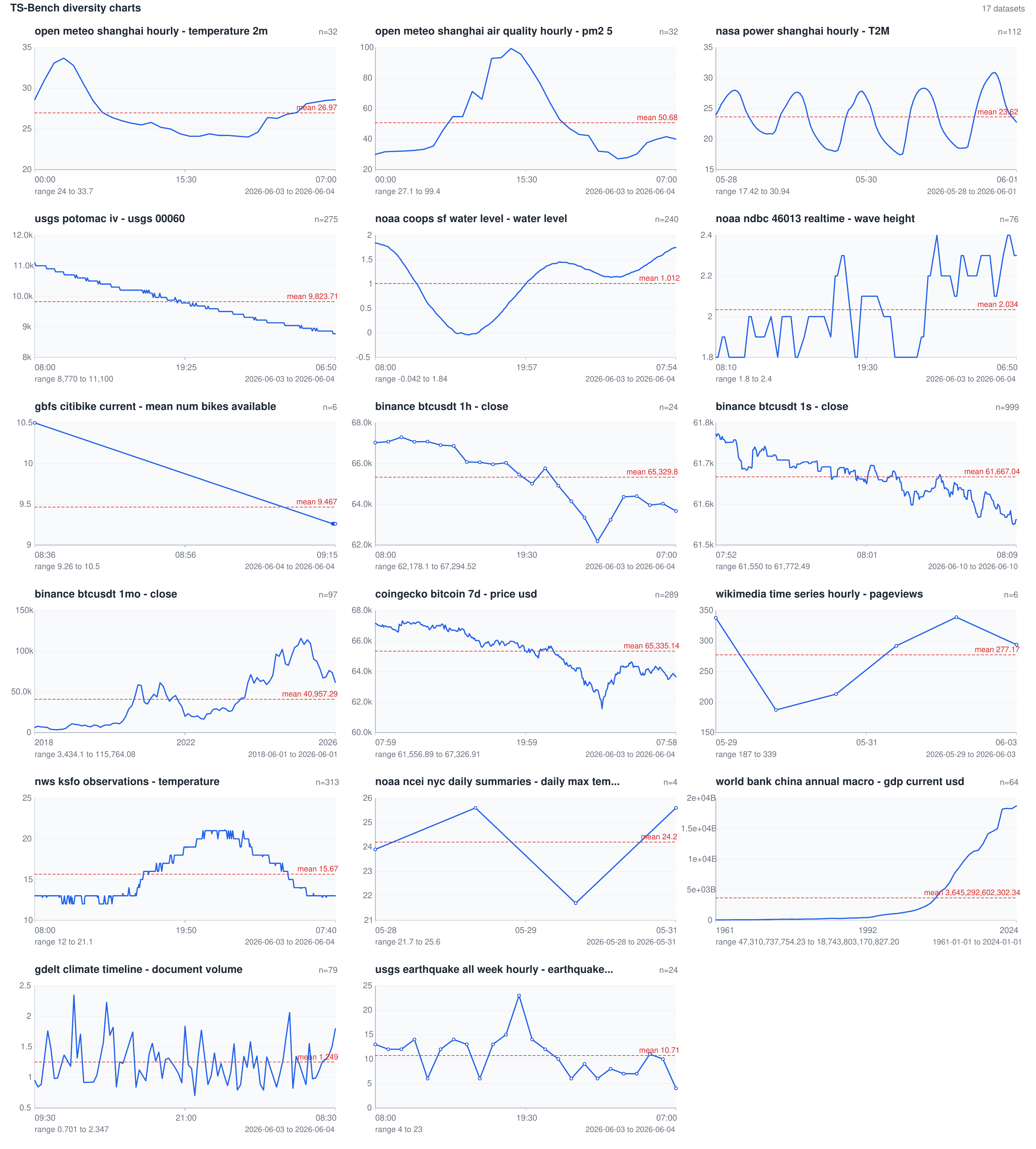}
    \caption{Verification-slice target trajectories for all 17 registry datasets. Each panel plots the locally parsed public sample for one dataset and target variable; panel titles give the dataset identifier and selected target.}
    \label{fig:verification-charts}
\end{figure*}

Figure~\ref{fig:verification-charts} shows the verification-slice target trajectories used to summarize the dynamic patterns in Table~\ref{tab:verification-patterns}.
Each panel corresponds to one registry dataset; together they illustrate the diversity of observed temporal behavior.
These are not fixed test windows but short public samples used to check that each source can be parsed into a regular numeric series. For the Citi Bike verification chart, each point averages available bikes across 50 stations before connecting the sequence; live evaluation can instead track individual stations or GBFS systems using the same public feed.

\subsection{Data attribution and usage}
\label{app:data-attribution}

We gratefully acknowledge Open-Meteo, NASA POWER, USGS, NOAA, GBFS/Citi Bike, Binance, CoinGecko, Wikimedia Foundation, World Bank Open Data, and GDELT as the public data providers summarized in Table~\ref{tab:app-dataset-background}.
Users operating a live deployment should respect each provider's terms of use, attribution requirements, and request-rate limits.

\section{Detailed Forecasting results}
\label{app_perdataset}
For completeness, we report per-dataset performance in Tables~\ref{tab_perdataset_rmse}, \ref{tab_perdataset_mape}, and~\ref{tab_perdataset_crps}, which present the RMSE, MAPE, and CRPS of every model in the frozen online benchmark snapshot on each dataset.

\begin{table*}[!t]
\centering
\caption{Per-dataset RMSE ($\downarrow$) on the online benchmark. In each column the best available model is in \textcolor{red}{\textbf{red bold}} and the second best is in \textcolor{blue}{\underline{blue underline}}.}
\label{tab_perdataset_rmse}
\resizebox{0.8 \linewidth}{!}{%
\begin{tabular}{l rrrrrrrrrr}
\toprule
Model & BTC & PM2.5 & Quake & Potomac & Water & Wave & T2M & KSFO & Temp2m & Wiki \\
\midrule
Chronos-2 & \textcolor{red}{\textbf{483.614}} & 11.420 & \textcolor{blue}{\underline{2.830}} & 19.675 & 0.318 & 0.892 & 1.293 & 1.078 & 1.074 & \textcolor{red}{\textbf{8.420}} \\
TiRex & 497.749 & \textcolor{red}{\textbf{10.843}} & 2.882 & \textcolor{red}{\textbf{16.958}} & 0.339 & 0.899 & 1.329 & 1.003 & \textcolor{blue}{\underline{1.025}} & 9.889 \\
TimesFM-2.5 & \textcolor{blue}{\underline{485.407}} & \textcolor{blue}{\underline{11.156}} & \textcolor{red}{\textbf{2.812}} & \textcolor{blue}{\underline{17.502}} & \textcolor{red}{\textbf{0.234}} & 0.844 & 1.332 & 1.060 & \textcolor{red}{\textbf{0.988}} & \textcolor{blue}{\underline{8.711}} \\
Toto-1.0 & 1.162e+03 & 40.693 & 5.314 & 58.600 & 0.539 & 1.093 & 1.578 & 1.208 & 1.835 & 96.250 \\
Moirai-2.0 & 569.016 & 19.226 & 3.619 & 24.663 & 0.361 & 0.308 & \textcolor{red}{\textbf{0.909}} & \textcolor{blue}{\underline{0.568}} & 1.511 & 13.782 \\
Chronos-Bolt & 802.832 & 20.359 & 3.917 & 28.495 & 0.634 & \textcolor{red}{\textbf{0.299}} & \textcolor{blue}{\underline{0.943}} & 0.578 & 1.630 & 33.431 \\
TabPFN-TS & 716.183 & 19.586 & 3.934 & 27.656 & 0.518 & \textcolor{blue}{\underline{0.307}} & 1.141 & \textcolor{red}{\textbf{0.515}} & 1.538 & 20.472 \\
Sundial & 1.344e+03 & 19.184 & 4.298 & 24.235 & \textcolor{blue}{\underline{0.294}} & 0.410 & 1.313 & 0.890 & 1.539 & 20.726 \\
\midrule
Moving-Average & 774.684 & 16.576 & 3.263 & 24.945 & 0.731 & 0.808 & 2.119 & 1.285 & 1.702 & 26.906 \\
ETS & 543.172 & 18.520 & 3.176 & 23.910 & 0.865 & 0.979 & 5.053 & 0.964 & 7.677 & 28.427 \\
ARIMA & 665.233 & 15.283 & 3.271 & 26.375 & 0.731 & 0.807 & 3.096 & 0.930 & 2.616 & 29.245 \\
Seasonal-Naive & 847.250 & 18.334 & 3.983 & 28.209 & 0.755 & 0.873 & 1.500 & 1.329 & 1.373 & 53.966 \\
\bottomrule
\end{tabular}}
\end{table*}

\begin{table*}[!t]
\centering
\caption{Per-dataset MAPE ($\downarrow$) on the online benchmark. In each column the best available model is in \textcolor{red}{\textbf{red bold}} and the second best is in \textcolor{blue}{\underline{blue underline}}.}
\label{tab_perdataset_mape}
\resizebox{0.8 \linewidth}{!}{%
\begin{tabular}{l rrrrrrrrrr}
\toprule
Model & BTC & PM2.5 & Quake & Potomac & Water & Wave & T2M & KSFO & Temp2m & Wiki \\
\midrule
Chronos-2 & 0.010 & 0.329 & 0.565 & 0.009 & 0.469 & 0.158 & 0.043 & 0.043 & 0.043 & 0.061 \\
TiRex & 0.010 & 0.315 & 0.578 & \textcolor{red}{\textbf{0.007}} & 0.591 & 0.164 & 0.043 & 0.038 & \textcolor{blue}{\underline{0.042}} & \textcolor{blue}{\underline{0.057}} \\
TimesFM-2.5 & 0.010 & \textcolor{blue}{\underline{0.314}} & \textcolor{blue}{\underline{0.493}} & \textcolor{blue}{\underline{0.008}} & \textcolor{red}{\textbf{0.283}} & \textcolor{red}{\textbf{0.145}} & 0.046 & 0.040 & 0.049 & 0.060 \\
Toto-1.0 & 0.011 & 0.333 & 0.509 & 0.010 & \textcolor{blue}{\underline{0.426}} & 0.171 & 0.034 & 0.031 & \textcolor{red}{\textbf{0.040}} & 0.210 \\
Moirai-2.0 & \textcolor{red}{\textbf{0.008}} & 0.341 & \textcolor{red}{\textbf{0.483}} & 0.009 & 0.437 & 0.178 & \textcolor{blue}{\underline{0.031}} & \textcolor{blue}{\underline{0.031}} & 0.045 & \textcolor{red}{\textbf{0.046}} \\
Chronos-Bolt & 0.011 & 0.364 & 0.546 & 0.010 & 0.722 & 0.176 & \textcolor{red}{\textbf{0.030}} & 0.031 & 0.045 & 0.134 \\
TabPFN-TS & 0.010 & 0.348 & 0.556 & 0.009 & 0.547 & 0.179 & 0.035 & \textcolor{red}{\textbf{0.026}} & 0.044 & 0.066 \\
Sundial & 0.020 & \textcolor{red}{\textbf{0.282}} & 0.534 & 0.008 & 0.600 & 0.204 & 0.040 & 0.046 & 0.045 & 0.073 \\
\midrule
Moving-Average & 0.014 & 0.477 & 0.628 & 0.011 & 1.858 & 0.158 & 0.075 & 0.067 & 0.068 & 0.200 \\
ETS & \textcolor{blue}{\underline{0.010}} & 0.713 & 0.585 & 0.010 & 1.340 & 0.213 & 0.176 & 0.044 & 0.321 & 0.203 \\
ARIMA & 0.014 & 0.439 & 0.580 & 0.012 & 1.378 & \textcolor{blue}{\underline{0.155}} & 0.100 & 0.038 & 0.115 & 0.191 \\
Seasonal-Naive & 0.015 & 0.518 & 0.641 & 0.011 & 1.863 & 0.166 & 0.052 & 0.065 & 0.052 & 0.361 \\
\bottomrule
\end{tabular}}
\end{table*}

\begin{table*}[!t]
\centering
\caption{Per-dataset CRPS ($\downarrow$) on the online benchmark. In each column the best available model is in \textcolor{red}{\textbf{red bold}} and the second best is in \textcolor{blue}{\underline{blue underline}}.}
\label{tab_perdataset_crps}
\resizebox{0.8 \linewidth}{!}{%
\begin{tabular}{l rrrrrrrrrr}
\toprule
Model & BTC & PM2.5 & Quake & Potomac & Water & Wave & T2M & KSFO & Temp2m & Wiki \\
\midrule
Chronos-2 & 0.112 & 0.445 & 0.345 & 0.139 & 0.166 & 0.445 & 0.059 & 0.438 & 0.092 & 0.122 \\
TiRex & 0.118 & 0.436 & 0.347 & 0.139 & 0.174 & 0.428 & 0.056 & 0.408 & 0.100 & 0.118 \\
TimesFM-2.5 & 0.136 & 0.518 & 0.481 & 0.172 & \textcolor{red}{\textbf{0.147}} & 0.505 & 0.071 & 0.499 & 0.111 & 0.161 \\
Toto-1.0 & \textcolor{blue}{\underline{0.010}} & \textcolor{red}{\textbf{0.245}} & \textcolor{blue}{\underline{0.273}} & 0.010 & 0.186 & \textcolor{blue}{\underline{0.128}} & 0.027 & \textcolor{blue}{\underline{0.027}} & \textcolor{red}{\textbf{0.036}} & 0.132 \\
Moirai-2.0 & \textcolor{red}{\textbf{0.008}} & 0.260 & \textcolor{red}{\textbf{0.262}} & \textcolor{blue}{\underline{0.008}} & 0.207 & \textcolor{red}{\textbf{0.124}} & \textcolor{red}{\textbf{0.027}} & 0.028 & 0.040 & \textcolor{red}{\textbf{0.050}} \\
Chronos-Bolt & 0.010 & 0.272 & 0.278 & 0.010 & 0.365 & 0.129 & \textcolor{blue}{\underline{0.027}} & 0.028 & 0.041 & 0.111 \\
TabPFN-TS & 0.010 & 0.334 & 0.371 & 0.009 & 0.339 & 0.171 & 0.036 & \textcolor{red}{\textbf{0.026}} & 0.045 & 0.063 \\
Sundial & 0.018 & \textcolor{blue}{\underline{0.257}} & 0.296 & \textcolor{red}{\textbf{0.007}} & \textcolor{blue}{\underline{0.151}} & 0.164 & 0.033 & 0.039 & \textcolor{blue}{\underline{0.040}} & \textcolor{blue}{\underline{0.054}} \\
\midrule
Moving-Average & 0.107 & 0.490 & 0.348 & 0.132 & 0.597 & 0.451 & 0.106 & 0.435 & 0.178 & 0.321 \\
ETS & 0.099 & 0.829 & 0.348 & 0.111 & 0.575 & 0.516 & 0.279 & 0.301 & 0.903 & 0.323 \\
ARIMA & 0.117 & 0.457 & 0.338 & 0.117 & 0.514 & 0.458 & 0.174 & 0.345 & 0.356 & 0.284 \\
Seasonal-Naive & 0.100 & 0.583 & 0.556 & 0.114 & 0.621 & 0.569 & 0.073 & 0.416 & 0.136 & 0.731 \\
\bottomrule
\end{tabular}}
\end{table*}

\begin{table*}[!t]
\centering
\caption{Per-dataset Temporal Stability ($\downarrow$) on the online benchmark, computed from release-level MSE histories. Cells with fewer than two releases are shown as --. In each column the best available model is in \textcolor{red}{\textbf{red bold}} and the second best is in \textcolor{blue}{\underline{blue underline}}.}
\label{tab_perdataset_stability}
\resizebox{0.8 \linewidth}{!}{%
\begin{tabular}{l rrrrrrrrrr}
\toprule
Model & BTC & PM2.5 & Quake & Potomac & Water & Wave & T2M & KSFO & Temp2m & Wiki \\
\midrule
Chronos-2 & 7.329e+05 & 494.984 & 8.908 & 2.860e+03 & 0.271 & 3.756 & 1.756 & 27.049 & 1.953 & 129.300 \\
TiRex & 6.805e+05 & 463.452 & 8.575 & \textcolor{blue}{\underline{2.414e+03}} & 0.310 & 3.479 & 1.653 & 27.449 & 1.788 & 167.781 \\
TimesFM-2.5 & 6.407e+05 & 499.814 & 8.574 & 2.685e+03 & \textcolor{blue}{\underline{0.137}} & 3.345 & 1.964 & 27.191 & 1.673 & 133.699 \\
Toto-1.0 & 3.767e+06 & 8.411e+03 & 26.727 & 9.212e+03 & 0.653 & 7.889 & 7.376 & 3.933 & 6.619 & 8.228e+03 \\
Moirai-2.0 & \textcolor{blue}{\underline{4.540e+05}} & \textcolor{blue}{\underline{185.713}} & \textcolor{blue}{\underline{5.250}} & 2.734e+03 & 0.192 & 0.088 & \textcolor{red}{\textbf{0.375}} & \textcolor{blue}{\underline{0.737}} & 1.538 & \textcolor{blue}{\underline{64.521}} \\
Chronos-Bolt & 6.818e+05 & \textcolor{red}{\textbf{169.701}} & 5.282 & 3.085e+03 & 0.358 & \textcolor{red}{\textbf{0.052}} & 1.235 & 0.981 & 2.238 & 238.712 \\
TabPFN-TS & \textcolor{red}{\textbf{4.069e+05}} & 307.897 & 5.972 & 2.749e+03 & 0.288 & \textcolor{blue}{\underline{0.081}} & 0.928 & \textcolor{red}{\textbf{0.701}} & \textcolor{red}{\textbf{1.273}} & 68.714 \\
Sundial & 1.307e+06 & 645.449 & \textcolor{red}{\textbf{5.049}} & \textcolor{red}{\textbf{2.020e+03}} & \textcolor{red}{\textbf{0.117}} & 0.102 & \textcolor{blue}{\underline{0.666}} & 2.253 & \textcolor{blue}{\underline{1.509}} & \textcolor{red}{\textbf{26.124}} \\
\midrule
Moving-Average & 9.664e+05 & 571.085 & 9.040 & 3.870e+03 & 1.352 & 3.555 & 5.080 & 26.020 & 2.855 & 1.196e+03 \\
ETS & 9.012e+05 & 7.217e+03 & 8.041 & 2.763e+03 & 1.431 & 4.353 & 29.082 & 26.072 & 209.632 & 1.770e+03 \\
ARIMA & 7.312e+05 & 943.393 & 8.244 & 3.046e+03 & 1.027 & 3.510 & 11.488 & 25.728 & 10.261 & 1.427e+03 \\
Seasonal-Naive & 1.002e+06 & 644.216 & 12.657 & 4.223e+03 & 1.389 & 3.750 & 2.021 & 25.672 & 1.829 & 4.084e+03 \\
\bottomrule
\end{tabular}}
\end{table*}

\begin{table*}[!t]
\centering
\caption{Per-dataset Improvement ($\downarrow$; more negative is better) on the online benchmark, computed as Kendall $\tau$ on release-level MSE histories. Cells with fewer than two releases are shown as --. In each column the best available model is in \textcolor{red}{\textbf{red bold}} and the second best is in \textcolor{blue}{\underline{blue underline}}.}
\label{tab_perdataset_improvement}
\resizebox{0.8 \linewidth}{!}{%
\begin{tabular}{l rrrrrrrrrr}
\toprule
Model & BTC & PM2.5 & Quake & Potomac & Water & Wave & T2M & KSFO & Temp2m & Wiki \\
\midrule
Chronos-2 & 0.560 & 0.588 & 0.398 & 0.224 & -0.033 & -0.500 & -0.138 & -0.025 & 0.527 & 0.704 \\
TiRex & 0.565 & 0.545 & 0.429 & 0.242 & -0.010 & \textcolor{blue}{\underline{-0.504}} & 0.144 & -0.016 & 0.547 & 0.873 \\
TimesFM-2.5 & 0.531 & 0.548 & 0.431 & 0.247 & -0.051 & -0.428 & -0.002 & -0.034 & 0.585 & 0.479 \\
Toto-1.0 & -0.309 & \textcolor{blue}{\underline{0.059}} & -0.273 & \textcolor{red}{\textbf{0.072}} & 0.022 & -0.085 & 0.344 & 0.123 & 0.022 & 1.000 \\
Moirai-2.0 & \textcolor{blue}{\underline{-0.355}} & \textcolor{red}{\textbf{-0.013}} & -0.351 & 0.132 & -0.028 & \textcolor{red}{\textbf{-0.639}} & -0.048 & \textcolor{red}{\textbf{-0.195}} & \textcolor{blue}{\underline{-0.385}} & \textcolor{red}{\textbf{-1.000}} \\
Chronos-Bolt & 0.090 & 0.307 & \textcolor{blue}{\underline{-0.371}} & 0.151 & 0.014 & -0.098 & -0.048 & \textcolor{blue}{\underline{-0.143}} & \textcolor{red}{\textbf{-0.429}} & 1.000 \\
TabPFN-TS & \textcolor{red}{\textbf{-0.505}} & 0.200 & \textcolor{red}{\textbf{-0.415}} & 0.121 & 0.000 & -0.237 & 0.023 & -0.086 & -0.114 & 0.333 \\
Sundial & 0.503 & 0.066 & -0.271 & \textcolor{blue}{\underline{0.084}} & -0.080 & -0.221 & \textcolor{red}{\textbf{-0.505}} & -0.134 & -0.096 & \textcolor{blue}{\underline{0.000}} \\
\midrule
Moving-Average & 0.458 & 0.520 & 0.351 & 0.176 & \textcolor{blue}{\underline{-0.321}} & -0.395 & -0.391 & -0.081 & 0.525 & 0.423 \\
ETS & 0.345 & 0.533 & 0.374 & 0.221 & -0.267 & -0.476 & -0.379 & 0.054 & 0.296 & 0.366 \\
ARIMA & 0.393 & 0.555 & 0.414 & 0.191 & -0.313 & -0.464 & \textcolor{blue}{\underline{-0.456}} & -0.079 & 0.364 & 0.592 \\
Seasonal-Naive & 0.484 & 0.459 & 0.381 & 0.134 & \textcolor{red}{\textbf{-0.329}} & -0.419 & -0.036 & -0.074 & 0.491 & 0.477 \\
\bottomrule
\end{tabular}}
\end{table*}

\begin{table*}[!t]
\centering
\caption{Overall zero-shot standing on the online benchmark aggregated over all datasets. For each metric we report Average Rank ($\downarrow$), Win Rate ($\uparrow$), and Elo ($\uparrow$) computed from the per-dataset values. In each row the best available model is in \textcolor{red}{\textbf{red bold}} and the second best is in \textcolor{blue}{\underline{blue underline}}.}
\label{tab_rq1_overall}
\resizebox{0.8 \linewidth}{!}{%
\begin{tabular}{l rrrrrrrrrrrr}
\toprule
Metric & Chr2 & TiRex & TFM & Toto1 & Moi2 & ChrB & TabPFN & Sundial & ARIMA & ETS & MovAvg & SNaive \\
\midrule
Rank (RMSE) $\downarrow$ & \textcolor{blue}{\underline{3.80}} & 4.00 & \textcolor{red}{\textbf{3.30}} & 10.70 & 4.70 & 7.10 & 5.80 & 6.40 & 7.45 & 8.00 & 7.75 & 9.00 \\
Win Rate (RMSE) $\uparrow$ & \textcolor{blue}{\underline{0.745}} & 0.727 & \textcolor{red}{\textbf{0.791}} & 0.118 & 0.664 & 0.445 & 0.564 & 0.509 & 0.414 & 0.364 & 0.386 & 0.273 \\
Elo (RMSE) $\uparrow$ & \textcolor{blue}{\underline{1225}} & 1194 & \textcolor{red}{\textbf{1254}} & 637 & 1218 & 947 & 1161 & 1070 & 834 & 789 & 867 & 804 \\
\addlinespace[3pt]
Rank (MAPE) $\downarrow$ & 4.90 & 4.30 & \textcolor{blue}{\underline{3.85}} & 5.05 & \textcolor{red}{\textbf{3.70}} & 6.15 & 5.10 & 6.45 & 8.70 & 9.80 & 9.65 & 10.35 \\
Win Rate (MAPE) $\uparrow$ & 0.645 & 0.700 & \textcolor{blue}{\underline{0.741}} & 0.632 & \textcolor{red}{\textbf{0.755}} & 0.532 & 0.627 & 0.505 & 0.300 & 0.200 & 0.214 & 0.150 \\
Elo (MAPE) $\uparrow$ & 1163 & \textcolor{blue}{\underline{1251}} & 1139 & 1055 & \textcolor{red}{\textbf{1253}} & 1064 & 1177 & 998 & 846 & 650 & 739 & 666 \\
\addlinespace[3pt]
Rank (CRPS) $\downarrow$ & 7.25 & 7.05 & 9.20 & \textcolor{blue}{\underline{2.95}} & \textcolor{red}{\textbf{2.30}} & 3.90 & 4.70 & 3.15 & 8.70 & 9.45 & 9.35 & 10.00 \\
Win Rate (CRPS) $\uparrow$ & 0.432 & 0.450 & 0.255 & \textcolor{blue}{\underline{0.823}} & \textcolor{red}{\textbf{0.882}} & 0.736 & 0.664 & 0.805 & 0.300 & 0.232 & 0.241 & 0.182 \\
Elo (CRPS) $\uparrow$ & 930 & 1003 & 780 & 1325 & \textcolor{red}{\textbf{1470}} & 1265 & 1243 & \textcolor{blue}{\underline{1358}} & 726 & 623 & 666 & 611 \\
\addlinespace[3pt]
Rank (Stability) $\downarrow$ & 6.90 & 5.80 & 5.30 & 10.50 & \textcolor{red}{\textbf{2.40}} & 4.80 & \textcolor{blue}{\underline{2.80}} & 3.60 & 8.40 & 9.60 & 8.80 & 9.10 \\
Win Rate (Stability) $\uparrow$ & 0.464 & 0.564 & 0.609 & 0.136 & \textcolor{red}{\textbf{0.873}} & 0.655 & \textcolor{blue}{\underline{0.836}} & 0.764 & 0.327 & 0.218 & 0.291 & 0.264 \\
Elo (Stability) $\uparrow$ & 989 & 992 & 1059 & 648 & \textcolor{red}{\textbf{1423}} & 1073 & \textcolor{blue}{\underline{1410}} & 1407 & 764 & 625 & 814 & 797 \\
\addlinespace[3pt]
Rank (Improvement) $\downarrow$ & 8.50 & 9.60 & 9.20 & 7.45 & \textcolor{red}{\textbf{2.95}} & 6.00 & 4.80 & \textcolor{blue}{\underline{4.40}} & 6.60 & 6.20 & 5.90 & 6.40 \\
Win Rate (Improvement) $\uparrow$ & 0.318 & 0.218 & 0.255 & 0.414 & \textcolor{red}{\textbf{0.823}} & 0.545 & 0.655 & \textcolor{blue}{\underline{0.691}} & 0.491 & 0.527 & 0.555 & 0.509 \\
Elo (Improvement) $\uparrow$ & 849 & 753 & 838 & 776 & \textcolor{red}{\textbf{1338}} & 990 & 1161 & \textcolor{blue}{\underline{1242}} & 981 & 1059 & 1036 & 977 \\
\bottomrule
\end{tabular}}
\end{table*}

\begin{table*}[!t]
\footnotesize
\tabcolsep=0.25cm
\centering
\caption{Zero-shot rankings on the online benchmark aggregated by domain. For each metric we report Average Rank ($\downarrow$), Win Rate ($\uparrow$), and Elo ($\uparrow$) computed from the per-dataset values within each domain. In each row the best available model is in \textcolor{red}{\textbf{red bold}} and the second best is in \textcolor{blue}{\underline{blue underline}}.}
\label{tab_rq1_domain}
\begin{tabularx}{\textwidth}{ll rrrrrrrrrrrr}
\toprule
Group & Metric & Chr2 & TiRex & TFM & Toto1 & Moi2 & ChrB & TabPFN & Sundial & ARIMA & ETS & MovAvg & SNaive \\
\midrule
\multirow{9}{*}{Air Quality} & Rank (RMSE) $\downarrow$ & 3.00 & \textcolor{red}{\textbf{1.00}} & \textcolor{blue}{\underline{2.00}} & 12.00 & 9.00 & 11.00 & 10.00 & 8.00 & 4.00 & 7.00 & 5.00 & 6.00 \\
 & Win Rate (RMSE) $\uparrow$ & 0.818 & \textcolor{red}{\textbf{1.000}} & \textcolor{blue}{\underline{0.909}} & 0.000 & 0.273 & 0.091 & 0.182 & 0.364 & 0.727 & 0.455 & 0.636 & 0.545 \\
 & Elo (RMSE) $\uparrow$ & 1451 & \textcolor{red}{\textbf{1786}} & \textcolor{blue}{\underline{1601}} & 218 & 684 & 397 & 548 & 807 & 1326 & 931 & 1191 & 1059 \\
 & Rank (MAPE) $\downarrow$ & 4.00 & 3.00 & \textcolor{blue}{\underline{2.00}} & 5.00 & 6.00 & 8.00 & 7.00 & \textcolor{red}{\textbf{1.00}} & 9.00 & 12.00 & 10.00 & 11.00 \\
 & Win Rate (MAPE) $\uparrow$ & 0.727 & 0.818 & \textcolor{blue}{\underline{0.909}} & 0.636 & 0.545 & 0.364 & 0.455 & \textcolor{red}{\textbf{1.000}} & 0.273 & 0.000 & 0.182 & 0.091 \\
 & Elo (MAPE) $\uparrow$ & 1312 & 1438 & \textcolor{blue}{\underline{1588}} & 1192 & 1068 & 817 & 950 & \textcolor{red}{\textbf{1783}} & 695 & 206 & 553 & 397 \\
 & Rank (CRPS) $\downarrow$ & 7.00 & 6.00 & 10.00 & \textcolor{red}{\textbf{1.00}} & 3.00 & 4.00 & 5.00 & \textcolor{blue}{\underline{2.00}} & 8.00 & 12.00 & 9.00 & 11.00 \\
 & Win Rate (CRPS) $\uparrow$ & 0.455 & 0.545 & 0.182 & \textcolor{red}{\textbf{1.000}} & 0.818 & 0.727 & 0.636 & \textcolor{blue}{\underline{0.909}} & 0.364 & 0.000 & 0.273 & 0.091 \\
 & Elo (CRPS) $\uparrow$ & 941 & 1062 & 560 & \textcolor{red}{\textbf{1783}} & 1444 & 1312 & 1178 & \textcolor{blue}{\underline{1606}} & 823 & 208 & 683 & 399 \\
\midrule
\multirow{9}{*}{Finance} & Rank (RMSE) $\downarrow$ & \textcolor{red}{\textbf{1.00}} & 3.00 & \textcolor{blue}{\underline{2.00}} & 11.00 & 5.00 & 9.00 & 7.00 & 12.00 & 6.00 & 4.00 & 8.00 & 10.00 \\
 & Win Rate (RMSE) $\uparrow$ & \textcolor{red}{\textbf{1.000}} & 0.818 & \textcolor{blue}{\underline{0.909}} & 0.091 & 0.636 & 0.273 & 0.455 & 0.000 & 0.545 & 0.727 & 0.364 & 0.182 \\
 & Elo (RMSE) $\uparrow$ & \textcolor{red}{\textbf{1777}} & 1441 & \textcolor{blue}{\underline{1603}} & 400 & 1190 & 688 & 938 & 216 & 1066 & 1321 & 819 & 540 \\
 & Rank (MAPE) $\downarrow$ & \textcolor{blue}{\underline{4.00}} & \textcolor{blue}{\underline{4.00}} & \textcolor{blue}{\underline{4.00}} & 7.50 & \textcolor{red}{\textbf{1.00}} & 7.50 & \textcolor{blue}{\underline{4.00}} & 12.00 & 9.50 & \textcolor{blue}{\underline{4.00}} & 9.50 & 11.00 \\
 & Win Rate (MAPE) $\uparrow$ & \textcolor{blue}{\underline{0.727}} & \textcolor{blue}{\underline{0.727}} & \textcolor{blue}{\underline{0.727}} & 0.409 & \textcolor{red}{\textbf{1.000}} & 0.409 & \textcolor{blue}{\underline{0.727}} & 0.000 & 0.227 & \textcolor{blue}{\underline{0.727}} & 0.227 & 0.091 \\
 & Elo (MAPE) $\uparrow$ & 1300 & 1301 & 1301 & 914 & \textcolor{red}{\textbf{1728}} & 909 & 1300 & 228 & 651 & \textcolor{blue}{\underline{1303}} & 652 & 414 \\
 & Rank (CRPS) $\downarrow$ & 9.00 & 11.00 & 12.00 & \textcolor{blue}{\underline{3.00}} & \textcolor{red}{\textbf{1.00}} & \textcolor{blue}{\underline{3.00}} & \textcolor{blue}{\underline{3.00}} & 5.00 & 10.00 & 6.00 & 8.00 & 7.00 \\
 & Win Rate (CRPS) $\uparrow$ & 0.273 & 0.091 & 0.000 & \textcolor{blue}{\underline{0.818}} & \textcolor{red}{\textbf{1.000}} & \textcolor{blue}{\underline{0.818}} & \textcolor{blue}{\underline{0.818}} & 0.636 & 0.182 & 0.545 & 0.364 & 0.455 \\
 & Elo (CRPS) $\uparrow$ & 680 & 410 & 218 & \textcolor{blue}{\underline{1450}} & \textcolor{red}{\textbf{1766}} & 1445 & 1447 & 1199 & 550 & 1077 & 814 & 945 \\
\midrule
\multirow{9}{*}{Hazards} & Rank (RMSE) $\downarrow$ & \textcolor{blue}{\underline{2.00}} & 3.00 & \textcolor{red}{\textbf{1.00}} & 12.00 & 7.00 & 8.00 & 9.00 & 11.00 & 6.00 & 4.00 & 5.00 & 10.00 \\
 & Win Rate (RMSE) $\uparrow$ & \textcolor{blue}{\underline{0.909}} & 0.818 & \textcolor{red}{\textbf{1.000}} & 0.000 & 0.455 & 0.364 & 0.273 & 0.091 & 0.545 & 0.727 & 0.636 & 0.182 \\
 & Elo (RMSE) $\uparrow$ & \textcolor{blue}{\underline{1595}} & 1449 & \textcolor{red}{\textbf{1782}} & 217 & 938 & 823 & 694 & 392 & 1056 & 1319 & 1190 & 544 \\
 & Rank (MAPE) $\downarrow$ & 7.00 & 8.00 & \textcolor{blue}{\underline{2.00}} & 3.00 & \textcolor{red}{\textbf{1.00}} & 5.00 & 6.00 & 4.00 & 9.00 & 10.00 & 11.00 & 12.00 \\
 & Win Rate (MAPE) $\uparrow$ & 0.455 & 0.364 & \textcolor{blue}{\underline{0.909}} & 0.818 & \textcolor{red}{\textbf{1.000}} & 0.636 & 0.545 & 0.727 & 0.273 & 0.182 & 0.091 & 0.000 \\
 & Elo (MAPE) $\uparrow$ & 935 & 828 & \textcolor{blue}{\underline{1603}} & 1453 & \textcolor{red}{\textbf{1777}} & 1182 & 1055 & 1317 & 696 & 548 & 399 & 207 \\
 & Rank (CRPS) $\downarrow$ & 6.00 & 7.00 & 11.00 & \textcolor{blue}{\underline{2.00}} & \textcolor{red}{\textbf{1.00}} & 3.00 & 10.00 & 4.00 & 5.00 & 8.50 & 8.50 & 12.00 \\
 & Win Rate (CRPS) $\uparrow$ & 0.545 & 0.455 & 0.091 & \textcolor{blue}{\underline{0.909}} & \textcolor{red}{\textbf{1.000}} & 0.818 & 0.182 & 0.727 & 0.636 & 0.318 & 0.318 & 0.000 \\
 & Elo (CRPS) $\uparrow$ & 1063 & 942 & 410 & \textcolor{blue}{\underline{1601}} & \textcolor{red}{\textbf{1778}} & 1452 & 553 & 1312 & 1175 & 750 & 752 & 212 \\
\midrule
\multirow{9}{*}{Hydrology} & Rank (RMSE) $\downarrow$ & 3.00 & \textcolor{red}{\textbf{1.00}} & \textcolor{blue}{\underline{2.00}} & 12.00 & 6.00 & 11.00 & 9.00 & 5.00 & 8.00 & 4.00 & 7.00 & 10.00 \\
 & Win Rate (RMSE) $\uparrow$ & 0.818 & \textcolor{red}{\textbf{1.000}} & \textcolor{blue}{\underline{0.909}} & 0.000 & 0.545 & 0.091 & 0.273 & 0.636 & 0.364 & 0.727 & 0.455 & 0.182 \\
 & Elo (RMSE) $\uparrow$ & 1456 & \textcolor{red}{\textbf{1784}} & \textcolor{blue}{\underline{1599}} & 216 & 1064 & 391 & 690 & 1187 & 812 & 1313 & 944 & 544 \\
 & Rank (MAPE) $\downarrow$ & 5.00 & \textcolor{red}{\textbf{1.00}} & \textcolor{blue}{\underline{2.50}} & 8.00 & 5.00 & 8.00 & 5.00 & \textcolor{blue}{\underline{2.50}} & 12.00 & 8.00 & 10.50 & 10.50 \\
 & Win Rate (MAPE) $\uparrow$ & 0.636 & \textcolor{red}{\textbf{1.000}} & \textcolor{blue}{\underline{0.864}} & 0.364 & 0.636 & 0.364 & 0.636 & \textcolor{blue}{\underline{0.864}} & 0.000 & 0.364 & 0.136 & 0.136 \\
 & Elo (MAPE) $\uparrow$ & 1176 & \textcolor{red}{\textbf{1763}} & 1500 & 831 & 1175 & 826 & 1179 & \textcolor{blue}{\underline{1509}} & 223 & 825 & 496 & 497 \\
 & Rank (CRPS) $\downarrow$ & 10.50 & 10.50 & 12.00 & 4.50 & \textcolor{blue}{\underline{2.00}} & 4.50 & 3.00 & \textcolor{red}{\textbf{1.00}} & 8.00 & 6.00 & 9.00 & 7.00 \\
 & Win Rate (CRPS) $\uparrow$ & 0.136 & 0.136 & 0.000 & 0.682 & \textcolor{blue}{\underline{0.909}} & 0.682 & 0.818 & \textcolor{red}{\textbf{1.000}} & 0.364 & 0.545 & 0.273 & 0.455 \\
 & Elo (CRPS) $\uparrow$ & 483 & 484 & 223 & 1253 & \textcolor{blue}{\underline{1592}} & 1250 & 1444 & \textcolor{red}{\textbf{1782}} & 804 & 1069 & 676 & 941 \\
\midrule
\multirow{9}{*}{Ocean} & Rank (RMSE) $\downarrow$ & 6.00 & 7.00 & \textcolor{blue}{\underline{4.00}} & 9.50 & \textcolor{blue}{\underline{4.00}} & 4.50 & \textcolor{blue}{\underline{4.00}} & \textcolor{red}{\textbf{3.00}} & 7.25 & 11.50 & 7.75 & 9.50 \\
 & Win Rate (RMSE) $\uparrow$ & 0.545 & 0.455 & \textcolor{blue}{\underline{0.727}} & 0.227 & \textcolor{blue}{\underline{0.727}} & 0.682 & \textcolor{blue}{\underline{0.727}} & \textcolor{red}{\textbf{0.818}} & 0.432 & 0.045 & 0.386 & 0.227 \\
 & Elo (RMSE) $\uparrow$ & 1002 & 924 & 1165 & 705 & 1251 & 1252 & \textcolor{blue}{\underline{1268}} & \textcolor{red}{\textbf{1304}} & 990 & 439 & 941 & 759 \\
 & Rank (MAPE) $\downarrow$ & \textcolor{blue}{\underline{3.75}} & 5.50 & \textcolor{red}{\textbf{1.00}} & 4.50 & 6.00 & 8.00 & 7.50 & 9.00 & 6.00 & 10.50 & 7.25 & 9.00 \\
 & Win Rate (MAPE) $\uparrow$ & \textcolor{blue}{\underline{0.750}} & 0.591 & \textcolor{red}{\textbf{1.000}} & 0.682 & 0.545 & 0.364 & 0.409 & 0.273 & 0.545 & 0.136 & 0.432 & 0.273 \\
 & Elo (MAPE) $\uparrow$ & \textcolor{blue}{\underline{1212}} & 1061 & \textcolor{red}{\textbf{1762}} & 1080 & 958 & 859 & 858 & 754 & 1066 & 611 & 967 & 812 \\
 & Rank (CRPS) $\downarrow$ & 5.00 & 5.00 & 5.50 & \textcolor{blue}{\underline{3.50}} & \textcolor{blue}{\underline{3.50}} & 5.50 & 6.00 & \textcolor{red}{\textbf{3.00}} & 9.00 & 10.50 & 9.50 & 12.00 \\
 & Win Rate (CRPS) $\uparrow$ & 0.636 & 0.636 & 0.591 & \textcolor{blue}{\underline{0.773}} & \textcolor{blue}{\underline{0.773}} & 0.591 & 0.545 & \textcolor{red}{\textbf{0.818}} & 0.273 & 0.136 & 0.227 & 0.000 \\
 & Elo (CRPS) $\uparrow$ & 1158 & 1180 & 1072 & \textcolor{blue}{\underline{1363}} & \textcolor{red}{\textbf{1383}} & 1201 & 1124 & 1361 & 775 & 522 & 725 & 136 \\
\midrule
\multirow{9}{*}{Weather} & Rank (RMSE) $\downarrow$ & 5.33 & 5.00 & 5.33 & 9.67 & \textcolor{red}{\textbf{2.67}} & 4.33 & \textcolor{blue}{\underline{3.33}} & 5.33 & 9.00 & 10.00 & 10.00 & 8.00 \\
 & Win Rate (RMSE) $\uparrow$ & 0.606 & 0.636 & 0.606 & 0.212 & \textcolor{red}{\textbf{0.848}} & 0.697 & \textcolor{blue}{\underline{0.788}} & 0.606 & 0.273 & 0.182 & 0.182 & 0.364 \\
 & Elo (RMSE) $\uparrow$ & 1125 & 1172 & 1168 & 720 & \textcolor{red}{\textbf{1306}} & 1111 & \textcolor{blue}{\underline{1245}} & 1075 & 776 & 668 & 706 & 928 \\
 & Rank (MAPE) $\downarrow$ & 5.83 & 4.67 & 7.67 & \textcolor{red}{\textbf{2.33}} & 3.67 & 3.33 & \textcolor{blue}{\underline{3.00}} & 7.00 & 9.17 & 11.00 & 10.67 & 9.67 \\
 & Win Rate (MAPE) $\uparrow$ & 0.561 & 0.667 & 0.394 & \textcolor{red}{\textbf{0.879}} & 0.758 & 0.788 & \textcolor{blue}{\underline{0.818}} & 0.455 & 0.258 & 0.091 & 0.121 & 0.212 \\
 & Elo (MAPE) $\uparrow$ & 1143 & 1265 & 884 & \textcolor{red}{\textbf{1518}} & 1281 & 1309 & \textcolor{blue}{\underline{1411}} & 951 & 681 & 426 & 490 & 641 \\
 & Rank (CRPS) $\downarrow$ & 8.00 & 7.00 & 9.33 & \textcolor{red}{\textbf{1.67}} & \textcolor{blue}{\underline{2.67}} & 3.17 & 3.67 & 3.83 & 9.67 & 10.00 & 10.00 & 9.00 \\
 & Win Rate (CRPS) $\uparrow$ & 0.364 & 0.455 & 0.242 & \textcolor{red}{\textbf{0.939}} & \textcolor{blue}{\underline{0.848}} & 0.803 & 0.758 & 0.742 & 0.212 & 0.182 & 0.182 & 0.273 \\
 & Elo (CRPS) $\uparrow$ & 741 & 850 & 596 & \textcolor{red}{\textbf{1760}} & \textcolor{blue}{\underline{1575}} & 1482 & 1418 & 1420 & 536 & 494 & 504 & 623 \\
\midrule
\multirow{9}{*}{Web} & Rank (RMSE) $\downarrow$ & \textcolor{red}{\textbf{1.00}} & 3.00 & \textcolor{blue}{\underline{2.00}} & 12.00 & 4.00 & 10.00 & 5.00 & 6.00 & 9.00 & 8.00 & 7.00 & 11.00 \\
 & Win Rate (RMSE) $\uparrow$ & \textcolor{red}{\textbf{1.000}} & 0.818 & \textcolor{blue}{\underline{0.909}} & 0.000 & 0.727 & 0.182 & 0.636 & 0.545 & 0.273 & 0.364 & 0.455 & 0.091 \\
 & Elo (RMSE) $\uparrow$ & \textcolor{red}{\textbf{1772}} & 1441 & \textcolor{blue}{\underline{1602}} & 215 & 1316 & 550 & 1195 & 1071 & 683 & 821 & 947 & 387 \\
 & Rank (MAPE) $\downarrow$ & 4.00 & \textcolor{blue}{\underline{2.00}} & 3.00 & 11.00 & \textcolor{red}{\textbf{1.00}} & 7.00 & 5.00 & 6.00 & 8.00 & 10.00 & 9.00 & 12.00 \\
 & Win Rate (MAPE) $\uparrow$ & 0.727 & \textcolor{blue}{\underline{0.909}} & 0.818 & 0.091 & \textcolor{red}{\textbf{1.000}} & 0.455 & 0.636 & 0.545 & 0.364 & 0.182 & 0.273 & 0.000 \\
 & Elo (MAPE) $\uparrow$ & 1316 & \textcolor{blue}{\underline{1600}} & 1447 & 407 & \textcolor{red}{\textbf{1774}} & 935 & 1193 & 1069 & 822 & 546 & 684 & 208 \\
 & Rank (CRPS) $\downarrow$ & 6.00 & 5.00 & 8.00 & 7.00 & \textcolor{red}{\textbf{1.00}} & 4.00 & 3.00 & \textcolor{blue}{\underline{2.00}} & 9.00 & 11.00 & 10.00 & 12.00 \\
 & Win Rate (CRPS) $\uparrow$ & 0.545 & 0.636 & 0.364 & 0.455 & \textcolor{red}{\textbf{1.000}} & 0.727 & 0.818 & \textcolor{blue}{\underline{0.909}} & 0.273 & 0.091 & 0.182 & 0.000 \\
 & Elo (CRPS) $\uparrow$ & 1063 & 1183 & 818 & 949 & \textcolor{red}{\textbf{1787}} & 1302 & 1440 & \textcolor{blue}{\underline{1596}} & 698 & 402 & 552 & 209 \\
\bottomrule
\end{tabularx}
\end{table*}

\begin{table*}[!t]
\centering
\caption{Zero-shot rankings on the online benchmark aggregated by sampling frequency. For each metric we report Average Rank ($\downarrow$), Win Rate ($\uparrow$), and Elo ($\uparrow$) computed from the per-dataset values within each frequency group. In each row the best available model is in \textcolor{red}{\textbf{red bold}} and the second best is in \textcolor{blue}{\underline{blue underline}}.}
\label{tab_rq1_frequency}
\resizebox{\linewidth}{!}{%
\begin{tabular}{ll rrrrrrrrrrrr}
\toprule
Group & Metric & Chr2 & TiRex & TFM & Toto1 & Moi2 & ChrB & TabPFN & Sundial & ARIMA & ETS & MovAvg & SNaive \\
\midrule
\multirow{9}{*}{6min} & Rank (RMSE) $\downarrow$ & 3.00 & 4.00 & \textcolor{red}{\textbf{1.00}} & 7.00 & 5.00 & 8.00 & 6.00 & \textcolor{blue}{\underline{2.00}} & 9.50 & 12.00 & 9.50 & 11.00 \\
 & Win Rate (RMSE) $\uparrow$ & 0.818 & 0.727 & \textcolor{red}{\textbf{1.000}} & 0.455 & 0.636 & 0.364 & 0.545 & \textcolor{blue}{\underline{0.909}} & 0.227 & 0.000 & 0.227 & 0.091 \\
 & Elo (RMSE) $\uparrow$ & 1436 & 1310 & \textcolor{red}{\textbf{1780}} & 944 & 1186 & 820 & 1062 & \textcolor{blue}{\underline{1602}} & 626 & 208 & 625 & 400 \\
 & Rank (MAPE) $\downarrow$ & 4.00 & 6.00 & \textcolor{red}{\textbf{1.00}} & \textcolor{blue}{\underline{2.00}} & 3.00 & 8.00 & 5.00 & 7.00 & 10.00 & 9.00 & 11.00 & 12.00 \\
 & Win Rate (MAPE) $\uparrow$ & 0.727 & 0.545 & \textcolor{red}{\textbf{1.000}} & \textcolor{blue}{\underline{0.909}} & 0.818 & 0.364 & 0.636 & 0.455 & 0.182 & 0.273 & 0.091 & 0.000 \\
 & Elo (MAPE) $\uparrow$ & 1310 & 1068 & \textcolor{red}{\textbf{1783}} & \textcolor{blue}{\underline{1599}} & 1446 & 817 & 1182 & 947 & 549 & 692 & 399 & 206 \\
 & Rank (CRPS) $\downarrow$ & 3.00 & 4.00 & \textcolor{red}{\textbf{1.00}} & 5.00 & 6.00 & 8.00 & 7.00 & \textcolor{blue}{\underline{2.00}} & 9.00 & 10.00 & 11.00 & 12.00 \\
 & Win Rate (CRPS) $\uparrow$ & 0.818 & 0.727 & \textcolor{red}{\textbf{1.000}} & 0.636 & 0.545 & 0.364 & 0.455 & \textcolor{blue}{\underline{0.909}} & 0.273 & 0.182 & 0.091 & 0.000 \\
 & Elo (CRPS) $\uparrow$ & 1436 & 1314 & \textcolor{red}{\textbf{1779}} & 1190 & 1067 & 818 & 948 & \textcolor{blue}{\underline{1602}} & 693 & 547 & 399 & 206 \\
\midrule
\multirow{9}{*}{10min} & Rank (RMSE) $\downarrow$ & 9.00 & 10.00 & 7.00 & 12.00 & 3.00 & \textcolor{red}{\textbf{1.00}} & \textcolor{blue}{\underline{2.00}} & 4.00 & 5.00 & 11.00 & 6.00 & 8.00 \\
 & Win Rate (RMSE) $\uparrow$ & 0.273 & 0.182 & 0.455 & 0.000 & 0.818 & \textcolor{red}{\textbf{1.000}} & \textcolor{blue}{\underline{0.909}} & 0.727 & 0.636 & 0.091 & 0.545 & 0.364 \\
 & Elo (RMSE) $\uparrow$ & 685 & 559 & 941 & 214 & 1451 & \textcolor{red}{\textbf{1784}} & \textcolor{blue}{\underline{1601}} & 1317 & 1184 & 399 & 1054 & 811 \\
 & Rank (MAPE) $\downarrow$ & 3.50 & 5.00 & \textcolor{red}{\textbf{1.00}} & 7.00 & 9.00 & 8.00 & 10.00 & 11.00 & \textcolor{blue}{\underline{2.00}} & 12.00 & 3.50 & 6.00 \\
 & Win Rate (MAPE) $\uparrow$ & 0.773 & 0.636 & \textcolor{red}{\textbf{1.000}} & 0.455 & 0.273 & 0.364 & 0.182 & 0.091 & \textcolor{blue}{\underline{0.909}} & 0.000 & 0.773 & 0.545 \\
 & Elo (MAPE) $\uparrow$ & 1379 & 1189 & \textcolor{red}{\textbf{1778}} & 936 & 692 & 816 & 560 & 405 & \textcolor{blue}{\underline{1595}} & 211 & 1375 & 1064 \\
 & Rank (CRPS) $\downarrow$ & 7.00 & 6.00 & 10.00 & \textcolor{blue}{\underline{2.00}} & \textcolor{red}{\textbf{1.00}} & 3.00 & 5.00 & 4.00 & 9.00 & 11.00 & 8.00 & 12.00 \\
 & Win Rate (CRPS) $\uparrow$ & 0.455 & 0.545 & 0.182 & \textcolor{blue}{\underline{0.909}} & \textcolor{red}{\textbf{1.000}} & 0.818 & 0.636 & 0.727 & 0.273 & 0.091 & 0.364 & 0.000 \\
 & Elo (CRPS) $\uparrow$ & 943 & 1062 & 557 & \textcolor{blue}{\underline{1602}} & \textcolor{red}{\textbf{1781}} & 1454 & 1178 & 1306 & 689 & 401 & 818 & 208 \\
\midrule
\multirow{9}{*}{15min} & Rank (RMSE) $\downarrow$ & 3.00 & \textcolor{red}{\textbf{1.00}} & \textcolor{blue}{\underline{2.00}} & 12.00 & 6.00 & 11.00 & 9.00 & 5.00 & 8.00 & 4.00 & 7.00 & 10.00 \\
 & Win Rate (RMSE) $\uparrow$ & 0.818 & \textcolor{red}{\textbf{1.000}} & \textcolor{blue}{\underline{0.909}} & 0.000 & 0.545 & 0.091 & 0.273 & 0.636 & 0.364 & 0.727 & 0.455 & 0.182 \\
 & Elo (RMSE) $\uparrow$ & 1456 & \textcolor{red}{\textbf{1784}} & \textcolor{blue}{\underline{1599}} & 216 & 1064 & 391 & 690 & 1187 & 812 & 1313 & 944 & 544 \\
 & Rank (MAPE) $\downarrow$ & 5.00 & \textcolor{red}{\textbf{1.00}} & \textcolor{blue}{\underline{2.50}} & 8.00 & 5.00 & 8.00 & 5.00 & \textcolor{blue}{\underline{2.50}} & 12.00 & 8.00 & 10.50 & 10.50 \\
 & Win Rate (MAPE) $\uparrow$ & 0.636 & \textcolor{red}{\textbf{1.000}} & \textcolor{blue}{\underline{0.864}} & 0.364 & 0.636 & 0.364 & 0.636 & \textcolor{blue}{\underline{0.864}} & 0.000 & 0.364 & 0.136 & 0.136 \\
 & Elo (MAPE) $\uparrow$ & 1176 & \textcolor{red}{\textbf{1763}} & 1500 & 831 & 1175 & 826 & 1179 & \textcolor{blue}{\underline{1509}} & 223 & 825 & 496 & 497 \\
 & Rank (CRPS) $\downarrow$ & 10.50 & 10.50 & 12.00 & 4.50 & \textcolor{blue}{\underline{2.00}} & 4.50 & 3.00 & \textcolor{red}{\textbf{1.00}} & 8.00 & 6.00 & 9.00 & 7.00 \\
 & Win Rate (CRPS) $\uparrow$ & 0.136 & 0.136 & 0.000 & 0.682 & \textcolor{blue}{\underline{0.909}} & 0.682 & 0.818 & \textcolor{red}{\textbf{1.000}} & 0.364 & 0.545 & 0.273 & 0.455 \\
 & Elo (CRPS) $\uparrow$ & 483 & 484 & 223 & 1253 & \textcolor{blue}{\underline{1592}} & 1250 & 1444 & \textcolor{red}{\textbf{1782}} & 804 & 1069 & 676 & 941 \\
\midrule
\multirow{9}{*}{1h} & Rank (RMSE) $\downarrow$ & \textcolor{blue}{\underline{3.67}} & \textcolor{blue}{\underline{3.67}} & \textcolor{red}{\textbf{3.50}} & 10.67 & 4.83 & 6.83 & 6.00 & 7.83 & 7.17 & 7.50 & 8.00 & 8.33 \\
 & Win Rate (RMSE) $\uparrow$ & \textcolor{blue}{\underline{0.758}} & \textcolor{blue}{\underline{0.758}} & \textcolor{red}{\textbf{0.773}} & 0.121 & 0.652 & 0.470 & 0.545 & 0.379 & 0.439 & 0.409 & 0.364 & 0.333 \\
 & Elo (RMSE) $\uparrow$ & 1165 & \textcolor{blue}{\underline{1200}} & \textcolor{red}{\textbf{1211}} & 686 & 1177 & 1025 & 1117 & 989 & 870 & 809 & 829 & 923 \\
 & Rank (MAPE) $\downarrow$ & 5.42 & 4.83 & 5.17 & \textcolor{blue}{\underline{3.75}} & \textcolor{red}{\textbf{3.17}} & 5.08 & 4.33 & 6.33 & 9.17 & 9.83 & 10.42 & 10.50 \\
 & Win Rate (MAPE) $\uparrow$ & 0.598 & 0.652 & 0.621 & \textcolor{blue}{\underline{0.750}} & \textcolor{red}{\textbf{0.803}} & 0.629 & 0.697 & 0.515 & 0.258 & 0.197 & 0.144 & 0.136 \\
 & Elo (MAPE) $\uparrow$ & 1123 & 1192 & 1027 & \textcolor{red}{\textbf{1350}} & 1254 & 1181 & \textcolor{blue}{\underline{1261}} & 1013 & 757 & 603 & 597 & 642 \\
 & Rank (CRPS) $\downarrow$ & 7.67 & 7.50 & 10.17 & \textcolor{red}{\textbf{1.83}} & \textcolor{blue}{\underline{2.17}} & 3.25 & 4.83 & 3.75 & 8.67 & 9.42 & 9.25 & 9.50 \\
 & Win Rate (CRPS) $\uparrow$ & 0.394 & 0.409 & 0.167 & \textcolor{red}{\textbf{0.924}} & \textcolor{blue}{\underline{0.894}} & 0.795 & 0.652 & 0.750 & 0.303 & 0.235 & 0.250 & 0.227 \\
 & Elo (CRPS) $\uparrow$ & 810 & 860 & 601 & \textcolor{red}{\textbf{1685}} & \textcolor{blue}{\underline{1574}} & 1413 & 1200 & 1356 & 670 & 588 & 608 & 637 \\
\midrule
\multirow{9}{*}{1D} & Rank (RMSE) $\downarrow$ & \textcolor{red}{\textbf{1.00}} & 3.00 & \textcolor{blue}{\underline{2.00}} & 12.00 & 4.00 & 10.00 & 5.00 & 6.00 & 9.00 & 8.00 & 7.00 & 11.00 \\
 & Win Rate (RMSE) $\uparrow$ & \textcolor{red}{\textbf{1.000}} & 0.818 & \textcolor{blue}{\underline{0.909}} & 0.000 & 0.727 & 0.182 & 0.636 & 0.545 & 0.273 & 0.364 & 0.455 & 0.091 \\
 & Elo (RMSE) $\uparrow$ & \textcolor{red}{\textbf{1772}} & 1441 & \textcolor{blue}{\underline{1602}} & 215 & 1316 & 550 & 1195 & 1071 & 683 & 821 & 947 & 387 \\
 & Rank (MAPE) $\downarrow$ & 4.00 & \textcolor{blue}{\underline{2.00}} & 3.00 & 11.00 & \textcolor{red}{\textbf{1.00}} & 7.00 & 5.00 & 6.00 & 8.00 & 10.00 & 9.00 & 12.00 \\
 & Win Rate (MAPE) $\uparrow$ & 0.727 & \textcolor{blue}{\underline{0.909}} & 0.818 & 0.091 & \textcolor{red}{\textbf{1.000}} & 0.455 & 0.636 & 0.545 & 0.364 & 0.182 & 0.273 & 0.000 \\
 & Elo (MAPE) $\uparrow$ & 1316 & \textcolor{blue}{\underline{1600}} & 1447 & 407 & \textcolor{red}{\textbf{1774}} & 935 & 1193 & 1069 & 822 & 546 & 684 & 208 \\
 & Rank (CRPS) $\downarrow$ & 6.00 & 5.00 & 8.00 & 7.00 & \textcolor{red}{\textbf{1.00}} & 4.00 & 3.00 & \textcolor{blue}{\underline{2.00}} & 9.00 & 11.00 & 10.00 & 12.00 \\
 & Win Rate (CRPS) $\uparrow$ & 0.545 & 0.636 & 0.364 & 0.455 & \textcolor{red}{\textbf{1.000}} & 0.727 & 0.818 & \textcolor{blue}{\underline{0.909}} & 0.273 & 0.091 & 0.182 & 0.000 \\
 & Elo (CRPS) $\uparrow$ & 1063 & 1183 & 818 & 949 & \textcolor{red}{\textbf{1787}} & 1302 & 1440 & \textcolor{blue}{\underline{1596}} & 698 & 402 & 552 & 209 \\
\bottomrule
\end{tabular}}
\end{table*}

\begin{table*}[t]
\centering
\caption{Zero-shot rankings on the online benchmark aggregated by forecast horizon. For each prediction length and metric, models are first ranked within each dataset and then ranked by their average rank across datasets sharing that prediction length (best${}={}1$). In each row the best available model is in \textcolor{red}{\textbf{red bold}} and the second best is in \textcolor{blue}{\underline{blue underline}}.}
\label{tab_rq1_predlen}
\resizebox{\linewidth}{!}{%
\begin{tabular}{ll rrrrrrrrrrrr}
\toprule
Pred. length & Metric & Chr2 & TiRex & TFM & Toto1 & Moi & Bolt & TabPFN & Sundial & ARIMA & ETS & MovAvg & SNaive \\
\midrule
\multirow{3}{*}{7} & Rank (RMSE) & \textcolor{red}{\textbf{1}} & 3 & \textcolor{blue}{\underline{2}} & 12 & 4 & 7 & 5 & 6 & 10 & 9 & 8 & 11 \\
 & Rank (MAPE) & 3 & \textcolor{blue}{\underline{2}} & 5 & 8 & \textcolor{red}{\textbf{1}} & 7 & 4 & 6 & 9 & 11 & 10 & 12 \\
 & Rank (CRPS) & 6 & 5 & 8 & 7 & \textcolor{blue}{\underline{2}} & 4 & 3 & \textcolor{red}{\textbf{1}} & 9 & 11 & 10 & 12 \\
\cmidrule(l){2-14}
\multirow{3}{*}{24} & Rank (RMSE) & \textcolor{blue}{\underline{2}} & 3 & \textcolor{red}{\textbf{1}} & 12 & 4 & 6 & 5 & 9 & 7 & 8 & 10 & 11 \\
 & Rank (MAPE) & 5 & 4 & 6 & \textcolor{blue}{\underline{2}} & \textcolor{red}{\textbf{1}} & 8 & 3 & 7 & 10 & 9 & 11 & 12 \\
 & Rank (CRPS) & 7 & 6 & 9 & \textcolor{blue}{\underline{2}} & \textcolor{red}{\textbf{1}} & 4 & 5 & 3 & 8 & 10 & 12 & 11 \\
\cmidrule(l){2-14}
\multirow{3}{*}{60} & Rank (RMSE) & 3 & 4 & \textcolor{red}{\textbf{1}} & 7 & 5 & 8 & 6 & \textcolor{blue}{\underline{2}} & 9 & 12 & 10 & 11 \\
 & Rank (MAPE) & 4 & 7 & \textcolor{red}{\textbf{1}} & \textcolor{blue}{\underline{2}} & 3 & 8 & 5 & 6 & 10 & 9 & 11 & 12 \\
 & Rank (CRPS) & 3 & 5 & \textcolor{red}{\textbf{1}} & 4 & 6 & 8 & 7 & \textcolor{blue}{\underline{2}} & 9 & 10 & 11 & 12 \\
\cmidrule(l){2-14}
\multirow{3}{*}{72} & Rank (RMSE) & 8 & 10 & 7 & 12 & \textcolor{red}{\textbf{1}} & 3 & \textcolor{blue}{\underline{2}} & 4 & 5 & 11 & 6 & 9 \\
 & Rank (MAPE) & 7 & 10 & \textcolor{red}{\textbf{1}} & 4 & 5 & 9 & 6 & 11 & \textcolor{blue}{\underline{2}} & 12 & 3 & 8 \\
 & Rank (CRPS) & 7 & 6 & 10 & \textcolor{blue}{\underline{2}} & \textcolor{red}{\textbf{1}} & 3 & 4 & 5 & 9 & 11 & 8 & 12 \\
\bottomrule
\end{tabular}}
\end{table*}

\begin{figure*}[!t]
\centering
\includegraphics[width=0.6\textwidth]{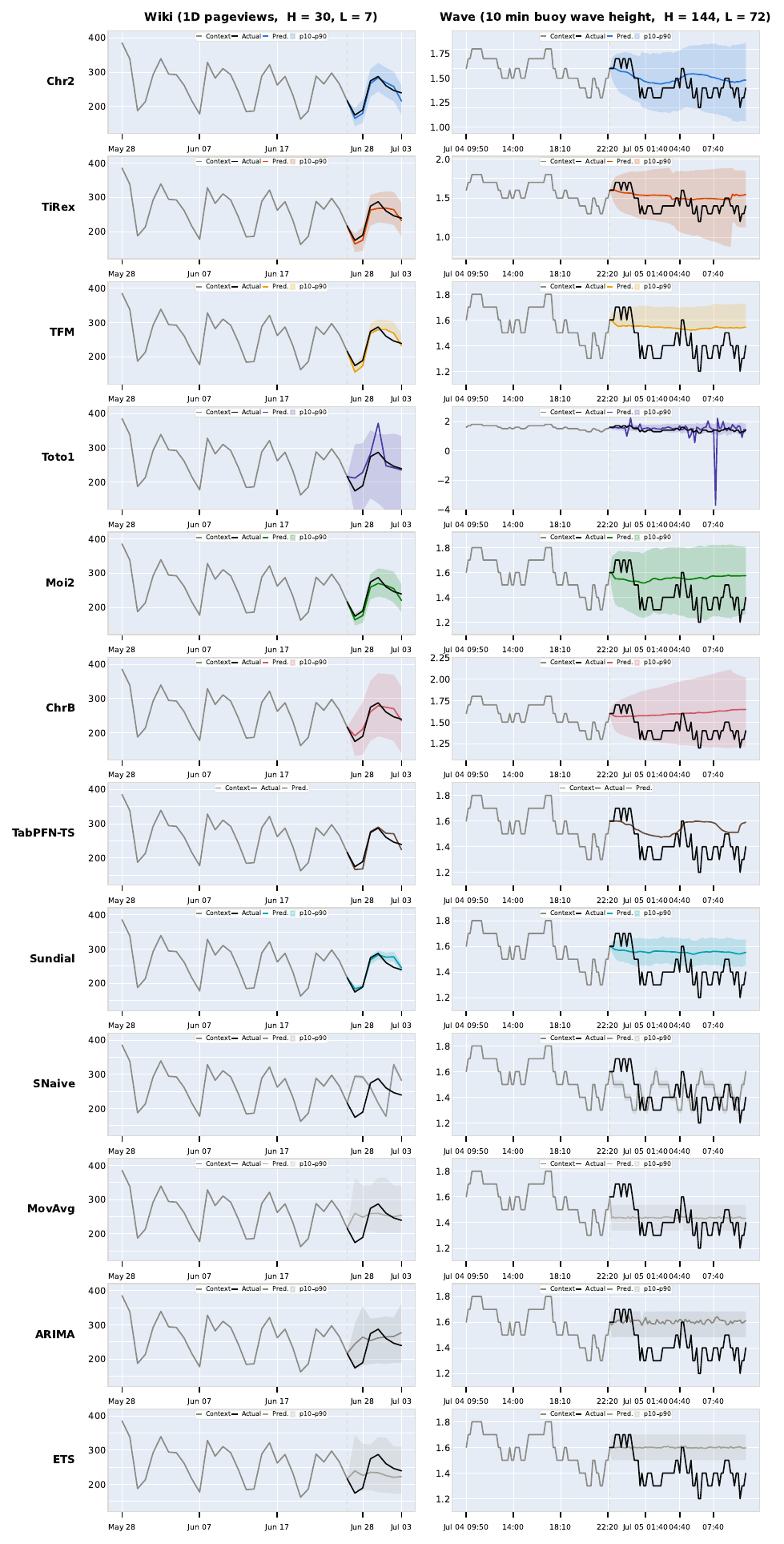}
\caption{Forecasting visualizations on Wiki (left column) and Wave (right column) for all baselines.
Each row corresponds to one baseline. 
The left side of the dash line shows the context windows, and the right side shows the forecasting horizons. }
\label{fig_appendix_all_models}
\end{figure*}

\begin{figure*}[!t]
\centering
\includegraphics[width=0.6\textwidth]{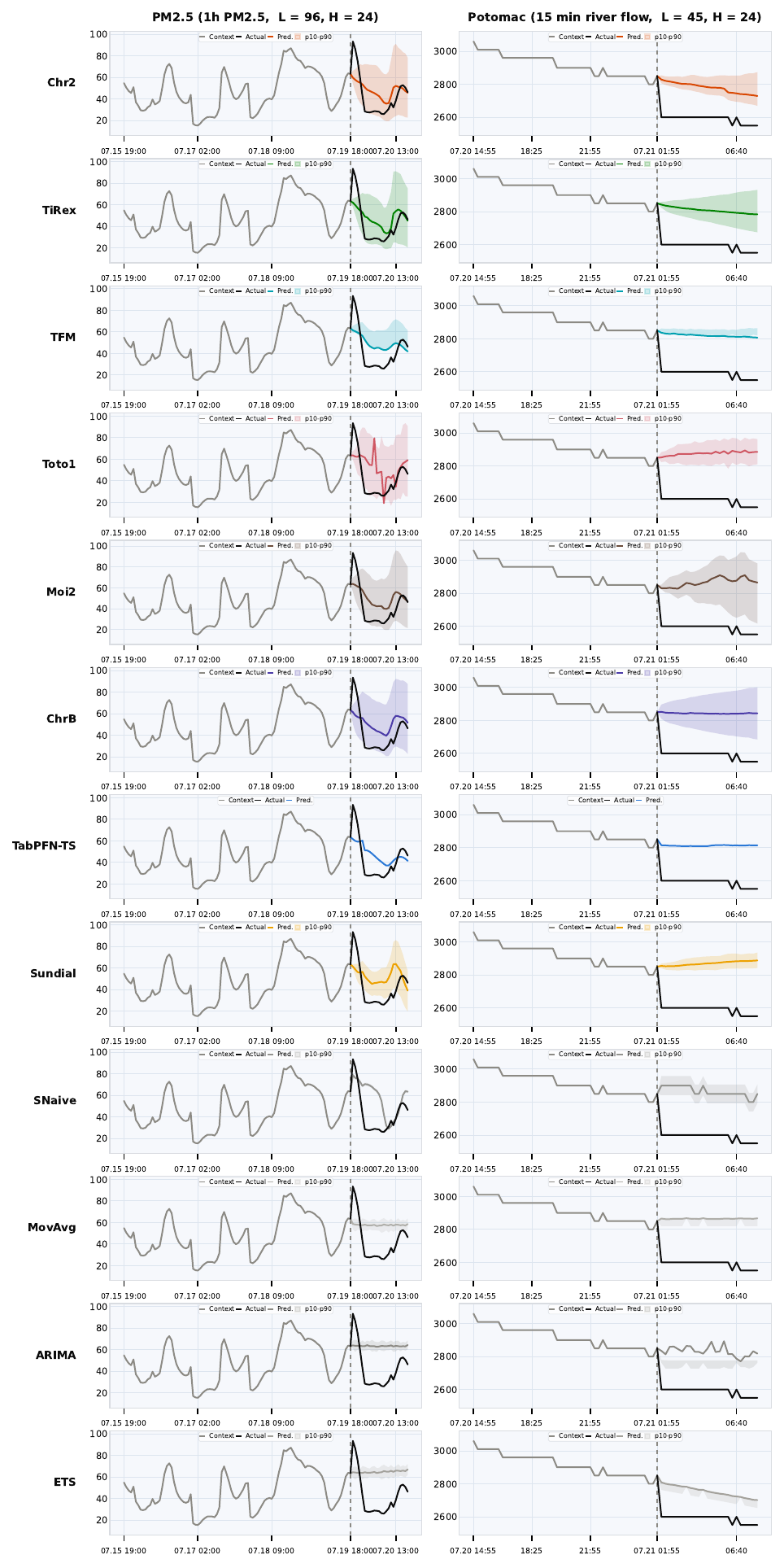}
\caption{Forecasting visualizations on PM~2.5 (left column) and Potomac (right column) for all baselines.
Each row corresponds to one baseline. 
The left side of the dash line shows the context windows, and the right side shows the forecasting horizons. }
\label{fig_appendix_all_models2}
\end{figure*}

\clearpage

\twocolumn[{%
\begin{minipage}{\textwidth}
    \centering

    \includegraphics[
        width=\textwidth,
        height=0.55\textheight,
        keepaspectratio
    ]{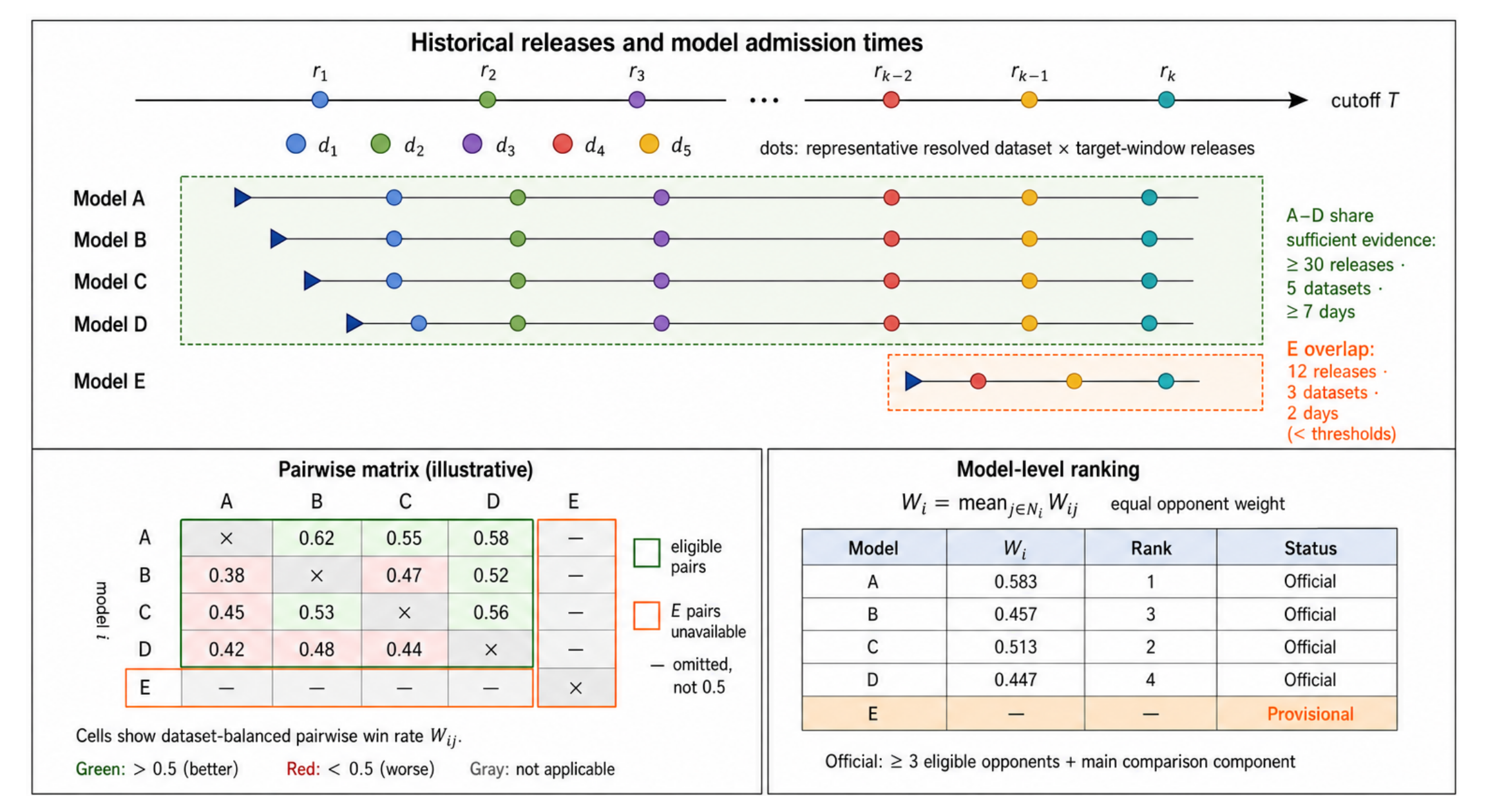}

    \vspace{-3mm}

    \captionof{figure}{
        Pairwise historical ranking under asynchronous model admission.
        The top panel shows representative resolved releases and model
        admission times. Models A--D have sufficient shared evidence, whereas
        Model E joined later and remains below the eligibility thresholds.
        The lower-left panel shows dataset-balanced pairwise win rates
        $W_{ij}$; unavailable comparisons involving Model E are omitted rather
        than treated as ties. The lower-right panel aggregates eligible
        pairwise comparisons into model-level scores and official ranks.
        Numerical values are illustrative.
    }
    \label{fig:pairwise-historical-ranking}

    \vspace{2mm}
\end{minipage}
}]

\section{Pairwise Historical Ranking}
\label{app:pairwise-historical-ranking}

Models may enter a live leaderboard at different times. Directly comparing
their average errors since admission can therefore be misleading: an early
model may have experienced both volatile and calm periods, whereas a recently
admitted model may have been evaluated only under the latest conditions. We
address this cohort mismatch by comparing each pair of models only on releases
that they completed in common.

 A \emph{release} is one resolved causal forecasting task for a particular
dataset and future target window. At issue time, all admitted models receive
the same historical context, sampling frequency, and forecast horizon. Their
forecasts are frozen before the future target becomes available. Once the
complete target window has been observed, the forecasts are scored and the
task becomes a resolved release. Models evaluated on the same release therefore
share the same context, target values, timestamps, normalization statistics,
and metric implementation.

Figure~\ref{fig:pairwise-historical-ranking} connects the three stages of the
ranking procedure. In the top panel, the horizontal axis represents historical time up to cutoff
$T$. Each colored dot denotes a representative resolved
dataset--target-window release, with colors indicating different datasets.
The dots are illustrative; the actual ranking uses every resolved release.
The blue triangle on each model row marks its admission time. A model is
evaluated only on releases issued after admission. Models A--D entered sufficiently early to accumulate overlapping evaluations
from at least 30 releases, 5 datasets, and 7 days. Model E entered later and
has only 12 shared releases from 3 datasets over 2 days. Its overlap is
therefore insufficient for an official comparison. The lower-left panel records the pairwise win rate $W_{ij}$ of row model $i$
against column model $j$. For example, $W_{AB}=0.62$ means that Model A obtains
a dataset-balanced win rate of $62\%$ against Model B. The reverse comparison
is $W_{BA}=0.38$. Green cells indicate values above $0.5$, red cells indicate
values below $0.5$, and gray cells are unavailable. In particular, missing
comparisons involving Model E are omitted; they are not assigned a neutral
value of $0.5$. The lower-right panel averages each model's eligible pairwise win rates with
equal weight per opponent. This produces the model-level score $W_i$ used for
ranking. Models A--D receive official ranks, while Model E remains
\emph{Provisional} until it accumulates sufficient shared evidence.

\subsection{Shared releases and pair eligibility}

Let $t_r$ denote the end time of the target window for release $r$. At
historical cutoff $T$, define

\begin{equation}
R_i(T)
=
\left\{
r:
t_r\leq T,\;
\mathrm{MSE}_{ir}\ \text{and}\ \mathrm{CRPS}_{ir}
\ \text{are valid}
\right\}
\end{equation}

as the valid release history of model $i$. The shared release history of
models $i$ and $j$ is

\begin{equation}
R_{ij}(T)
=
R_i(T)\cap R_j(T).
\label{eq:shared-releases}
\end{equation}

Thus, a release completed by only one member of the pair does not enter their
comparison. Releases issued before the later model's admission are
automatically excluded. Let $d_r$ be the dataset associated with release $r$. The shared dataset set is

\begin{equation}
D_{ij}(T)
=
\left\{
d_r:r\in R_{ij}(T)
\right\},
\end{equation}

and the temporal coverage of the pair is

\begin{equation}
\operatorname{span}\!\left(R_{ij}(T)\right)
=
\max_{r\in R_{ij}(T)}t_r
-
\min_{r\in R_{ij}(T)}t_r.
\end{equation}

A pair is eligible only if

\begin{equation}
|R_{ij}(T)|\geq 30,
\qquad
|D_{ij}(T)|\geq 5,
\qquad
\operatorname{span}\!\left(R_{ij}(T)\right)
\geq 7\ \text{days}.
\label{eq:pair-eligibility}
\end{equation}

These requirements prevent a model from receiving an official comparison
based on a short period, a small number of releases, or a narrow selection of
datasets. A pair that fails any requirement is treated as unavailable rather
than as a tie.

\subsection{Comparing point and probabilistic forecasts}

The leaderboard includes both probabilistic models and models that produce
only point forecasts. Both model types are evaluated using MSE and CRPS under
the same context-only normalization.

Let $z_{rh}$ be the normalized target at horizon step $h$. A probabilistic
model provides a predictive mean $\mu_{irh}$ and quantiles
$Q_{irh}(\tau)$ for

\begin{equation}
\mathcal{Q}
=
\{0.1,0.2,\ldots,0.9\}.
\end{equation}

Its MSE is computed from the predictive mean, while its CRPS is approximated
using the quantile forecasts:

\begin{equation}
\mathrm{CRPS}_{ir}
=
\frac{2}{H_r|\mathcal{Q}|}
\sum_{h=1}^{H_r}
\sum_{\tau\in\mathcal{Q}}
\rho_\tau\!\left(
z_{rh}-Q_{irh}(\tau)
\right),
\label{eq:quantile-crps}
\end{equation}

where

\begin{equation}
\rho_\tau(u)
=
u\left(
\tau-\mathbb{I}\{u<0\}
\right)
\end{equation}

is the pinball loss.

A point-only model provides one forecast $\hat z_{irh}$ per horizon step. We
represent it as a degenerate predictive distribution by assigning the point
forecast to every required quantile:

\begin{equation}
Q^{\mathrm{point}}_{irh}(\tau)
=
\hat z_{irh},
\qquad
\forall\tau\in\mathcal{Q}.
\label{eq:point-quantile-conversion}
\end{equation}

This conversion does not add artificial uncertainty: the model assigns all
predictive mass to its point forecast. Because the quantile grid is symmetric
around $0.5$, substituting
Eq.~\eqref{eq:point-quantile-conversion} into
Eq.~\eqref{eq:quantile-crps} gives

\begin{equation}
\mathrm{CRPS}^{\mathrm{point}}_{ir}
=
\frac{1}{H_r}
\sum_{h=1}^{H_r}
\left|
z_{rh}-\hat z_{irh}
\right|.
\label{eq:point-crps-mae}
\end{equation}

Hence, the CRPS of a point-only model reduces to its normalized MAE.
Probabilistic models are evaluated on both the location and dispersion of
their predictive distributions, while point models are evaluated as
zero-uncertainty distributions. Since CRPS is defined for both ordinary and
degenerate predictive distributions, their scores remain directly comparable. Model output type and any benchmark-side procedure used to construct
quantiles are fixed before the target is observed. Native quantiles are used
when available; a declared point-only output is converted using
Eq.~\eqref{eq:point-quantile-conversion}.

\begin{figure*}[t]
    \centering
    \includegraphics[width=0.9\textwidth]
    {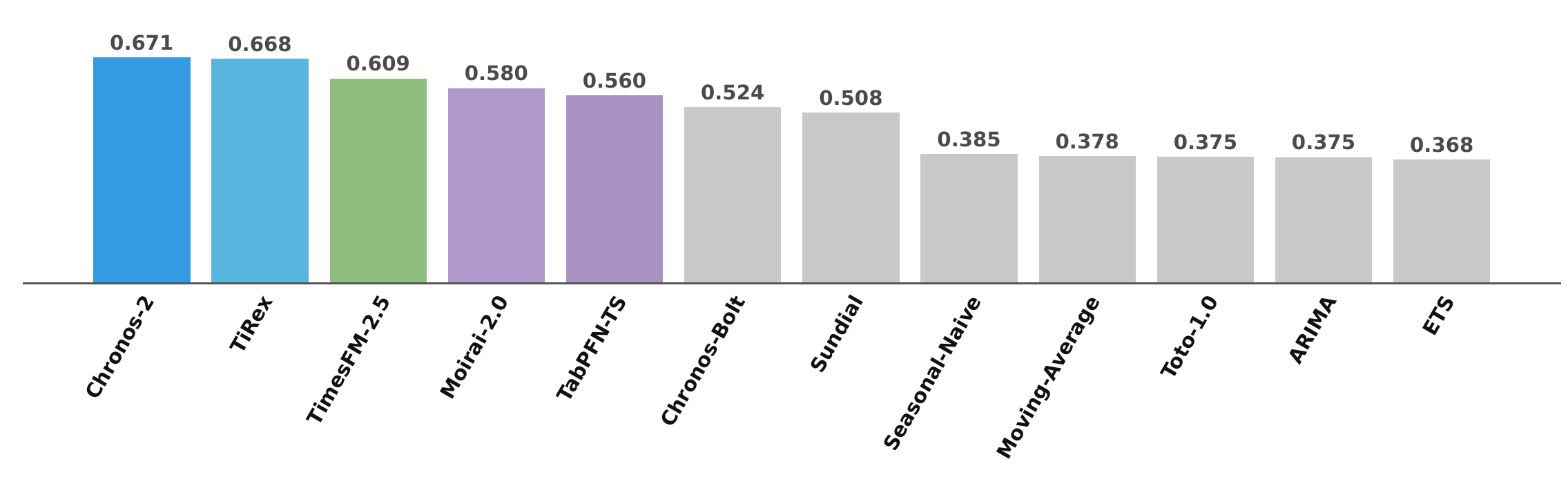}
    \vspace{-5mm}
    \caption{Overall model ranking in the latest leaderboard snapshot. Bars
    show the dataset-balanced pairwise win rate aggregated over eligible
    metric-specific comparisons, historical releases, and opponents.
    Chronos-2 achieves the highest overall score, followed closely by TiRex.
    Higher values are better.}
    \label{fig:overall-ranking}
\end{figure*}

\begin{figure*}[t]
    \centering
    \includegraphics[width=0.9\textwidth]{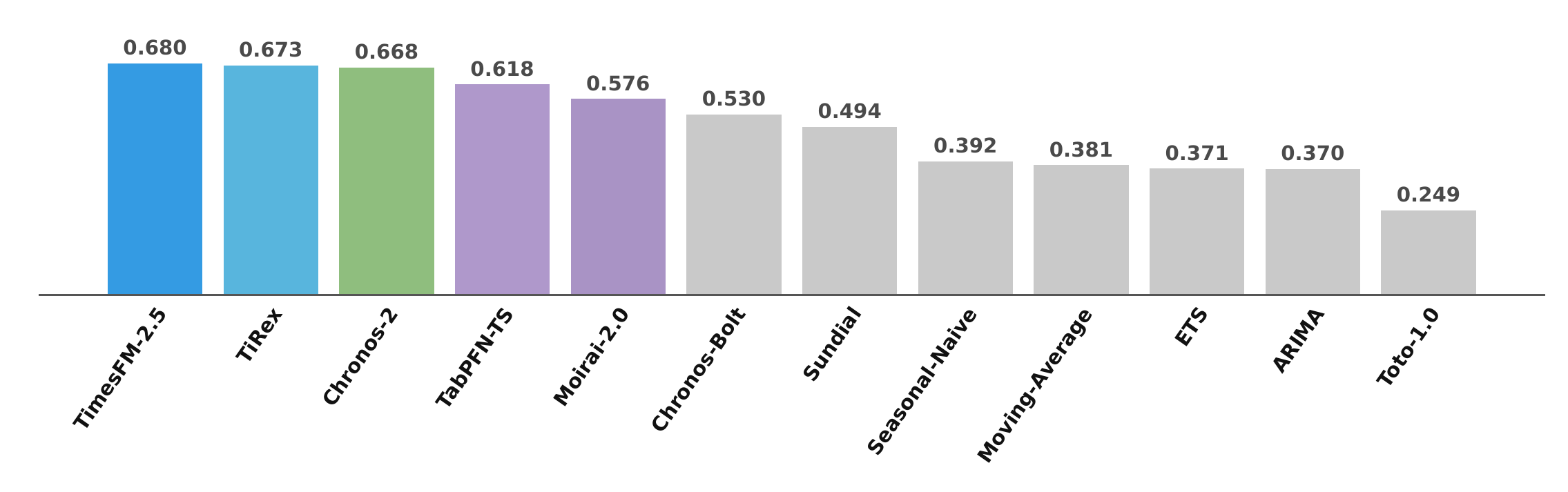}
    \vspace{-5mm}
    \caption{MSE-based ranking within the point-forecast track. Bars report
    dataset-balanced pairwise win rates over eligible historical releases and
    opponents. TimesFM-2.5 achieves the highest score. Higher values are
    better.}
    \label{fig:overall-ranking-mse}
\end{figure*}

\begin{figure*}[t]
    \centering    \includegraphics[width=0.9\textwidth]{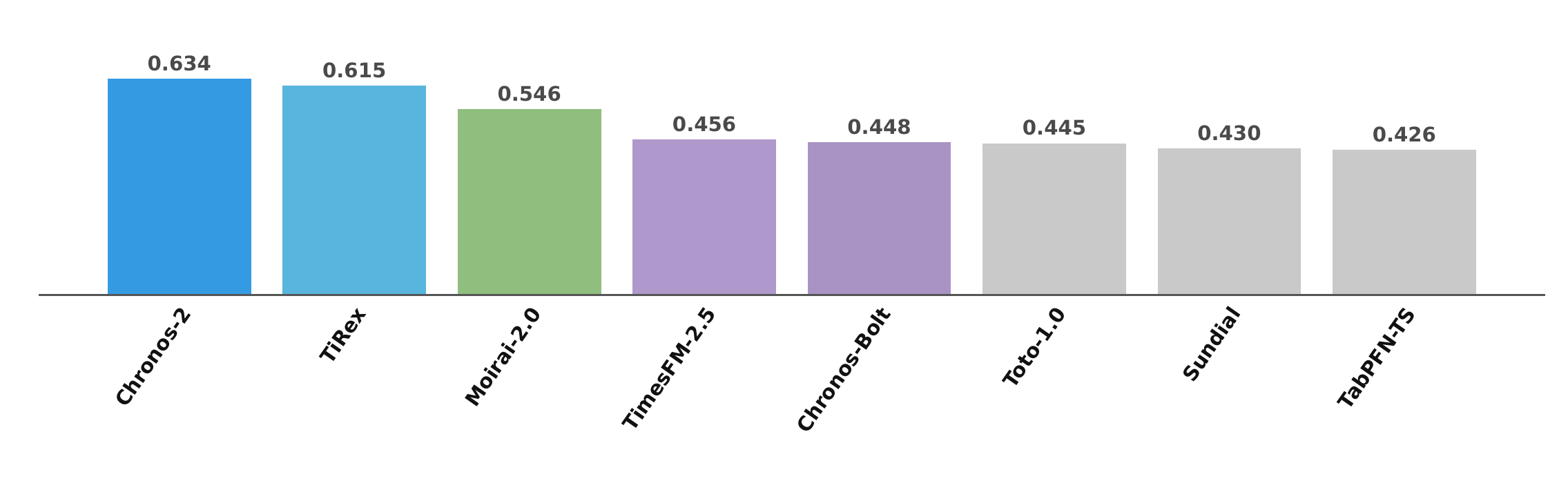}
    \vspace{-5mm}
    \caption{CRPS-based ranking restricted to the eight TSFMs whose model
    families natively support probabilistic forecasting through quantiles or
    forecast samples. Chronos-2 achieves the highest score. Higher values are
    better.}
    \label{fig:overall-ranking-crps}
\end{figure*}

\subsection{Release-level pairwise score}

Both MSE and CRPS are lower-is-better. For two error values $a$ and $b$, let

\begin{equation}
c(a,b)
=
\begin{cases}
1,   & a<b,\\
0.5, & a=b,\\
0,   & a>b.
\end{cases}
\label{eq:elementary-win}
\end{equation}

On shared release $r$, the score of model $i$ against model $j$ is

\begin{equation}
s_{ijr}
=
\frac{1}{2}
\left[
c\!\left(\mathrm{MSE}_{ir},\mathrm{MSE}_{jr}\right)
+
c\!\left(\mathrm{CRPS}_{ir},\mathrm{CRPS}_{jr}\right)
\right].
\label{eq:release-win}
\end{equation}

The two metrics receive equal weight. A model receives $s_{ijr}=1$ if it wins
on both metrics and $s_{ijr}=0$ if it loses on both. A split decision gives
$0.5$; a win and a tie give $0.75$; and a loss and a tie give $0.25$.
Therefore,

\begin{equation}
s_{ijr}\in\{0,0.25,0.5,0.75,1\},
\qquad
s_{jir}=1-s_{ijr}.
\end{equation}

The same rule applies to point--point, probabilistic--probabilistic, and
point--probabilistic comparisons. The MSE component compares central forecast
accuracy, while the CRPS component compares the corresponding predictive
distributions.

\subsection{Dataset-balanced pairwise win rate}

Datasets resolve releases at different rates. Pooling all releases directly
would allow high-frequency streams to dominate the ranking. We therefore
average scores within each shared dataset before averaging across datasets.

Let
\begin{equation}
R_{ij,d}(T)
=
\left\{
r\in R_{ij}(T):d_r=d
\right\}
\end{equation}

be the releases shared by models $i$ and $j$ for dataset $d$. Their
dataset-specific win rate is

\begin{equation}
W_{ij,d}(T)
=
\frac{1}{|R_{ij,d}(T)|}
\sum_{r\in R_{ij,d}(T)}
s_{ijr}.
\label{eq:dataset-win-rate}
\end{equation}

The dataset-balanced pairwise win rate is

\begin{equation}
W_{ij}(T)
=
\frac{1}{|D_{ij}(T)|}
\sum_{d\in D_{ij}(T)}
W_{ij,d}(T).
\label{eq:pairwise-win-rate}
\end{equation}

Every shared dataset therefore receives equal weight, regardless of how many
releases it produces. Complementarity of the release-level score implies

\begin{equation}
W_{ji}(T)=1-W_{ij}(T).
\end{equation}

This is visible in the lower-left panel of
Figure~\ref{fig:pairwise-historical-ranking}: for example,
$W_{AB}=0.62$ and $W_{BA}=0.38$. A value above $0.5$ means that model $i$
wins more often than model $j$ after dataset balancing; it is a ranking score,
not a statistical significance test.

\subsection{Model-level ranking and status}

Let $N_i(T)$ be the set of opponents for which model $i$ satisfies
Eq.~\eqref{eq:pair-eligibility}. The model-level score is the macro-average

\begin{equation}
W_i(T)
=
\frac{1}{|N_i(T)|}
\sum_{j\in N_i(T)}
W_{ij}(T).
\label{eq:model-win-rate}
\end{equation}

Each eligible opponent receives equal weight. Thus, an opponent with a longer
shared history does not dominate the final score merely because the pair has
more releases.

In Figure~\ref{fig:pairwise-historical-ranking}, Model A has

\begin{equation}
W_A
=
\frac{W_{AB}+W_{AC}+W_{AD}}{3}
=
\frac{0.62+0.55+0.58}{3}
=
0.583,
\end{equation}

which gives it rank 1. Models are ordered by decreasing $W_i(T)$.

We construct an undirected comparison graph whose vertices are models and
whose edges are eligible model pairs. A model receives an official rank only
if

\begin{enumerate}
    \item it has at least three eligible opponents; and
    \item it belongs to the main connected component of the graph.
\end{enumerate}

Models that fail either condition remain \emph{Provisional}. Missing pairs are
omitted from Eq.~\eqref{eq:model-win-rate}; they are never imputed as $0.5$.
This is why Model E in the figure has neither a model-level score nor a rank,
despite having completed some releases.

Alongside the rank, the leaderboard reports the number of eligible opponents,
shared releases, shared datasets, and covered time span. These fields expose
the amount of evidence supporting each result and distinguish an established
ranking from a provisional one.

\section{Overall Model Ranking}
\label{app:overall-ranking}

Figures~\ref{fig:overall-ranking}--\ref{fig:overall-ranking-crps} provide three
complementary views of the latest leaderboard snapshot up to cutoff time \(T\).
Each score is a dataset-balanced pairwise win rate, averaged with equal weight
over eligible opponents and historical releases; higher values indicate better
relative performance. Figure~\ref{fig:overall-ranking} shows the overall
eligibility-aware aggregation, where Chronos-2 ranks first (\(0.671\)), narrowly
ahead of TiRex (\(0.668\)), followed by TimesFM-2.5 (\(0.609\)).

To separate point accuracy from distributional forecast quality,
Figure~\ref{fig:overall-ranking-mse} ranks models using MSE within the
point-forecast track. TimesFM-2.5 achieves the highest MSE-based win rate
(\(0.680\)), followed by TiRex (\(0.673\)) and Chronos-2 (\(0.668\)).
Figure~\ref{fig:overall-ranking-crps} restricts the comparison to the eight
TSFMs whose model families natively support probabilistic forecasts through
quantiles or samples. Under CRPS, Chronos-2 ranks first (\(0.634\)), followed by
TiRex (\(0.615\)) and Moirai-2.0 (\(0.546\)).

Together, the three views show that point-forecast accuracy and probabilistic
forecast quality need not produce the same ordering: TimesFM-2.5 leads the MSE
track, whereas Chronos-2 leads the CRPS track and the overall aggregation.
TiRex remains consistently competitive across all three views. Because the MSE
and CRPS rankings use different eligible model sets and opponent groups, their
absolute scores should be interpreted within each figure rather than compared
directly across tracks. These rankings summarize relative historical
performance and do not by themselves establish statistical significance or
long-term rank stability.


\end{document}